\documentclass{article}
\usepackage[utf8]{inputenc}
\usepackage[T1]{fontenc}
\usepackage{graphicx}
\usepackage{grffile}
\usepackage{longtable}
\usepackage{wrapfig}
\usepackage{rotating}
\usepackage[normalem]{ulem}
\usepackage{amsmath}
\usepackage{textcomp}
\usepackage{amssymb}
\usepackage{capt-of}
\usepackage{multicol}
\usepackage{cancel}
\usepackage{MnSymbol}
\usepackage[dvipsnames]{xcolor}
\usepackage[boxruled,lined]{algorithm2e}
\usepackage{subfiles}
\usepackage{enumitem}

\usepackage[accepted]{icml2024}

\usepackage{hyperref}
\definecolor{mydarkblue}{rgb}{0,0.08,0.45}
\hypersetup{ %
  pdftitle={},
  pdfsubject={Proceedings of the International Conference on Machine Learning 2024},
  pdfkeywords={},
  pdfborder=0 0 0,
  pdfpagemode=UseNone,
  colorlinks=true,
  linkcolor=mydarkblue,
  citecolor=mydarkblue,
  filecolor=mydarkblue,
  urlcolor=mydarkblue,
  }

\usepackage{amsmath,amssymb}
\usepackage{mathtools}
\usepackage{amsthm}
\usepackage{thmtools,thm-restate}

\newcommand{\SecVAW}{The Vovk-Azoury-Warmuth Forecaster}
\newcommand{\SecDiscountedVAW}{Dynamic Regret via Discounting}
  \newcommand{\SecTuning}{Learning the Optimal Discount Factor}
  \newcommand{\SecSmallLoss}{Small-loss Bounds via Self-confident Predictions}

\newcommand{\SecLB}{Dimension-dependent Lower Bound}

\newcommand{\StaticVAWUpdateDiscussion}[1]{}

\makeatletter
\newcommand{\doublewidetilde}[1]{{%
  \mathpalette\double@widetilde{#1}%
}}
\newcommand{\double@widetilde}[2]{%
  \sbox\z@{$\m@th#1\widetilde{#2}$}%
  \ht\z@=.9\ht\z@
  \widetilde{\box\z@}%
}
\makeatother

\makeatletter
\newcommand{\oset}[3][0ex]{%
  \mathrel{\mathop{#3}\limits^{
    \vbox to#1{\kern-2\ex@
    \hbox{$\scriptstyle#2$}\vss}}}}
\makeatother

\usepackage{stackengine}
\newcommand\URcorneraccent{\rule[\dimexpr1.7pt-.5pt]{1ex}{.5pt}\rule{.5pt}{1.7pt}}
\newcommand\URcorner[1]{\ensurestackMath{\stackon[.8pt]{#1}{\URcorneraccent}}}

\newcommand{\Clip}{\mathrm{Clip}}

\newcommand{\Ring}{\circ{}}
\newcommand{\yhint}{\ytilde}
\newcommand{\yref}{\y^{\text{Ref}}}
\newcommand{\ypred}{\ybar}
\newcommand{\yclip}{\URcorner{\y}}

\newcommand{\x}{x}
\newcommand{\xt}{\x_{t}}
\newcommand{\xtmm}{\x_{\tmm}}

\newcommand{\y}{y}
\newcommand{\yt}{\y_{t}}

\newcommand{\ytmm}{\y_{\tmm}}
\newcommand{\yhat}{\hat\y}

\renewcommand*\hat\widehat
\renewcommand*\tilde\widetilde
\renewcommand*{\vec}[1]{\boldsymbol{#1}}

\newcommand{\Mod}[1]{\ (\mathrm{mod}\ #1)}

\newcommand{\ybar}{\overline{y}}

\newcommand{\Meta}{\text{Meta}}
\newcommand{\ytilde}{\tilde{\y}}

\newcommand{\Ceil}[1]{\left\lceil#1\right\rceil}
\newcommand{\Floor}[1]{\left\lfloor#1\right\rfloor}

\usepackage{bbm}

\newcommand{\Sign}[1]{\text{Sign}\left(#1\right)}

\newcommand{\cmp}{u}

\newcommand{\w}{w}
\newcommand{\wt}{\w_t}
\newcommand{\wtpp}{\w_\tpp}
\newcommand{\wtmm}{\w_\tmm}

\newcommand{\xtpp}{x_{\tpp}}

\newcommand{\maxOp}{\vee}
\newcommand{\minOp}{\wedge}
\newcommand{\Max}[1]{\max\Set{#1}}
\newcommand{\Min}[1]{\min\Set{#1}}

\newcommand{\wrt}{w.r.t}

\newcommand{\sumtT}{\sum_{t=1}^T}

\usepackage{tikz}

\newcommand{\R}{\mathbb{R}}

\newcommand{\N}{\mathbb{N}}

\newcommand{\Set}[1]{\left\{#1\right\}}
\newcommand{\sbrac}[1]{\left[#1\right]}
\newcommand{\brac}[1]{\left(#1\right)}
\newcommand{\cA}{\mathcal{A}}

\newcommand{\cB}{\mathcal{B}}

\newcommand{\cS}{\mathcal{S}}
\newcommand{\cY}{\mathcal{Y}}

\newcommand{\cV}{\mathcal{V}}

\DeclareMathOperator*{\argmin}{arg\ min~}

\newcommand{\norm}[1]{\left\|#1\right\|}

\newcommand{\grad}{\nabla}
\newcommand{\Det}[1]{\text{Det}\left(#1\right)}
\newcommand{\Tr}[1]{\text{Tr}\left(#1\right)}
\newcommand{\inner}[1]{\left\langle #1 \right\rangle}

\newcommand{\Log}[1]{\log\left(#1\right)}
\newcommand{\Exp}[1]{\exp\left(#1\right)}

\newcommand{\inv}{{-1}}

\newcommand{\E}{\mathbb E}

\newcommand{\EE}[1]{\E\left[ #1 \right]}

\newcommand{\abs}[1]{\left|#1\right|}

\newcommand{\zeros}{\mathbf{0}}

\newcommand{\half}{\frac{1}{2}}

\newcommand{\tpp}{{t+1}}
\newcommand{\tmm}{{t-1}}

\theoremstyle{plain}
\newtheorem{theorem}{Theorem}[section]

\theoremstyle{definition}

\theoremstyle{remark}
\newtheorem{remark}[theorem]{Remark}

\newtheorem{manualtheoreminner}{\normalfont{\textbf{Theorem}}}
\newenvironment{manualtheorem}[1]{%
  \renewcommand\themanualtheoreminner{\textbf{#1}}%
  \manualtheoreminner
}{\endmanualtheoreminner}

\newtheorem{manuallemmainner}{\normalfont{\textbf{Lemma}}}
\newenvironment{manuallemma}[1]{%
  \renewcommand\themanuallemmainner{#1}%
  \manuallemmainner
}{\endmanuallemmainner}

\SetKw{KwRequire}{Require}
\SetKw{KwInput}{Input}
\SetKw{KwInitialize}{Initialize}

\usepackage{tcolorbox}
\newcounter{BoxCounter}
\newcounter{PreBoxCounter}
\newenvironment{Boxed*}[2][\small]
{
  \begin{figure*}[!t]
    \begin{tcolorbox}[title=\textbf{Box \arabic{PreBoxCounter}:} #2, colback=white, colbacktitle=white, coltitle=black, arc=0pt,outer arc=0pt, fontupper=#1]
    }{
    \end{tcolorbox}
  \end{figure*}
}

\newcommand{\scratch}[1]{
  \ifscratchwork
  {\color{blue}
    \relax\ifmmode
    #1
    \else
    \ifscratchtxt
    #1
    \fi
    \fi
  }
  \else
  \relax\ifmmode
  \\[-2\baselineskip]
  \fi
  \fi
}

\def\ie{\textit{i.e.}}
\def\eg{\textit{e.g.}}

\usepackage[capitalize,noabbrev]{cleveref}

\icmltitlerunning{Online Linear Regression in Dynamic Environments via Discounting}

\begin{document}

\twocolumn[
\icmltitle{Online Linear Regression in Dynamic Environments via Discounting}

\icmlsetsymbol{equal}{*}

\begin{icmlauthorlist}
\icmlauthor{Andrew Jacobsen}{ualberta}
\icmlauthor{Ashok Cutkosky}{bu}
\end{icmlauthorlist}

\icmlaffiliation{ualberta}{Department of Computing Science, University of Alberta, Edmonton, Canada}
\icmlaffiliation{bu}{Department of Electrical and Computer Engineering, Boston University, Boston, Massachussetts}

\icmlcorrespondingauthor{Andrew Jacobsen}{ajjacobs@ualberta.ca}

\icmlkeywords{online learning, vovk-azoury-warmuth forecaster, dynamic regret,
  non-stationary, parameter-free, Machine Learning, ICML}

\vskip 0.3in
]

\printAffiliationsAndNotice{} %

\begin{abstract}
  We develop algorithms for online linear regression which achieve optimal static and dynamic regret guarantees \emph{even in the complete absence of prior knowledge}.  We present a novel analysis showing that a discounted variant of the Vovk-Azoury-Warmuth forecaster achieves dynamic regret of the form $R_{T}(\vec{u})\le O\left(d\log(T)\vee \sqrt{dP_{T}^{\gamma}(\vec{u})T}\right)$, where $P_{T}^{\gamma}(\vec{u})$ is a measure of variability of the comparator sequence, and show that the discount factor achieving this result can be learned on-the-fly. We show that this result is optimal by providing a matching lower bound.  We also extend our results to \emph{strongly-adaptive} guarantees which hold over every sub-interval $[a,b]\subseteq[1,T]$ simultaneously.
\end{abstract}

\section{Online Linear Regression}%
\label{sec:online-regression}

This paper presents new techniques and
analyses for online linear regression,
a variant of the classic least-squares regression problem
tailored to streaming data \citep{azoury2001relative,vovk2001competitive,orabona2015generalized,foster2017sparse}.
Formally, consider $T$ rounds of interaction between a learner and
an environment, in which learner's objective is to accurately
predict some observable target signal $\yt\in\R$ before it's revealed.
On each round, a vector of \emph{features} $\xt\in\R^{d}$ is first
revealed, representing the context of the environment
at the start of the round, and the learner predicts
$\yhat_{t}=\inner{\xt,\wt}$ by means of a weight vector $\wt\in\R^{d}$. The signal
$\yt\in\R^{d}$ is then observed, and the learner incurs a
loss proportional to the prediction error, $\ell_{t}(\wt)=\half(\yt-\inner{\xt,\wt})^{2}$.
Since $\wt$ is  allowed to depend on $\xt$, this protocol is sometimes referred to as \emph{improper} online regression,
as the learner is able to make predictions outside of the
class of linear models.
Indeed, since $\xt$ is revealed \emph{before}
the learner must make their prediction, it is always possible to make
predictions $\yhat_{t}=f_{t}(\xt)$ for any arbitrary transformation $f_{t}:\R^{d}\to\R$,
for instance by setting $\wt=f_{t}(\xt)\xt/\norm{\xt}^{2}$.

The classical measure of the learner's performance in this setting is
\emph{regret},
the cumulative prediction error relative to some fixed benchmark point
$\cmp\in\R^{d}$:
\begin{align*}
  R_{T}(\cmp)=\sumtT\ell_{t}(\wt)-\ell_{t}(\cmp).
\end{align*}
Notice that this performance measure can only properly reflect prediction
accuracy when there exists a \emph{fixed} $\cmp\in\R^{d}$ which predicts well on
average. For example, this may occur when when the $(\xt,\yt)$ pairs are all generated
\textit{i.i.d.} from some well-behaved distribution. However, in many true streaming settings the
data-generating distribution may change over time due to changes in
the environment. \emph{Dynamic} regret attempts to model such
settings by comparing against a \emph{sequence} of comparators
$\vec{\cmp}=(\cmp_{1},\ldots,\cmp_{T})$:
\begin{align*}
  R_{T}(\vec{\cmp})
  &=
    \sumtT\ell_{t}(\wt)-\ell_{t}(\cmp_{t}).
\end{align*}
Notice that dynamic regret captures the usual notion of regret (referred to as
\emph{static} regret) as a special case
by setting $\cmp_{1}=\ldots=\cmp_{T}$. Our goal in this work is
to make \emph{favorable dynamic regret guarantees
  even in the complete absence of any prior knowledge
  of the underlying data-generating process}.
Naturally, because such an algorithm leverages no
prior knowledge, it necessarily must be adaptive to
all problem-dependent quantities without
requiring any instance-specific hyperparameter tuning.

\textbf{Contributions. } In this work we achieve
the goal laid out above and
develop the first algorithms for online linear regression
that require no prior knowledge about the
data stream, yet still make strong performance guarantees.
In particular, our contributions are as follows:
\begin{itemize}[itemsep=0em,leftmargin=1em]
  \item We show that even in the absence of any
        boundedness assumptions,
        a discounted variant of the VAW forecaster
        with a well-chosen discount factor
        achieves dynamic regret $R_{T}(\vec{\cmp})\le O\big(d\Log{T}\maxOp \sqrt{d P_{T}^{\gamma}(\vec{\cmp})T}\big)$,
        where $ P_{T}^{\gamma}(\vec{\cmp})$ is a measure of variability of the
        comparator sequence (\ie{} the magnitude of $P_{T}^{\gamma}(\vec{\cmp})$
        is related to how drastically the comparator changes over time).
        We also obtain \emph{small-loss} guarantees
        of the form
        $R_{T}(\vec{\cmp})\le O\Big(d\Log{T}\maxOp \sqrt{d P_{T}^{\gamma}(\vec{\cmp})\sumtT\ell_{t}(\cmp_{t})}\Big)$, so that the algorithm will automatically perform better on ``easy'' data where the comparator has low loss.
  \item We provide a matching lower bound of the form
        $R_{T}(\vec{\cmp})\ge \Omega\Big(d\Log{T}\maxOp \sqrt{d TP_{T}^{\gamma}(\vec{\cmp})}\Big)$,
        demonstrating
        optimality of the discounted VAW forecaster.

  \item We show that the discount factors required to obtain the results
        in the first point
        can be learned on-the-fly, leading to algorithms
        that make guarantees matching our lower bound.
        Moreover,
        we show how to extend our approach to
        achieve bounds of a similar form
        over
        \emph{every sub-interval
        $[a,b]\subseteq[1,T]$ simultaneously}.
        These are the first
        strongly-adaptive guarantees have been
        achieved in the absence of all boundedness
        assumptions.
\end{itemize}

\subsection{Related Works}\label{sec:rel}
Despite being a well-studied problem setting, there
are no prior works which approach online linear regression
with sufficient generality to be considered free from prior knowledge.
The closest works to our own are
\citet{vovk2001competitive,azoury2001relative,orabona2015generalized,mayo2022scalefree},
each of which consider the same improper online learning
setting as this work and present
algorithms that can be run in an unbounded
domain (hence requiring no prior knowledge about the comparator)
and without any prior knowledge of the data stream.
Yet these works provide guarantees that only hold for \emph{static} regret---the \emph{dynamic} regret of
the algorithms in these works may be arbitrarily bad.
In this sense, deploying any such algorithm
implicitly requires rather strong prior knowledge: that the
data-generating distribution is not changing over time.

A closely related problem setting which
does account for potential non-stationarity is the classic
\emph{filtering} problem \cite{kalman1960filter,simon2006optimal,kozdoba2019online,hazan2022introduction}.
This problem setting assumes that the $\yt$ are generated from
a dynamical system of a specific form, and seeks to estimate the hidden
state of the system. Thus, these works revolve around
strong structural assumptions about the data-generating
process from the outset.
Similarly, there is a large literature on \emph{adaptive} filtering
which seeks to solve the filtering problem without
\emph{a priori} knowledge of the system \cite{kivinen2006pnorm,hazan2017learning,hazan2018spectral,rashidinejad2020slip,tsiamis2022online,ghai2020noregret},
though these works still implicitly require prior knowledge that
the underlying dynamical system is from some specific class,
as any performance guarantees may otherwise fail to hold.

Alternatively, there are several related problem settings that
one might hope to leverage results from,
but these all inevitably
require additional assumptions of some form
to be applied to the online linear regression
problem.
For instance, many prior works develop algorithms for
general online regression settings that capture linear
regression as a
special case
\cite{orabona2015generalized,luo2016efficient,kotlowski2017scale,kempka2019adaptive,mhammedi2020lipschitz}.
Even more generally,
one might hope to approach online linear regression via reduction to a
more general online convex optimization setting
\citep{zhang2018adaptive,yuan2019trading,zhao2020dynamic,baby2021dynamic,baby2021optimal,
 luo2022corralling, jacobsen2022parameter,
zhang2023unconstrained,zhao2024adaptivity}. Unfortunately, all of these works require additional
boundedness
assumptions on the losses such as Lipschitzness or exp-concavity,
both of which require a bounded domain in the context of losses $\ell_{t}(\w)=\half(\yt-\inner{\xt,\w})^{2}$.
Yet assuming a bounded domain amounts
amounts to having strong prior knowledge that the
comparator sequence $\vec{\cmp}=(\cmp_{1},\ldots,\cmp_{T})$ lies entirely within
some bounded subset $W\subset\R^{d}$, which must be known and accounted for
\emph{a priori}
for the guarantees to hold.

One recent exception to the limitations mentioned above is
the work of \citet{jacobsen2023unconstrained}. They develop
an approach that
can be applied to
any loss functions satisfying $\norm{\grad\ell_{t}(\w)}\le G_{t}+L_{t}\norm{\w}$ for
some non-negative constants $G_{t}$ and $L_{t}$, and hence could be applied in our setting
for $G_{t}=\abs{\yt}\norm{\xt}$ and $L_{t}=\norm{\xt}^{2}$.
Their algorithm achieves a dynamic regret
guarantee on the order of $O(M^{3/2}\sqrt{P_{T}T})$ where
$M=\max_{t}\norm{\cmp_{t}}$
and $P_{T}=\sum_{t=2}^{T}\norm{\cmp_{t}-\cmp_{\tmm}}$. However, their
approach fails to achieve logarithmic regret against
a fixed comparator,
and their approach requires prior knowledge of
a $G_{\max}\ge G_{t}$ and $L_{\max}\ge L_{t}$ for all $t$.
Moreover their approach requires
$O(dT\Log{T})$ per-round computation, making it inappropriate
for many of the long-running problems where non-stationarity
naturally emerges due to subtle changes in the environment over time.

\subsection{Notations}
\begin{minipage}{\columnwidth}
We define
$\ell_{0}(\w)=\frac{\lambda}{2}\norm{\w}^{2}_{2}$, so that
updates can be written purely in terms of losses $\ell_{t}$.
Given a positive definite matrix $M$, the weighted norm \wrt{} $M$ is
$\norm{\w}_{M}=\sqrt{\inner{w,Mw}}$.
For any sequence $a_{1},a_{2},\ldots$, we denote $a_{\max}=\max_{t}\abs{a_{t}}$.
Positive thresholding is denoted
as $[\cdot]_{+}=\Max{\cdot,0}$.
The Bregman
divergence \textit{w.r.t.} a differentiable function $\psi$
is
$D_{\psi}(x|y)=\psi(x)-\psi(y)-\inner{\grad\psi(y),x-y}$.
We denote $a\maxOp b = \Max{a,b}$ and $a\minOp b=\Min{a,b}$,
$[N]=\Set{1,\ldots,N}$, $\N=\Set{0,1,\ldots}$ denotes the natural numbers, and $\mathbf{1}_{N}$ is the
$N$-dimensional vector of ones.
We use the short-hand $\Clip_{[a,b]}(\y)=(\y\maxOp a)\minOp b$
and the compressed sum notations $g_{i:j}=\sum_{t=i}^{j}g_{t}$ and $\norm{g}_{a:b}^{2}=\sum_{t=a}^{b}\norm{g_{t}}^{2}$.
The $N$-dimensional simplex is denoted $\Delta_{N}$.
$O(\cdot)$ hides constant factors and $\hat O(\cdot)$ hides constant and $\log\log$ factors.
\end{minipage}

\section{\SecVAW}%
\label{sec:vaw}

In the context of \emph{static} regret,
it is well known that the optimal strategy in
our improper online linear regression setting
is the
Vovk-Azoury-Warmuth (VAW) forecaster, discovered independently by
\citet{azoury2001relative} and \citet{vovk2001competitive}. On each round, the standard
VAW forecaster sets
\begin{align}
  \wt &= \brac{\lambda I + \sum_{s=1}^{t}\x_{s}\x_{s}^{\top}}^{\inv}\sum_{s=1}^{\tmm}\y_{s}\x_{s}.\label{eq:vaw:closed-form}
\end{align}

The VAW forecaster is well-known for the following
regret guarantee \cite{azoury2001relative,vovk2001competitive,orabona2015generalized}.
\begin{restatable}{theorem}{StaticVAW}\label{thm:static-vaw}
  For any $\cmp\in\R^{d}$ and any sequences $(\yt)_{t=1}^{T}$ in $\R$ and $(\x_{t})_{t=1}^{T}$ in $\R^{d}$,
  the VAW forecaster
  guarantees
  \begin{align*}
    R_{T}(\cmp)\le \frac{\lambda}{2}\norm{\cmp}^{2}_{2}+ \frac{d\max_{t}\yt^{2} }{2}\Log{1+\frac{\sumtT\norm{\xt}^{2}_{2}}{\lambda d}},
  \end{align*}
\end{restatable}
Let us briefly pause to appreciate some of the subtleties of
this result, as it represents a very high standard of excellence in
online learning.
First, note that the result holds using \emph{no prior knowledge about the data}
---
there are no underlying assumptions about how the features
$\xt$ or the targets $\yt$ are distributed, the algorithm requires no
specific statistics or bounds such as $\abs{\yt}\le Y$ or $\norm{\xt}\le X$, and
the algorithm works in an unbounded domain --- a relative rarity in
adversarial settings.
Yet despite this incredible degree of generality,
the VAW forecaster boasts a strong
\emph{logarithmic} regret guarantee, which
can be shown to be optimal up to constant
factors (See, \eg{}, \citet[Theorem 11.9]{cesa2006prediction}).
Thus, the VAW forecaster achieves a harmony between
theory and practice
which is quite rare in online learning,
requiring no problem-specific information or assumptions
while still guaranteeing optimal regret.

However, a major caveat to the above discussion 
is that these favorable properties hold only within the 
context of \emph{static} regret.
The \emph{dynamic} regret of the VAW forecaster can be arbitrarily bad
in general. To see why, let us consider the simple case where $d=1$
and $\x_{t}=1$ for all $t$. In this case, the VAW forecaster
predicts
$\yhat_{t}=\xt\wt=(\lambda + t)^{\inv}\sum_{s=1}^{t-1}\y_{s}$, which
approximates an empirical average of the targets observed up to
round $t$. It is easy to see that any such prediction strategy
can fail when competing with a time-varying comparator.
For instance, if the first $T/2$ targets are $-1$ but the second half
are $+1$, the VAW forecaster will quickly converge to predicting $-1$
in the first $T/2$ rounds, but will be unable to quickly adapt
after the
change in the latter $T/2$ rounds, leading to linear regret overall.
In this sense, the VAW forecaster actually \emph{implicitly requires
  quite strong prior knowledge} about the data: that it is, in
some sense, \emph{stationary}. Because of this,
its predictions can \emph{not} be trusted in the
absence of prior knowledge, but rather only
when the practitioner knows
they are dealing with data that can be reasonably predicted using only a
single fixed hypothesis $\cmp\in\R^{d}$.
In the next section, we will see that this issue can
be alleviated by incorporating a suitable recency bias to
the statistics of the VAW forecaster.

\section{\SecDiscountedVAW}
\label{sec:discounted-vaw}

Despite
making strong static regret guarantees,
we saw in the previous section that the standard VAW forecaster
may fail to attain low regret when competing against a
time-varying comparator.
Loosely speaking, the problem is that
the VAW forecaster treats all time-steps as equally important.
Indeed, it can be shown that VAW forecaster
can be understood as updating
\begin{align*}
  \wt=\argmin_{\w\in\R^{d}}\half\norm{\w}^{2}_{\Lambda_{t}}+ \sum_{s=1}^{t-1}\ell_{s}(\w),
\end{align*}
where $\Lambda_{t}=\lambda I + \xt\xt^{\top}$.\footnote{The equivalence to
  \Cref{eq:vaw:closed-form} is readily checked via the first-order
  optimality condition, though this claim can also be derived as a special case
  of a more general claim \Cref{prop:discounted-vaw-update} provided in the appendix.}
The latter term $\sum_{s=1}^{\tmm}\ell_{s}(\w)$
forces the VAW forecaster to choose a $\w$ which balances
all of the losses encountered so-far.
Yet in dynamic scenarios, the losses that contain the most-relevant information
for predicting $\yt$ are typically the ones that have been observed
the most recently.
In order to more closely track these recently-observed losses,
we make two modifications to the VAW forecaster. First, we
incorporate a \emph{forgetting} or \emph{discount}
factor $\gamma$ in to the algorithm's statistics, placing
less emphasis on losses observed far in the past. Second, we allow the update
to additionally make use of a sequence of ``predicted labels'' or ``hints''  $\yhint_t$ that are available before we commit to $\yhat_t$. Intuitively, we would like our algorithm to do better when $\yhint_t=\yt$. Later, we will provide some concrete ways to set $\yhint_t$ that yield strong regret bounds.

The variant of the VAW forecaster described above is provided concretely in
\Cref{alg:discounted-vaw}.
Observe that
by unrolling the recursions for $\theta_{t}$ and $\Sigma_{t}$, the
update $\wt= \Sigma_{t}^{\inv}\sbrac{\yhint_{t}\xt+\gamma\theta_{t}}$
can be written in closed-form as
\begin{align*}
  \wt
  &=\brac{\gamma^{t}\lambda I + \sum_{s=1}^{t}\gamma^{t-s}\x_{s}\x_{s}^{\top} }^{\inv}\sbrac{\yhint_{t}\xt+\gamma\sum_{s=1}^{\tmm}\gamma^{\tmm-s}\y_{s}\x_{s}}.%
\end{align*}
By setting $\gamma=1$ and $\ytilde_{t}=0$, the update precisely reduces to
\Cref{eq:vaw:closed-form}, so the discounted VAW forecaster is a strict
generalization of the standard VAW forecaster.
Likewise, the following
theorem shows that \Cref{alg:discounted-vaw} obtains a regret
guarantee which captures \Cref{thm:static-vaw} as a special case. Proof can be
found in \Cref{app:general-discounted-vaw}.
\begin{restatable}{theorem}{GeneralDiscountedVAW}\label{thm:general-discounted-vaw}
    Let $\lambda>0$ and $\gamma\in(0,1]$. Then
    for any sequence $\vec{\cmp}=(\cmp_{1},\ldots,\cmp_{T})$ in $\R^{d}$,
    \Cref{alg:discounted-vaw} guarantees dynamic regret $R_{T}(\vec{\cmp})$
    bounded above by
    \begin{align*}
      &\frac{\gamma\lambda}{2}\norm{\cmp_{1}}^{2}_{2}+\frac{d}{2}\max_{t}(\yt-\yhint_{t})^{2}\Log{1+\frac{\sumtT \gamma^{T-t}\norm{\x_{t}}^{2}_{2}}{\lambda d}}\\
    &+\gamma\sum_{t=1}^{T-1}\sbrac{F_{t}^{\gamma}(\cmp_{\tpp})-F_{t}^{\gamma}(\cmp_{t})}
    +\frac{d}{2}\Log{1/\gamma}\sumtT(\yt-\yhint_{t})^{2}
    \end{align*}
    where
    $F_{t}^{\gamma}(\w)=\gamma^{t}\frac{\lambda}{2}\norm{\w}^{2}_{2}+\sum_{s=1}^{t}\gamma^{t-s}\ell_{s}(\w)$.
\end{restatable}
\begin{algorithm}[t]
  \SetAlgoLined
  \KwInput{$\lambda>0$, $\gamma\in (0,1]$}\\
  \KwInitialize{$\w_{1}=\zeros$, $\Sigma_{0}=\lambda I$, $\theta_{1}=\zeros$}\\
  \For{$t=1:T$}{
    Receive features $\xt\in\R^{d}$\\
    Set $\Sigma_{t}=\xt\xt^{\top}+\gamma \Sigma_{\tmm}$,
    choose $\yhint_{t}\in\R$\\
    Update
    \(
      \wt= \Sigma_{t}^{\inv}\sbrac{\yhint_{t}\xt+\gamma \theta_{t}}
    \)\\

    \hfill\\
    Predict $\inner{\xt,\wt}$ and observe $\yt$\\
    Incur loss $\ell_{t}(\wt)=\half(\yt-\inner{\xt,\wt})^{2}$\\
    Set $\theta_{\tpp}=\yt\xt + \gamma \theta_{t}$
  }
  \caption{Discounted VAW Forecaster}
  \label{alg:discounted-vaw}
\end{algorithm}
The regret decomposition obtained in \Cref{thm:general-discounted-vaw}
is appealing for two reasons. First,
it captures \Cref{thm:static-vaw} as a special case:
setting $\gamma=1$, $\yhint_{t}=0$, and $\cmp_{1}=\ldots=\cmp_{T}=\cmp$, the last two terms of
the bound evaluate to zero, so the regret is bounded by
\(
  \frac{\lambda}{2}\norm{\cmp}^{2}_{2}+\frac{d}{2}\max_{t}\yt^{2}\Log{1+\frac{\sumtT\norm{\xt}_{2}^{2}}{\lambda d}},
\)
which is precisely the guarantee promised by \Cref{thm:static-vaw}.
Second, the decomposition displays a clean separation
of concerns.
The terms
in the first line are the unavoidable penalties associated with \emph{static}
regret, which are of course also unavoidable here in the more general dynamic regret
setting. In the second line,
any penalties incurred as a result of a changing comparator sequence are
captured entirely by the \emph{variability term}
$\gamma\sum_{t=1}^{T}F_{t}^{\gamma}(\cmp_{\tpp})-F_{t}^{\gamma}(\cmp_{t})$,
while the
term $d\Log{1/\gamma}\sumtT\half (\yt-\yhint_{t})^{2}$ represents a
\emph{stability penalty} incurred due to discounting.

Intuitively, the terms in the second line represent a tracking/stability
trade-off: against a volatile comparator sequence, we
would ideally like to set the discount factor $\gamma$ to be
small to control the variability penalty,
yet this will come at the expense of
increasing the stability penalty $d\Log{1/\gamma}\sumtT\half(\yt-\yhint_{t})^{2}$.
In its current form, however,
this trade-off is still a bit mysterious.
The variability term
$\gamma\sum_{t=1}^{T-1}F_{t}^{\gamma}(\cmp_{\tpp})-F_{t}^{\gamma}(\cmp_{t})$
is not necessarily monotonic
as a function of $\gamma$ nor is it necessarily positive, making it difficult
to meaningfully analyze or understand how it relates to the
stability penalty $\frac{d}{2}\Log{1/\gamma}\sumtT(\yt-\yhint_{t})^{2}$.
If we instead consider a modest upper bound on these
terms we can reveal a more explicit trade-off. We provide proof of a
slightly more general statement of the following lemma in \Cref{app:discounted-terms}.

\begin{restatable}{lemma}{DiscountedTerms}\label{lemma:discounted-terms}
  (simplified)
  Let
  $\ell_{0},\ell_{1},\ldots,\ell_{T}$ be arbitrary non-negative functions,
  $\gamma\in(0,1)$,
  and
  $F_{t}^{\gamma}(\w)=\sum_{s=0}^{t}\gamma^{t-s}\ell_{s}(\w)$.
  For all $t$, define
  \begin{align*}
    \bar d_{t}^{\gamma}(u,v)&=\sum_{s=0}^{t}\frac{\gamma^{t-s}}{\sum_{s'=0}^{t}\gamma^{t-s'}}\sbrac{\ell_{s}(u)-\ell_{s}(v)}_{+}
  \end{align*}
  and $P_{T}^{\gamma}(\vec{\cmp})=\sum_{t=1}^{T-1}\bar d_{t}^{\gamma}(\cmp_{\tpp},\cmp_{t})$.
  Then for any $V_{T}\ge 0$,
  \begin{align*}
    &\gamma \sum_{t=1}^{T-1} \sbrac{F_{t}^{\gamma}(\cmp_{\tpp})-F_{t}^{\gamma}(\cmp_{t})}+\Log{\frac{1}{\gamma}}V_{T}
    \\
      &\qquad\qquad\le\frac{\gamma}{1-\gamma} P_{T}^{\gamma}(\vec{\cmp})+\frac{1-\gamma}{\gamma}V_{T}
  \end{align*}
\end{restatable}
The lemma bounds the variability term
$\gamma\sum_{t=1}^{T-1}\sbrac{F_{t}^{\gamma}(\cmp_{\tpp})-F_{t}^{\gamma}(\cmp_{t})}$ from \Cref{thm:general-discounted-vaw}
in terms of a new one $P_{T}^{\gamma}(\vec{\cmp})$.
To understand this new measure of variability,
for each $t$ let us first define a $\gamma$-exponentially-decaying distribution over
time-steps $s\le t$ as
$p_{t}^{\gamma}(s)=\frac{\gamma^{t-s}}{\sum_{s'=0}^{t}\gamma^{t-s'}}$.
Then, given $\gamma$
we can express $P_{T}^{\gamma}(\vec{u})$ as
\begin{align*}
  P_{T}^{\gamma}(\vec{u})
  &=
    \sum_{t=1}^{T-1}\overbrace{\sum_{s=0}^{t}p_{t}^{\gamma}(s)[\ell_{s}(u_{t+1})-\ell_{s}(u_{t})]_{+}}^{ \bar d_{t}^{\gamma}(u_{t+1},u_{t})}\\
  &=
    \sum_{t=1}^{T-1}\mathbb{E}_{s\sim p_{t}^{\gamma}}\Big[(\ell_{s}(u_{t+1})-\ell_{s}(u_{t}))_{+}\Big],
\end{align*}
so %
each term of $P_{T}^{\gamma}(\vec{\cmp})$
is a measure of
how different the prediction errors of $u_{t}$ and $u_{t+1}$ are on average across ``recent'' losses.
The quantity $P_{T}^{\gamma}(\vec{\cmp})$ can also be naively related to
the more common measure of variability --- the path-length
$P_{T}^{\|\cdot\|}=\sum_{t=1}^{T-1}\|u_{t}-u_{t+1}\|$ --- as follows:
\begin{align*}
  P_{T}^{\gamma}(\vec{\cmp})
  &\le
    \sum_{t=1}^{T-1}\max_{s}\|\nabla \ell_{s}(u_{t+1})\|\|u_{t}-u_{t+1}\|\\
  &\le
    \max_{t,s}\|\nabla\ell_{s}(u_{t})\|P_{T}^{\|\cdot\|}\le O\brac{\max_{t}\norm{\cmp_{t}}P_{T}^{\|\cdot\|}}.
\end{align*}
Thus, $P_{T}^{\gamma}(\vec{\cmp})$ is proportional to the usual path-length.
Note that a multiplicative penalty of $\max_{t}\norm{\cmp_{t}}$
is the same worst-case
penalty that appears in prior works,
even in bounded domains \citep{zhang2018adaptive,jacobsen2022parameter,zhang2023unconstrained,zhao2024adaptivity}.

Letting $\eta=\frac{\gamma}{1-\gamma}$, \Cref{lemma:discounted-terms} tells us that that latter terms of \Cref{thm:general-discounted-vaw}
are bounded by
\[\eta  P_{T}^{\gamma}(\vec{\cmp}) + \frac{d}{2\eta}\sumtT(\yt-\yhint_{t})^{2},\]
a trade-off which can be
optimized by choosing
$\eta=\sqrt{\frac{\frac{d}{2}\sumtT(\yt-\yhint_{t})^{2}}{P_{T}^{\gamma}(\vec{\cmp})}}$ to get
\vspace{-0.5em}
\begin{align*}
 \eta P_{T}^{\gamma}(\vec{\cmp}) +\frac{d}{2\eta}\sumtT(\yt-\yhint_{t})^{2}
  &=
    2\sqrt{d P_{T}^{\gamma}(\vec{\cmp})\sumtT \half(\yt-\yhint_{t})^{2}}.
\end{align*}
This is very promising; as we will see in \Cref{sec:lb}, a penalty of this
form is unavoidable in general.
Plugging this choice of $\eta$ back into
$\eta=\frac{\gamma}{1-\gamma}$ and solving for $\gamma$, we find that
the ideal choice of discount factor would be a $\gamma\in[0,1]$ satisfying
\vspace{-1.25em}
\begin{align*}
  \gamma=\frac{\sqrt{\frac{d}{2}\sumtT(\yt-\yhint_{t})^{2}}}{\sqrt{\frac{d}{2}\sumtT(\yt-\yhint_{t})^{2}}+\sqrt{P_{T}^{\gamma}(\vec{\cmp})}}.
\end{align*}
Notice in particular that $\gamma$ appears on both sides of the expression, and
solving for this $\gamma$ explicitly is non-trivial in general. Nonetheless,
the following theorem shows that a discount factor satisfying the above
expression always exists, and if it could
somehow be provided to the discounted VAW forecaster we would achieve
dynamic regret matching the lower bound in \Cref{sec:lb}.
Proof can be found in \Cref{app:oracle-tuning-discounted-vaw}.
\begin{restatable}{theorem}{OracleTuningDiscountedVAW}\label{thm:oracle-tuning-discounted-vaw}
  For any sequences $\y_{1},\ldots,\y_{T}$ and  $\yhint_{1},\ldots,\yhint_{T}$ in
  $\R$ and any sequence $\vec{\cmp}=(\cmp_{1},\ldots,\cmp_{T})$ in $\R^{d}$,
  there is a discount factor $\gamma^{*}\in[0,1]$ satisfying
  \begin{align}
    \gamma^{*}=\frac{\sqrt{\frac{d}{2}\sumtT(\yt-\yhint_{t})^{2}}}{\sqrt{\frac{d}{2}\sumtT(\yt-\yhint_{t})^{2}}+\sqrt{ P_{T}^{\gamma^{*}}(\vec{\cmp})}}\label{eq:discounted-vaw:optimal-discount}
  \end{align}
  with which the regret of
  \Cref{alg:discounted-vaw} is bounded above by
  \begin{align*}
    R_{T}(\vec{\cmp})
    &\le
    O\bigg(d\max_{t}(\yt-\yhint_{t})^{2}\Log{T}\\
    &\qquad+ \sqrt{d P_{T}^{\gamma^{*}}(\vec{\cmp})\sumtT(\yt-\yhint_{t})^{2}}\bigg)
  \end{align*}
\end{restatable}
While this result is promising, it is important to note that
it still falls short of our desired goal of prior-knowledge-free
learning. Indeed, it seems that we require
\emph{exceptionally strong} prior knowledge to choose the
prescribed discount factor $\gamma^{*}$ satisfying
\Cref{eq:discounted-vaw:optimal-discount}.
We will return to this issue in
\Cref{sec:tuning} to show that this discount factor can be learned
on-the-fly, resulting in algorithms that \emph{are} truly free
of prior knowledge.

Interestingly, the discount factor $\gamma^{*}$ in
\Cref{thm:oracle-tuning-discounted-vaw} can help to
shed some light on the variability measure
$P_{T}^{\gamma^{*}}(\vec{\cmp})$.
Observe from the relation in \Cref{eq:discounted-vaw:optimal-discount}
that $\gamma^{*}$ can be near zero only when $P_{T}^{\gamma^{*}}(\vec{\cmp})$ is
very large relative to the stability penalty, and likewise, if $\gamma^{*}$
is near 1 then $P_{T}^{\gamma^{*}}(\vec{\cmp})$ must be inconsequentially small.
In this sense, the $P_{T}^{\gamma^{*}}(\vec{\cmp})$ corresponding to small
$\gamma^{*}$ can be regarded as the worst-case measures of variability.
Yet as $\gamma^{*}$ approaches zero,
$P_{T}^{\gamma^{*}}(\vec{\cmp})$ approaches $\sum_{t=1}^{t-1}\sbrac{\ell_{t}(\cmp_{\tpp})-\ell_{t}(\cmp_{t})}_{+}$,
which can be naturally related other standard measures of variability. Indeed, this
penalty is similar in spirit to the
temporal variability
$\sum_{t=1}^{T-1}\abs{\ell_{\tpp}(\cmp_{t})-\ell_{t}(\cmp_{t})}$ studied in works such as
\citet{campolongo2021closer,besbes2015nonstationary},
and can be related to the path-length
$\sum_{t=1}^{T-1} \norm{\cmp_{t}-\cmp_{\tpp}}$ via convexity of $\ell_{t}$.
In this sense, $P_{T}^{\gamma^{*}}(\vec{\cmp})$ can be thought of as a relaxation
of the more common measures of variability.

\subsection{\SecSmallLoss}%
\label{sec:small-loss}

In the previous section, we saw that the discounted VAW forecaster
can achieve regret scaling as
$O\brac{\sqrt{d P_{T}^{\gamma^{*}}(\vec{\cmp})\sumtT(\yt-\yhint_{t})^{2}}}$,
where $\yhint_{t}\in\R$ is an arbitrary ``hint'' available
before observing the true $\yt$.
One particularly interesting option is to use \emph{the learner's own
  prediction} as a hint, $\yhint_{t}=\inner{\xt,\wt}$.
The reasoning is that any learner achieving low dynamic regret
must be predicting $\yt$ reasonably well on average,
so their own predictions would naturally make
for
reasonable predicted labels $\yhint_{t}$.
Concretely,  observe that by choosing
$\yhint_{t}=\inner{\xt,\wt}$ we would have
$\sumtT(\yt-\yhint_{t})^{2}=\sumtT (\yt-\inner{\xt,\wt})^{2}=2\sumtT \ell_{t}(\wt)$,
and hence for some $\gamma\in[0,1]$ the guarantee in
\Cref{thm:oracle-tuning-discounted-vaw} would scale as
\begin{align*}
  R_{T}(\vec{\cmp})
  &=
    \sumtT \ell_{t}(\wt)-\ell_{t}(\cmp_{t})
  \le
     \tilde O\brac{\sqrt{d P_{T}^{\gamma}(\vec{\cmp})\sumtT \ell_{t}(\wt)}},
\end{align*}
where the $\tilde O(\cdot)$ hides the logarithmic factor.
Now notice that $\sumtT \ell_{t}(\wt)$ appears on both sides of this inequality.
Solving for
$\sumtT\ell_{t}(\wt)$, one finds that this implies that
$\sqrt{\sumtT\ell_{t}(\wt)}\le O\brac{\sqrt{d P_{T}^{\gamma}(\vec{\cmp})}+\sqrt{\sumtT\ell_{t}(\cmp_{t})}}$,
so plugging this back into the regret bound we have
\begin{align*}
  R_{T}(\vec{\cmp})\le \tilde O\brac{P_{T}^{\gamma}(\vec{\cmp})+\sqrt{P_{T}^{\gamma}(\vec{\cmp})\sumtT\ell_{t}(\cmp_{t})}}.
\end{align*}
Bounds of this form, sometimes called \emph{small-loss} or $L^{*}$ bounds,
are highly desirable because they naturally adapt to the total loss
of the comparator sequence, potentially leading to lower regret than
more naive hint choices such as $\yhint_{t}=\ytmm$ or $\yhint_{t}=0$.

Unfortunately, the above argument does not quite go through because the
now the logarithmic penalty in \Cref{thm:oracle-tuning-discounted-vaw} scales as
$O\brac{d\max_{t}(\yt-\yhint_{t})^{2}\Log{T}}=O\brac{d\max_{t}\ell_{t}(\wt)\Log{T}}$,
and this $\max_{t}\ell_{t}(\wt)$ could be arbitrarily large. Fortunately,
it turns out that this issue can be remedied by a simple
trust-region argument. On each round, instead of
directly using hints $\yhint_{t}=\inner{\xt,\wt}$, we can constrain these
predictions to be close to some arbitrary reference point
$\yref_{t}$. In particular, in \Cref{lemma:clipped-predictions}
we show %
by clipping the learner's predictions to a suitable interval centered
at $\yref_{t}$ we can guarantee
$(\yt-\yhint_{t})^{2}\le O\brac{\max_{t}(\yt-\yref_{t})^{2}\minOp \ell_{t}(\wt)}$.
This gives us the best-of-both-worlds: a similar self-bounding argument
to above still yields a small-loss penalty
$O\Big(\sqrt{d P_{T}^{\gamma}(\vec{\cmp})\sumtT\ell_{t}(\cmp_{t})}\Big)$, while the
logarithmic penalty can be bounded as
$O\brac{d\max_{t}(\yt-\yref_{t})^{2}\Log{T}}\le O(d\max_{t}\yt^{2}\Log{T})$
by setting $\yref_{t}=\ytmm$ or $\yref_{t}=0$.
The following theorem follows this above argument through,
demonstrating that the discounted VAW forecaster can achieve
small-loss bounds when using a well-chosen discount factor.
\begin{restatable}{theorem}{OracleTuningSmallLoss}\label{thm:oracle-tuning-small-loss}
  Let $\yref_{t}\in\R$ be an arbitrary reference
  point and let
  $\cB_{t}=[\yref_{t}-M_{t},\yref_{t}+M_{t}]$ for
  $M_{t}=\max_{s<t}\abs{\y_{s}-\yref_{s}}$.
  Suppose that we apply \Cref{alg:discounted-vaw} with
  hints $\yhint_{t}=\Clip_{\cB_{t}}(\inner{\xt,\wt})$.
  Then for any sequence of losses $\ell_{1},\ldots,\ell_{T}$ and any
  sequence $\vec{\cmp}=(\cmp_{1},\ldots,\cmp_{T})$
  in $\R^{d}$, there is a $\gamma^{\Ring}\in[0,1]$ satisfying
  \begin{align}
    \gamma^{\Ring}=\frac{\sqrt{d\sumtT\ell_{t}(\cmp_{t})}}{\sqrt{d\sumtT\ell_{t}(\cmp_{t})}+\sqrt{
      P_{T}^{\gamma^{\Ring}}(\vec{\cmp})}}.\label{eq:small-loss:optimal-discount}
  \end{align}
  Moreover, running \Cref{alg:discounted-vaw} with discount
  $\gamma^{\Ring}\maxOp\gamma_{\min}$
  for $\gamma_{\min}=\frac{2d}{2d+1}$ ensures regret bounded above by
  \begin{align*}
    R_{T}(\vec{\cmp})
    &\le
      O\bigg(d P_{T}^{\gamma_{\min}}(\vec{\cmp})+ d\max_{t}(\yt-\yref_{t})^{2}\Log{T}\\
    &\qquad
      + \sqrt{d P_{T}^{\gamma^{\Ring}}(\vec{\cmp})\sumtT\ell_{t}(\cmp_{t})}\bigg),
  \end{align*}
\end{restatable}

Notice that unlike the previous section,
there are two different variability penalties,
$P_{T}^{\gamma^{\Ring}}(\vec{\cmp})$ and $P_{T}^{\gamma_{\min}}(\vec{\cmp})$.
The first mirrors the measure encountered in the last section.
The other, $P_{T}^{\gamma_{\min}}(\vec{\cmp})$,
is rather annoying; in high dimensions $\gamma_{\min}=\frac{2d}{2d+1}$ is
generally quite large, so $P_{T}^{\gamma_{\min}}(\vec{\cmp})$ may evaluate
losses at irrelevant comparators that are far away in time.
Nevertheless, notice that this term satisfies
$P_{T}^{\gamma_{\min}}(\vec{\cmp})
\le \sum_{t=1}^{T-1}\max_{s}\sbrac{\ell_{s}(\cmp_{\tpp})-\ell_{s}(\cmp_{t})}_{+}$,
a penalty which we will show is unavoidable in general
in \Cref{thm:tight-lb}.

\subsection{\SecLB}%
\label{sec:lb}

In this section, we show that the regret penalties observed
in the previous sections are unavoidable without further assumptions.
The following lower bound is proven in \Cref{app:tight-lb}.
\begin{minipage}{\columnwidth}
\begin{restatable}{theorem}{TightLB}\label{thm:tight-lb}
  For any $d,T\ge 1$ and $P,Y>0$ such that $dP\le 2TY^{2}$, there is a sequence of
  losses $\ell_{t}(\w)=\half(\yt-\inner{\xt,\w})^{2}$ and
  a comparator sequence $\vec{\cmp}=(\cmp_{1},\ldots,\cmp_{T})$
  satisfying
  $\max_{t}\abs{\yt}\le Y$ and
  $\sum_{t=1}^{T-1}\max_{s}\sbrac{\ell_{s}(\cmp_{\tpp})-\ell_{s}(\cmp_{t})}_{+}\le P$
  such that
  \begin{align*}
    R_{T}(\vec{\cmp}) \ge \Omega\brac{ dY^{2}\Log{T}+dP+ \sqrt{dP\sum_{t=2}^{T}(\yt-\ytmm)^{2}}}\ .
  \end{align*}
\end{restatable}
\end{minipage}

The key observation is that there is always a sequence of losses such that
$\sumtT\ell_{t}(\cmp_{t})=0$ can be ensured
using only $T/d$ different comparators. Indeed,
letting the features $\xt$ cycle through the standard basis vectors,
for any sub-interval $[s,s+d]\subseteq[1,T]$ we can choose a
single $\cmp\in\R^{d}$ such that $\inner{\xt,\cmp}=y_{t}$ for each $t$
in the interval. Then by sampling the $\yt$ randomly from $\Set{-Y\sigma,Y\sigma}$
for some $\sigma\in[0,1]$, we can ensure variability of at most
$O(TY^{2}\sigma^{2}/d)\le P$ but regret of at least
$\Omega(TY^{2}\sigma^{2})\ge\Omega\brac{\sqrt{dP\sbrac{ \sumtT (\yt-\ytmm)^{2}\maxOp dP}}}$.

Note that the condition $dP\le 2TY^{2}$ captures a natural restriction of the
problem setting, in that for larger $P$
the vacuous lower bound $R_{T}(\vec{\cmp})\ge \Omega(TY^{2})$ can be constructed.
Indeed, in the boundary case where $dP=2TY^{2}$,
\Cref{thm:tight-lb} tells us that there is a sequence
such that
$R_{T}(\vec{\cmp})\ge \Omega\brac{\sqrt{dP\cV_{T}}}=\Omega\brac{ dP}=\Omega\brac{TY^{2}}$.
Yet this bound is achieved against \emph{any} comparator sequence by the algorithm that naively
predicts $\zeros$ on every round:
$R_{T}(\vec{\cmp})=\sumtT\ell_{t}(\zeros)-\ell_{t}(\cmp_{t})\le \sumtT\half\yt^{2}\le\half TY^{2}$.
Hence, no lower bound can exceed $\half TY^{2}$, so it is sufficient to
consider comparator sequences with variability bounded by $P\le 2TY^{2}$.

If we instead consider a more restricted problem setting
by assuming a bounded domain, then the losses
$\ell_{t}(\w)=\half(\yt-\inner{\xt,\w})^{2}$ can be
considered to be exp-concave.
In this setting,
\citet{baby2021optimal} have shown a lower bound of
\begin{align}
R_{T}(\vec{u})\ge \Omega \left(Y^{4/3}d^{1/3}T^{1/3}C_{T}^{2/3}\right),\label{eq:baby-lb}
\end{align}
where $C_{T}=\sum_{t=1}^{T-1}\norm{\cmp_{t}-\cmp_{\tmm}}_{1}$.
A natural question is whether similar results
also hold in the unbounded setting, and how they
compare to our lower bound in \Cref{thm:tight-lb}.
Note that even in the exp-concave
setting, the bound in \Cref{eq:baby-lb} is not necessarily tight. Indeed,
\citet{baby2021optimal} provide
an algorithm which
guarantees
\begin{align*}
  R_{T}(\vec{u})\le \tilde O(Y^{4/3}d^{3.5}T^{1/3}C_{T}^{2/3}),
\end{align*}
which does not match the lower bound \wrt{} the dimension $d$.
In contrast, our lower bound in \Cref{thm:tight-lb}
matches our upper bounds in all involved quantities (see \Cref{sec:discounted-vaw,sec:tuning}).
Regardless, we also demonstrate in \Cref{app:baby-comparison}
that the same $\tilde O(Y^{4/3}d^{3.5}T^{1/3}C_{T}^{2/3})$ upper bound can be attained,
even in unbounded domains,
using the strongly-adaptive guarantees
developed in \Cref{sec:strongly-adaptive}.

\section{\SecTuning}%
\label{sec:tuning}

Recall that our goal from the outset has been to design algorithms
that achieve favourable dynamic regret guarantees using
\emph{no prior knowledge}. To this end,
we showed in \Cref{sec:discounted-vaw} that the discounted VAW forecaster
can achieve dynamic regret guarantees of the form
$R_{T}(\vec{\cmp})\le O\brac{\sqrt{dP_{T}^{\gamma}(\vec{\cmp})T}\maxOp d\Log{T}}$ where
$P_{T}^{\gamma}(\vec{\cmp})$ is a certain measure of variability of the comparator sequence,
and in \Cref{sec:lb} we showed that these penalties are unavoidable in general.
However, these results hold under the assumption that the
learner chooses discount rates satisfying special conditions (\Cref{eq:discounted-vaw:optimal-discount,eq:small-loss:optimal-discount}),
either of which would require exceptionally strong prior knowlege to ensure. Indeed, the
learner would need to know the future!
In order to achieve our goal of learning in the complete absence of prior knowledge,
we need to ensure that the learner can adequately guess or learn
these ideal discount factors on-the-fly.

A common way to achieve runtime parameter-tuning of this sort would be to run many
instances of the algorithm for different choices of $\gamma$ in parallel, and combine the predictions
using a suitable meta-algorithm. In particular,
suppose we have a collection of
algorithms $\cA_{1},\ldots,\cA_{N}$ and on each round we
can query each $\cA_{i}$ for a prediction $\yt^{(i)}\in\R$. Moreover,
suppose we have a meta-algorithm $\cA_{\Meta}$ which tells us how to
combine these predictions by outputting a $p_{t}$ from the $N$-dimensional simplex $\Delta_{N}$.
Then by predicting $\ypred_{t}=\sum_{i=1}^{N}p_{ti}\yt^{(i)}$,
\footnote{
  Recall from the introduction that because the features $\xt$
  are provided at the start of the round, we can work directly in the output
  space $\R$ if we so choose by setting $\wt=\ypred_{t}\xt/\norm{\xt}^{2}$.
  Hence, given $\ypred\in \R$ we allow a slight abuse of notation by
  letting $\ell_{t}(\ypred)=\half(\yt-\ypred)^{2}$.
}
for any
benchmark sequence
$\vec{\cmp}=(\cmp_{1},\ldots,\cmp_{T})$ and any $j\in[N]$ we
have
\begin{align*}
  R_{T}(\vec{\cmp})
  &=
    \sumtT\ell_{t}(\ypred_{t})-\ell_{t}(\cmp_{t})\\
  &=
    \underbrace{\sumtT \ell_{t}(\yt^{(j)})-\ell_{t}(\cmp_{t})}_{\smash{=:R_{T}^{\cA_{j}}(\vec{\cmp})}}+\underbrace{\sumtT\ell_{t}(\ypred_{t})-\ell_{t}(\yt^{(j)})}_{\smash{=:R_{T}^{\Meta}(e_{j})}}
\end{align*}
where the last line observes that $\yt^{(j)}=\langle\xt,\wt^{(j)}\rangle$.
Hence, we may achieve our goal if we can
ensure 1) that there is a $j\in[N]$ such that
$\cA_{j}$ uses a near-optimal discount factor $\gamma_{j}$,
and 2) we can provide a meta-algorithm
which guarantees low regret $R_{T}^{\Meta}(e_{j})$.
We first investigate the latter point, and
return to the former in \Cref{thm:grid-tuning-discounted-vaw,thm:grid-tuning-small-loss}.

The obvious approach to bounding the meta-algorithm's regret would be
to observe that the losses $\ell_{t}(\ypred_{t})=\half(\yt-\ybar_{t})^{2}$
are $\alpha_{t}$-exp-concave for $\alpha_{t}=\frac{1}{2\max_{i}\ell_{t}(\yt^{(i)})}$
(\Cref{lemma:squared-loss-exp-concave}), which will allow us to
apply an instance of the fixed-share algorithm \cite{cesa2012mirror}
to get:
\begin{align*}
  R_{T}^{\Meta}(e_{j})
  &\le
    O\brac{\frac{\Log{NT}}{\alpha_{T+1}}}
  \le
    O\Big(\mspace{-3mu}\max_{t,i}\ell_{t}(\yt^{(i)})\Log{NT}\mspace{-5mu}\Big),
\end{align*}
as shown in \Cref{thm:adaptive-fixed-share}.
However, just like in \Cref{sec:small-loss}, the term
$\max_{t,i}\ell_{t}(\yt^{(i)})$ is hard to quantify and
could be be arbitrarily large in general. Fortunately the
very same clipping trick used in \Cref{sec:small-loss}
also works here: instead of having the meta-algorithm
combine the \emph{raw} predictions $\yt^{(i)}$, we
can simply clip the predictions to a trust-region
around a given reference point $\yref_{t}$.
In \Cref{lemma:exp-concave-experts-redux} we show that
the clipping strategy detailed in \Cref{alg:clipped-meta}
incurs only an additional constant penalty in the regret.
Then, using \Cref{lemma:clipped-predictions},
using these clipped predictions leads to
\begin{align*}
  R_{T}^{\Meta}(e_{j})
  &\le
    O(\max_{t}(\yt-\yref_{t})^{2}\Log{NT}).
\end{align*}
Note that a penalty of a similar order is already present
in the regret of the VAW forecaster (\eg{} \Cref{thm:general-discounted-vaw})
so this result will be sufficient for our purposes.
Overall, the following theorem formalizes
the argument described above. We provide a simplified statement here for
brevity, but the full statement and its proof can be found in \Cref{app:dynamic-regret-tuning}.
\begin{algorithm}[t]
  \SetAlgoLined
  \KwInput{Online learning algorithms $\cA_{1},\ldots, \cA_{N}$, experts algorithm
  $\cA_{\Meta}$ over the simplex $\Delta_{N}$.}\\
  \KwInitialize{$\cA_{\Meta},\cA_{1},\ldots,\cA_{N}$, and set $M_{1}=0$}\\
  \For{$t=1:T$}{
    Receive features $\xt$\\
    Choose reference point $\yref_{t}$\\
    Define $\cB_{t}=[\yref_{t}-M_{t},\yref_{t}+M_{t}]$\\
    \For{$i=1,\ldots,N$}{
      Send $\xt$ to $\cA_{i}$\\
      Get prediction $\yt^{(i)}=\langle\xt,\wt^{(i)}\rangle$ from $\cA_{i}$\\
      Compute $\yclip_{t}^{(i)}=\Clip_{\cB_{t}}(\yt^{(i)})$\\
    }
    Get $p_{t}\in\Delta_{N}$ from $\cA_{\Meta}$\\
    Predict $\ypred_{t}=\sum_{i=1}^{N}p_{ti}\yclip_{t}^{(i)}$ and observe
    $\yt$\\
    Update $M_{\tpp}=M_{t}\maxOp \abs{\yt-\yref_{t}}$\\
    \hfill\\
    Send $\ell_{t}(\w)=\half(\yt-\inner{\xt,\w})^{2}$ to $\cA_{i}$ $\forall i$\\
    Send $\ell_{t}(\yclip_{t}^{(1)}),\ldots,\ell_{t}(\yclip_{t}^{(N)})$ to
    $\cA_{\Meta}$
  }
  \caption{Range-clipped Meta-algorithm}
  \label{alg:clipped-meta}
\end{algorithm}

\begin{restatable}{theorem}{DynamicRegretTuning}\label{thm:dynamic-regret-tuning}
  (simplified)
  Let $\cA_{\Meta}$ be the instance of fixed-share characterized in \Cref{thm:adaptive-fixed-share}.
  Then
  for any sequence $\vec{\cmp}=\brac{\cmp_{1},\ldots,\cmp_{T}}$ in $\R$ and any $j\in[N]$,
  \Cref{alg:clipped-meta} guarantees
  \begin{align*}
    R_{T}(\vec{\cmp})
    &\le
      \hat O\brac{ R_{T}^{\cA_{j}}(\vec{\cmp})+\max_{t}\brac{\yt-\yref_{t}}^{2}\Log{NT}},
  \end{align*}
  where $\hat O(\cdot)$ hides $\log\log$ terms.
\end{restatable}
A similar target-clipping strategy was recently used by
\citet{mayo2022scalefree} to prove a static regret result for scale-free unconstrained online
regression.
\Cref{thm:dynamic-regret-tuning} generalizes their approach by clipping
to a trust-region of an arbitrary center $\yref_{t}\in\R$,
and offers a somewhat streamlined argument which does not appeal to
probabilistic notions such as mixibility.

Finally, with \Cref{thm:dynamic-regret-tuning} in hand,
we can achieve our desired result by running \Cref{alg:clipped-meta}
with the base algorithms $\cA_{i}$ being instances of the discounted VAW forecaster
with different discount factors $\gamma$.
The following theorems show that for a well-chosen
set of discount factors, we can make guarantees that match
the bounds attained under oracle tuning of $\gamma$
(\Cref{thm:oracle-tuning-discounted-vaw,thm:oracle-tuning-small-loss}),
yet require no prior knowledge of any sort.
Proofs can be found in
\Cref{app:grid-tuning-discounted-vaw,app:grid-tuning-small-loss} respectively.
\begin{restatable}{theorem}{GridTuningDiscountedVAW}\label{thm:grid-tuning-discounted-vaw}
  Let $b>1$, $\eta_{\min}=2d$, $\eta_{\max}=dT$, and for all $i\in\N$ let
  $\eta_{i}=\eta_{\min}b^{i}\minOp\eta_{\max}$, and construct the set of discount factors
  \(
    \cS_{\gamma}=\Set{\gamma_{i}=\frac{\eta_{i}}{1+\eta_{i}}: i\in \N}\cup\Set{0}.
  \)
  For any $\gamma$ in $\cS_{\gamma}$, let $\cA_{\gamma}$ denote an instance of \Cref{alg:discounted-vaw}
  with discount $\gamma$.\footnote{
    For brevity, here we refer to an algorithm that
    directly predicts $\yhint_{t}$ on every round as being an instance of the
    discounted VAW forecaster with $\gamma=0$. This terminology can be justified
    by \Cref{remark:gamma-zero}, but for our purposes here it's sufficient to
    consider it convenient alias.
  } Let $\cA_{\Meta}$ be an instance of the
  algorithm characterized in \Cref{thm:dynamic-regret-tuning}, and suppose we
  set $\yref_{t}=\yhint_{t}$ for all $t$.
  Then for any $\vec{\cmp}=(\cmp_{1},\ldots,\cmp_{T})$ in $\R^{d}$,
  \Cref{alg:clipped-meta} guarantees
  \begin{align*}
    R_{T}(\vec{\cmp})
    &\le
      O\bigg(d\max_{t}(\yt-\yref_{t})^{2}\Log{T}\\
    &\qquad
      + b\sqrt{d P_{T}^{\gamma^{*}}(\vec{\cmp})\sumtT (\yt-\yhint_{t})^{2}}\bigg)
  \end{align*}
  where $\gamma^{*}\in[0,1]$ satisfies \Cref{eq:discounted-vaw:optimal-discount}.
\end{restatable}

\begin{restatable}{theorem}{GridTuningSmallLoss}\label{thm:grid-tuning-small-loss}
  Under the same conditions as \Cref{thm:grid-tuning-discounted-vaw},
  suppose each $\cA_{\gamma}$ sets hints
  $\yhint_{t}=\yclip_{t}^{\gamma}=\Clip_{\cB_{t}}(\inner{,\xt,\wt^{\gamma}})$,
  where $\cB_{t}=[\yref_{t}-M_{t},\yref_{t}+M_{t}]$
  and $M_{t}=\max_{s<t}\abs{\y_{s}-\yref_{s}}$.
  Then for any $\vec{\cmp}=(\cmp_{1},\ldots,\cmp_{T})$ in $\R^{d}$,
  \Cref{alg:clipped-meta} guarantees
  \begin{align*}
    R_{T}(\vec{\cmp})
    &\le
      O\bigg(d P_{T}^{\gamma_{\min}}(\vec{\cmp})+ d\max_{t}\brac{\yt-\yref_{t}}^{2}\Log{T}\\
    &\qquad
      + b\sqrt{d P_{T}^{\gamma^{\Ring}}(\vec{\cmp})\sumtT\ell_{t}(\cmp_{t})}\bigg)
  \end{align*}
  where
  $\gamma_{\min}=\frac{2d}{2d+1}$ and
  $\gamma^{\Ring}\in[0,1]$ satisfies \Cref{eq:small-loss:optimal-discount}.
\end{restatable}

It is worth noting that
\Cref{thm:grid-tuning-discounted-vaw,thm:grid-tuning-small-loss} use knowledge
of the horizon $T$ to construct the set of experts.
All of our results extend
immediately
to the unknown $T$ setting as well via the standard
doubling trick \citep{cesa2006prediction},
so for simplicity we treat $T$ as part of the problem setting rather than a
potentially unknown property of the data.
An interesting direction for future development would be to construct the
set of experts in a more on-the-fly way,
so as to avoid using the doubling trick
to adapt to unknown $T$.

\section{Strongly-Adaptive Guarantees}%
\label{sec:strongly-adaptive}

While our original goal was only to achieve
dynamic regret guarantees in the absence of prior knowledge, it turns out that
we can actually achieve an even stronger result:
dynamic regret guarantees that hold over
\emph{every sub-interal $[a,b]\subseteq[1,T]$ simultaneously}.
To our knowledge, \emph{strongly-adaptive} guarantees of this sort have
previously
only been achieved under various boundedness assumptions
\cite{baby2021dynamic,baby2022optimal,baby2022lqr,jun2017improved,cutkosky2020parameter,daniely2015strongly}.

The results can be derived using the results in the previous section.
As shown in \Cref{app:dynamic-regret-tuning},
for any $[s,\tau]\subseteq[1,T]$, $\vec{\cmp}=(\cmp_{s},\ldots,\cmp_{\tau})$, and
$\gamma\in\cS_{\gamma}$,
\Cref{alg:clipped-meta}
more generally guarantees that
\begin{align*}
  R_{[s,\tau]}(\vec{\cmp})
  &\le
    \hat O\brac{
    R_{[s,\tau]}^{\cA_{\gamma}}(\vec{\cmp})+\max_{t}(\yt-\yref_{t})^{2}\Log{N\tau}
    },
\end{align*}
where $R_{[s,\tau]}$ denotes the regret over sub-interval $[s,\tau]\subseteq[1,T]$.
The only caveat is that the regret guarantees of the
discounted VAW forecaster only hold when the algorithm
\emph{begins learning} on round  $s$.\footnote{
  More generally, it can be seen from the analysis
  that if the algorithm starts at time $t=1$ and we try to bound the regret over
  $[s,\tau]$, then after telescoping the divergence terms
  we will end up with a non-trivial term $D_{\psi_{s}}(\cmp_{s}|\w_{s})$ which is
  hard to quantify in general for $s>1$ without further assumptions.}
However, suppose that for each $s\in[1,T]$ and each $\gamma\in\cS_{\gamma}$ we define
an algorithm $\cA_{\gamma,s}$ which uses discount $\gamma$
but begins learning at time $s$. Then for any $[s,\tau]$
\Cref{lemma:grid-tuning-discounted-vaw-existence} implies that
there is a $\cA_{\gamma,s}$ such that
\(
R_{[s,\tau]}^{\cA_{\gamma,s}}(\vec{\cmp})
  \le
      O(d\max_{t}(\yt-\yref_{t})^{2}\Log{\tau-s}
      + b\sqrt{d P_{[s,\tau]}^{\gamma^{*}}(\vec{\cmp})\sum_{t=s}^{\tau} (\yt-\yhint_{t})^{2}}).
      \)
      Plugging this back into the previous display and choosing
$\abs{\cS_{\gamma}}\le O(\Log{T})$, we have $N\le O(T\Log{T})$ and
an overall regret bound of
\begin{align*}
  R_{[s,\tau]}(\vec{\cmp})
  &\le
    \hat O\Bigg(d\max_{t}(\yt-\yhint_{t})^{2}\Log{T}\\
  &\qquad
      + b\sqrt{d P_{[s,\tau]}^{\gamma^{*}}(\vec{\cmp})\sum_{t=s}^{\tau} (\yt-\yhint_{t})^{2}}\Bigg).
\end{align*}
This is the essence of the Follow the Leading History
algorithm of \citet{hazan2007adaptive,hazan2009efficient}.

While the above approach leads to a strongly-adaptive guarantee,
it would be excessively expensive in general, since we'd now have
$O(T\Log{T})$ total experts to update on every round.
We may instead lower this to $O(\log^{2}(T))$ experts
using the geometric covering intervals of
\citet{daniely2015strongly,veness2013partition}. The idea is as follows:
instead of initializing a new instance of each $\cA_{\gamma}$
on every round $s\in[T]$, we will construct a set
of intervals $S$ such that
any $[s,\tau]\subseteq[1,T]$ can be
covered using only a small number of intervals from $S$.
Then for each $\gamma\in\cS_{\gamma}$ and each $I\in S$, we can
define an instance of the discounted VAW forecaster $\cA_{\gamma,I}$
which is run only during the interval $I$. The geometric covering intervals
are constructed in such a way that
$1)$ any round $t$
can fall into at most $O(\Log{T})$ of the intervals,
and $2)$ any $[s,\tau]\subseteq[1,T]$ can be covered using only
$O(\Log{\tau-s})$ disjoint intervals from $S$.
The first property ensures that there at most $O(\log^{2}(T))$
active experts on each round,
while the second property implies that
there is a disjoint set of intervals $I_{1},\ldots,I_{K}$ such that
$R_{[s,\tau]}(\vec{\cmp})=\sum_{i=1}^{K}R_{I_{i}}(\vec{\cmp})$, so bounding
each of these using a similar argument to the above
followed by an application of Cauchy-Schwarz inequality yields
\begin{align*}
  R_{[s,\tau]}(\vec{\cmp})
  &\le
    \hat O\bigg(d\max_{t}(\yt-\yhint_{t})^{2}\log^{2}(T)\\
  &\qquad
  + b\sqrt{d P_{[s,\tau]}^{\gamma^{*}}(\vec{\cmp})\sum_{t=s}^{\tau}(\yt-\yhint_{t})^{2}}\bigg),
\end{align*}
where $P_{[s,\tau]}(\vec{\cmp})$ is the total variability over the intervals
and we've used $K\Log{T}\le O(\log^{2}(T))$. Hence, overall the penalty
we incur for using the
geometric covering is a modest increase from  $\Log{T}$ to
$K\Log{T}\le O(\log^{2}(T))$ in the leading term. Likewise, a similar argument holds for our
small-loss bounds.
We provide a formal statement and proof of these results in
\Cref{app:strongly-adaptive}.

\section{Conclusion}%
\label{sec:conclusion}

In this paper, we designed
algorithms for online linear regression which
achieve optimal dynamic regret guarantees,
even in the absence of all prior knowledge.
We developed a novel analysis of
a discounted variant of the Vovk-Azoury-Warmuth
forecaster, showing that it can guarantee dynamic regret of the form
$R_{T}(\vec{\cmp})\le O\brac{d\Log{T}\maxOp\sqrt{d P_{T}^{\gamma}(\vec{\cmp})T}}$
when equipped with an appropriate
discount factor (\Cref{sec:discounted-vaw}).
We also provided a matching lower bound, demonstrating that
these penalties
are unavoidable in general (\Cref{sec:lb}). We then
showed that the ideal discount factors can be learned on-the-fly,
resulting in algorithms that
can be applied with no prior knowledge yet
still make optimal dynamic regret guarantees (\Cref{sec:tuning})
and strongly-adaptive guarantees (\Cref{sec:strongly-adaptive}).
These are the
first algorithms for online linear regression that
make meaningful guarantees without making assumptions of any kind
on the underlying data.

An important direction for future work is to reduce the computational
complexity of the algorithms. Similar to the traditional VAW forecaster,
the approach developed here can be infeasible for very high-dimensional
features, requiring roughly $O(d^{2}\Log{T})$ computation every round.
The $d^{2}$ factor likely can be reduced by extending our analysis
to use modern sketching techniques \cite{luo2016efficient},
and the $\Log{T}$ factor can possibly be reduced
using similar techniques to
the recent work of \citet{lu2022efficient}.

\section*{Acknowledgements}
We thank David Janz for valuable feedback on the initial draft of this work. AJ was supported by an NSERC CGS-D scholarship. AC acknowledges support from NSF grant no. 2211718.

\section*{Impact Statement} This paper presents theoretical work that improves online linear regression. We do not anticipate any significant  negative societal consequences.

\bibliographystyle{icml2024}
\bibliography{refs}

\begin{thebibliography}{44}
\providecommand{\natexlab}[1]{#1}
\providecommand{\url}[1]{\texttt{#1}}
\expandafter\ifx\csname urlstyle\endcsname\relax
  \providecommand{\doi}[1]{doi: #1}\else
  \providecommand{\doi}{doi: \begingroup \urlstyle{rm}\Url}\fi

\bibitem[Azoury \& Warmuth(2001)Azoury and Warmuth]{azoury2001relative}
Azoury, K.~S. and Warmuth, M.~K.
\newblock Relative loss bounds for on-line density estimation with the
  exponential family of distributions.
\newblock \emph{Machine learning}, 43:\penalty0 211--246, 2001.

\bibitem[Baby \& Wang(2021)Baby and Wang]{baby2021optimal}
Baby, D. and Wang, Y.-X.
\newblock Optimal dynamic regret in exp-concave online learning.
\newblock In \emph{Proceedings of Thirty Fourth Conference on Learning Theory}.
  PMLR, 2021.

\bibitem[Baby \& Wang(2022{\natexlab{a}})Baby and Wang]{baby2022lqr}
Baby, D. and Wang, Y.-X.
\newblock Optimal dynamic regret in lqr control.
\newblock In \emph{Advances in Neural Information Processing Systems},
  2022{\natexlab{a}}.

\bibitem[Baby \& Wang(2022{\natexlab{b}})Baby and Wang]{baby2022optimal}
Baby, D. and Wang, Y.-X.
\newblock Optimal dynamic regret in proper online learning with strongly convex
  losses and beyond.
\newblock In \emph{Proceedings of The 25th International Conference on
  Artificial Intelligence and Statistics}. PMLR, 2022{\natexlab{b}}.

\bibitem[Baby et~al.(2021)Baby, Hasson, and Wang]{baby2021dynamic}
Baby, D., Hasson, H., and Wang, Y.
\newblock Dynamic regret for strongly adaptive methods and optimality of online
  krr, 2021.

\bibitem[Besbes et~al.(2015)Besbes, Gur, and Zeevi]{besbes2015nonstationary}
Besbes, O., Gur, Y., and Zeevi, A.
\newblock Non-stationary stochastic optimization.
\newblock \emph{Operations Research}, 2015.

\bibitem[Campolongo \& Orabona(2021)Campolongo and
  Orabona]{campolongo2021closer}
Campolongo, N. and Orabona, F.
\newblock A closer look at temporal variability in dynamic online learning,
  2021.

\bibitem[Cesa-Bianchi \& Lugosi(2006)Cesa-Bianchi and
  Lugosi]{cesa2006prediction}
Cesa-Bianchi, N. and Lugosi, G.
\newblock \emph{Prediction, learning, and games}.
\newblock Cambridge university press, 2006.

\bibitem[Cesa-Bianchi et~al.(2012)Cesa-Bianchi, Gaillard, Lugosi, and
  Stoltz]{cesa2012mirror}
Cesa-Bianchi, N., Gaillard, P., Lugosi, G., and Stoltz, G.
\newblock Mirror descent meets fixed share (and feels no regret).
\newblock \emph{Advances in Neural Information Processing Systems}, 25, 2012.

\bibitem[Cutkosky(2020)]{cutkosky2020parameter}
Cutkosky, A.
\newblock Parameter-free, dynamic, and strongly-adaptive online learning.
\newblock In \emph{Proceedings of the 37th International Conference on Machine
  Learning}, 2020.

\bibitem[Daniely et~al.(2015)Daniely, Gonen, and
  Shalev-Shwartz]{daniely2015strongly}
Daniely, A., Gonen, A., and Shalev-Shwartz, S.
\newblock Strongly adaptive online learning.
\newblock In \emph{Proceedings of the 32nd International Conference on Machine
  Learning}. PMLR, 2015.

\bibitem[Foster et~al.(2016)Foster, Kale, and Karloff]{foster2017sparse}
Foster, D., Kale, S., and Karloff, H.
\newblock Online sparse linear regression.
\newblock In \emph{29th Annual Conference on Learning Theory}. PMLR, 2016.

\bibitem[Gaillard et~al.(2019)Gaillard, Gerchinovitz, Huard, and
  Stoltz]{gaillard2019uniform}
Gaillard, P., Gerchinovitz, S., Huard, M., and Stoltz, G.
\newblock Uniform regret bounds over $\mathbb{R}^d$ for the sequential linear
  regression problem with the square loss.
\newblock In \emph{Proceedings of the 30th International Conference on
  Algorithmic Learning Theory}. PMLR, 2019.

\bibitem[Ghai et~al.(2020)Ghai, Lee, Singh, Zhang, and Zhang]{ghai2020noregret}
Ghai, U., Lee, H., Singh, K., Zhang, C., and Zhang, Y.
\newblock No-regret prediction in marginally stable systems.
\newblock In \emph{Proceedings of Thirty Third Conference on Learning Theory}.
  PMLR, 2020.

\bibitem[Hazan(2019)]{hazan2019introduction}
Hazan, E.
\newblock Introduction to online convex optimization.
\newblock \emph{CoRR}, abs/1909.05207, 2019.

\bibitem[Hazan \& Seshadhri(2007)Hazan and Seshadhri]{hazan2007adaptive}
Hazan, E. and Seshadhri, C.
\newblock Adaptive algorithms for online decision problems.
\newblock In \emph{Electronic colloquium on computational complexity (ECCC)},
  number 088, 2007.

\bibitem[Hazan \& Seshadhri(2009)Hazan and Seshadhri]{hazan2009efficient}
Hazan, E. and Seshadhri, C.
\newblock Efficient learning algorithms for changing environments.
\newblock In \emph{Proceedings of the 26th Annual International Conference on
  Machine Learning}, 2009.

\bibitem[Hazan \& Singh(2022)Hazan and Singh]{hazan2022introduction}
Hazan, E. and Singh, K.
\newblock Introduction to online nonstochastic control, 2022.

\bibitem[Hazan et~al.(2017)Hazan, Singh, and Zhang]{hazan2017learning}
Hazan, E., Singh, K., and Zhang, C.
\newblock Learning linear dynamical systems via spectral filtering.
\newblock \emph{Advances in Neural Information Processing Systems}, 2017.

\bibitem[Hazan et~al.(2018)Hazan, Lee, Singh, Zhang, and
  Zhang]{hazan2018spectral}
Hazan, E., Lee, H., Singh, K., Zhang, C., and Zhang, Y.
\newblock Spectral filtering for general linear dynamical systems.
\newblock \emph{Advances in Neural Information Processing Systems}, 2018.

\bibitem[Jacobsen \& Cutkosky(2022)Jacobsen and
  Cutkosky]{jacobsen2022parameter}
Jacobsen, A. and Cutkosky, A.
\newblock Parameter-free mirror descent.
\newblock In \emph{Proceedings of Thirty Fifth Conference on Learning Theory}.
  PMLR, 2022.

\bibitem[Jacobsen \& Cutkosky(2023)Jacobsen and
  Cutkosky]{jacobsen2023unconstrained}
Jacobsen, A. and Cutkosky, A.
\newblock Unconstrained online learning with unbounded losses.
\newblock In \emph{International Conference on Machine Learning (ICML)}. PMLR,
  2023.

\bibitem[Jun et~al.(2017)Jun, Orabona, Wright, and Willett]{jun2017improved}
Jun, K.-S., Orabona, F., Wright, S., and Willett, R.
\newblock {Improved Strongly Adaptive Online Learning using Coin Betting}.
\newblock In \emph{Proceedings of the 20th International Conference on
  Artificial Intelligence and Statistics}. PMLR, 2017.

\bibitem[Kalman(1960)]{kalman1960filter}
Kalman, R.~E.
\newblock A new approach to linear filtering and prediction problems.
\newblock \emph{Transactions of the ASME--Journal of Basic Engineering}, 1960.

\bibitem[Kempka et~al.(2019)Kempka, Kotlowski, and Warmuth]{kempka2019adaptive}
Kempka, M., Kotlowski, W., and Warmuth, M.~K.
\newblock Adaptive scale-invariant online algorithms for learning linear
  models.
\newblock In \emph{Proceedings of the 36th International Conference on Machine
  Learning}. PMLR, 2019.

\bibitem[Kivinen et~al.(2006)Kivinen, Warmuth, and Hassibi]{kivinen2006pnorm}
Kivinen, J., Warmuth, M., and Hassibi, B.
\newblock The p-norm generalization of the lms algorithm for adaptive
  filtering.
\newblock \emph{IEEE Transactions on Signal Processing}, 2006.

\bibitem[Kotłowski(2017)]{kotlowski2017scale}
Kotłowski, W.
\newblock Scale-invariant unconstrained online learning.
\newblock In \emph{Proceedings of the 28th International Conference on
  Algorithmic Learning Theory}. PMLR, 2017.

\bibitem[Kozdoba et~al.(2019)Kozdoba, Marecek, Tchrakian, and
  Mannor]{kozdoba2019online}
Kozdoba, M., Marecek, J., Tchrakian, T., and Mannor, S.
\newblock On-line learning of linear dynamical systems: Exponential forgetting
  in kalman filters.
\newblock In \emph{Proceedings of the AAAI Conference on Artificial
  Intelligence}, 2019.

\bibitem[Lu \& Hazan(2022)Lu and Hazan]{lu2022efficient}
Lu, Z. and Hazan, E.
\newblock Efficient adaptive regret minimization, 2022.

\bibitem[Luo et~al.(2016)Luo, Agarwal, Cesa-Bianchi, and
  Langford]{luo2016efficient}
Luo, H., Agarwal, A., Cesa-Bianchi, N., and Langford, J.
\newblock Efficient second order online learning by sketching.
\newblock In \emph{Advances in Neural Information Processing Systems}, 2016.

\bibitem[Luo et~al.(2022)Luo, Zhang, Zhao, and Zhou]{luo2022corralling}
Luo, H., Zhang, M., Zhao, P., and Zhou, Z.-H.
\newblock Corralling a larger band of bandits: A case study on switching regret
  for linear bandits.
\newblock In \emph{Proceedings of Thirty Fifth Conference on Learning Theory}.
  PMLR, 2022.

\bibitem[Mayo et~al.(2022)Mayo, Hadiji, and van Erven]{mayo2022scalefree}
Mayo, J.~J., Hadiji, H., and van Erven, T.
\newblock Scale-free unconstrained online learning for curved losses.
\newblock In \emph{Proceedings of Thirty Fifth Conference on Learning Theory},
  2022.

\bibitem[Mhammedi \& Koolen(2020)Mhammedi and Koolen]{mhammedi2020lipschitz}
Mhammedi, Z. and Koolen, W.~M.
\newblock Lipschitz and comparator-norm adaptivity in online learning.
\newblock In Abernethy, J. and Agarwal, S. (eds.), \emph{Proceedings of Thirty
  Third Conference on Learning Theory}. PMLR, 2020.

\bibitem[Orabona et~al.(2015)Orabona, Crammer, and
  Cesa-Bianchi]{orabona2015generalized}
Orabona, F., Crammer, K., and Cesa-Bianchi, N.
\newblock A generalized online mirror descent with applications to
  classification and regression.
\newblock \emph{Mach. Learn.}, 2015.

\bibitem[Rashidinejad et~al.(2020)Rashidinejad, Jiao, and
  Russell]{rashidinejad2020slip}
Rashidinejad, P., Jiao, J., and Russell, S.
\newblock Slip: Learning to predict in unknown dynamical systems with long-term
  memory.
\newblock In \emph{Advances in Neural Information Processing Systems}, 2020.

\bibitem[Simon(2006)]{simon2006optimal}
Simon, D.
\newblock \emph{Optimal state estimation: Kalman, H infinity, and nonlinear
  approaches}.
\newblock John Wiley \& Sons, 2006.

\bibitem[Tsiamis \& Pappas(2022)Tsiamis and Pappas]{tsiamis2022online}
Tsiamis, A. and Pappas, G.~J.
\newblock Online learning of the kalman filter with logarithmic regret.
\newblock \emph{IEEE Transactions on Automatic Control}, 2022.

\bibitem[Veness et~al.(2013)Veness, White, Bowling, and
  György]{veness2013partition}
Veness, J., White, M., Bowling, M., and György, A.
\newblock Partition tree weighting.
\newblock In \emph{2013 Data Compression Conference}, 2013.

\bibitem[Vovk(2001)]{vovk2001competitive}
Vovk, V.
\newblock Competitive on-line statistics.
\newblock \emph{International Statistical Review}, 2001.

\bibitem[Yuan \& Lamperski(2019)Yuan and Lamperski]{yuan2019trading}
Yuan, J. and Lamperski, A.~G.
\newblock Trading-off static and dynamic regret in online least-squares and
  beyond.
\newblock \emph{CoRR}, abs/1909.03118, 2019.

\bibitem[Zhang et~al.(2018)Zhang, Lu, and Zhou]{zhang2018adaptive}
Zhang, L., Lu, S., and Zhou, Z.-H.
\newblock Adaptive online learning in dynamic environments.
\newblock In \emph{Advances in Neural Information Processing Systems}, 2018.

\bibitem[Zhang et~al.(2023)Zhang, Cutkosky, and
  Paschalidis]{zhang2023unconstrained}
Zhang, Z., Cutkosky, A., and Paschalidis, Y.
\newblock Unconstrained dynamic regret via sparse coding.
\newblock In \emph{Advances in Neural Information Processing Systems}, 2023.

\bibitem[Zhao et~al.(2020)Zhao, Zhang, Zhang, and Zhou]{zhao2020dynamic}
Zhao, P., Zhang, Y.-J., Zhang, L., and Zhou, Z.-H.
\newblock Dynamic regret of convex and smooth functions.
\newblock In \emph{Advances in Neural Information Processing Systems}, 2020.

\bibitem[Zhao et~al.(2024)Zhao, Zhang, Zhang, and Zhou]{zhao2024adaptivity}
Zhao, P., Zhang, Y.-J., Zhang, L., and Zhou, Z.-H.
\newblock Adaptivity and non-stationarity: Problem-dependent dynamic regret for
  online convex optimization.
\newblock \emph{Journal of Machine Learning Research}, 2024.

\end{thebibliography}

\newpage
\onecolumn

\appendix

\section{Proofs for Section~\ref{sec:discounted-vaw} (\SecDiscountedVAW)}%
\label{app:discounted-vaw-proofs}

\subsection{Equivalence to FTRL and Mirror Descent}%
\label{app:discounted-vaw-update}
We accomplish our analysis of the discounted VAW forecaster using the
equivalence
in the following proposition, proving both optimistic FTRL and
and optimistic mirror descent interpretations of the discounted VAW forecaster.
\Cref{eq:discounted-vaw-update:ftrl}
is perhaps the most natural interpretation of the update: it says that the
discounted VAW forecaster chooses the $\w$
which minimizes the \emph{discounted sum}
$h_{t}(\w)+\gamma\ell_{\tmm}(\w)+\gamma^{2}\ell_{t-2}(\w)+\ldots$, thus placing greater
emphasis on the most-recent losses and the hint function $h_{t}(\w)$.
However, it is not at all obvious how to analyze the dynamic regret
of the discounted VAW forecaster when interpreted in this FTRL-like form.
Rather, the key to our results
in this work is to instead approach the analysis through the lens of the mirror
descent update (\Cref{eq:discounted-vaw-update:md}). Interestingly,
a similar mirror descent interpretation was used in the seminal work of
\citet{azoury2001relative}, though they did not account for an arbitrary
$\yhint_{t}$ and they did not refer to the algorithm in terms of mirror descent.
\begin{restatable}{proposition}{DiscountedVAWUpdate}\label{prop:discounted-vaw-update}
  (Discounted VAW Forecaster)
  Let $\gamma\in(0,1]$, $\lambda>0$, $\yhint_{1}=0$, and $\yhint_{t}\in\R$  for $t>1$.
  Define
  $h_{t}(\w)=\half(\yhint_{t}-\inner{\xt,\w})^{2}$ and $\ell_{0}(\w)=\frac{\lambda}{2}\norm{\w}^{2}_{2}$.
  Recursively define $\Sigma_{t}=\xt\xt^{\top}+\gamma \Sigma_{\tmm}$ starting from
  $\Sigma_{0}=\lambda I$, let $\psi_{t}(\w)=\half\norm{\w}^{2}_{\Sigma_{t}}$ and set
  $w_{1}=\argmin_{w\in \R^{d}}\psi_{1}(w)=\zeros$.
  Then the following are equivalent
    \vspace{-1em}
  \begin{align}
    &\Sigma_{t}^{\inv}\Big[\yhint_{t}\xt+\gamma\sum_{s=1}^{\tmm}\gamma^{t-1-s}\y_{s}\x_{s}\Big]\label{eq:discounted-vaw-update:closed-form}\\
    \argmin_{\w\in \R^{d}}&h_{t}(\w)+\gamma\sum_{s=0}^{\tmm}\gamma^{t-1-s}\ell_{s}(s) \label{eq:discounted-vaw-update:ftrl}\\
    \argmin_{\w\in\R^{d}}& (\gamma \ell_{\tmm}-\gamma h_{\tmm}+h_{t})(\w)+\gamma D_{\psi_{\tmm}}(\w|\wtmm)\label{eq:discounted-vaw-update:md}
  \end{align}
\end{restatable}
\begin{remark}\label{remark:gamma-zero}
Note that with $\gamma=0$, \Cref{eq:discounted-vaw-update:ftrl,eq:discounted-vaw-update:md}
prescribe choosing any $\wt$ satisfying
$\inner{\wt,\xt}=\yhint_{t}$. The choice is not unique, but nevertheless
it will often be convenient to refer to an algorithm which greedily predicts
$\yhint_{t}$ on each round as an instance of \Cref{alg:discounted-vaw} with $\gamma=0$.
\end{remark}
\begin{proof}
  The result follows by showing that
  \Cref{eq:discounted-vaw-update:ftrl,eq:discounted-vaw-update:md}
  are both equivalent to \Cref{eq:discounted-vaw-update:closed-form}.
  First consider the former, \Cref{eq:discounted-vaw-update:ftrl}. From the first-order optimality condition we have
  \begin{align*}
    \zeros
    &=
      \grad h_{t}(\wt)+\gamma\sum_{s=0}^{\tmm}\gamma^{\tmm-s}\grad\ell_{s}(\wt)\\
    &=
      -(\yhint_{t}-\inner{\xt,\wt})\xt-\gamma \sum_{s=1}^{\tmm}\gamma^{\tmm-s}(\y_{s}-\inner{\x_{s},\wt})\x_{s}+\gamma^{t}\lambda\wt,
  \end{align*}
  where the last line recalls that we defined $\ell_{0}(\w)=\frac{\lambda}{2}\norm{\w}^{2}_{2}$.
  Hence,
  \begin{align*}
    \brac{\gamma^{t}\lambda I + \sum_{s=1}^{t}\gamma^{t-s}\x_{s}\x_{s}^{\top}}\wt&= \yhint_{t}\xt+\sum_{s=1}^{t}\gamma^{t-s}\y_{s}\x_{s}\\
    \implies
    \wt&= \brac{\gamma^{t}\lambda I + \sum_{s=1}^{t}\gamma^{t-s}\x_{s}\x_{s}^{\top}}^{\inv}\sbrac{\yhint_{t}\xt+\gamma\sum_{s=1}^{\tmm}\gamma^{\tmm-s}y_{s}\x_{s}}\\
     &= \Sigma_{t}^{\inv}\sbrac{\yhint_{t}\xt+\gamma\sum_{s=1}^{\tmm}\gamma^{\tmm-s}y_{s}\x_{s}},
  \end{align*}
  where the last line can be seen by unrolling the recursion for $\Sigma_{t}$.

  Likewise, consider \Cref{eq:discounted-vaw-update:md}. From the first-order optimality condition
  $\wt= \argmin_{\w\in\R^{d}} (\gamma \ell_{\tmm}-\gamma h_{\tmm}+h_{t})(\w)+\gamma D_{\psi_{\tmm}}(\w|\wt)$,
  we have
  \begin{align*}
    \zeros
    &=
      \gamma(\grad\ell_{\tmm}(\wt)-\grad h_{\tmm}(\wt))+\grad h_{t}(\wt)+\gamma \sbrac{\grad\psi_{\tmm}(\wt)-\grad\psi_{\tmm}(\wtmm)}\\
    &=
      -\gamma\ytmm\xtmm+\gamma\yhint_{\tmm}\xtmm -\yhint_{t}\xt + \xt\xt^{\top}\wt + \gamma \Sigma_{\tmm}\wt -\gamma \Sigma_{\tmm}\wtmm\\
    &=
      -\gamma\ytmm\xtmm+\gamma\yhint_{\tmm}\xtmm -\yhint_{t}\xt + \Sigma_{t}\wt-\gamma \Sigma_{\tmm}\wtmm,
  \end{align*}
  where the last line observes that $\Sigma_{t}=\xt\xt^{\top}+\gamma \Sigma_{\tmm}$ by
  construction.
  Hence, re-arranging we have
  \begin{align*}
    \Sigma_{t}\wt
    &=
      \yhint_{t}\xt+\gamma\ytmm\xtmm -\gamma\yhint_{\tmm}\xtmm+ \gamma \Sigma_{\tmm}\wtmm \\
    \intertext{and unrolling the recursion:}
    &=
       \yhint_{t}\xt+\gamma\ytmm\xtmm -\gamma\yhint_{\tmm}\xtmm+\gamma \sbrac{\yhint_{\tmm}\xtmm +\gamma\y_{t-2}\x_{t-2} - \gamma\yhint_{t-2}\xt + \gamma \Sigma_{t-2}\w_{t-2}}\\
    &=
      \yhint_{t}\xt+\gamma\ytmm\xtmm+\gamma^{2}\y_{t-2}\x_{t-2}-\gamma^{2}\yhint_{t-2}\x_{t-2}+\gamma^{2} \Sigma_{t-2}\w_{t-2} \\
    &=
      \ldots\\
    &=
     \yhint_{t}\xt-\gamma^{\tmm}\yhint_{1}\x_{1} + \gamma \sum_{s=1}^{\tmm}\gamma^{\tmm-s}\y_{s}\x_{s}\\
    &=
     \yhint_{t}\xt + \gamma \sum_{s=1}^{\tmm}\gamma^{\tmm-s}\y_{s}\x_{s},
  \end{align*}
  for $\yhint_{1}=0$. Hence, applying $\Sigma_{t}^{\inv}$ to both sides we have
  \begin{align*}
    \wt
    &=
      \Sigma_{t}^{\inv}\sbrac{\yhint_{t}\xt + \gamma\sum_{s=1}^{\tmm}\gamma^{\tmm-s}\y_{s}\x_{s}}\\
  \end{align*}

\end{proof}

\subsection{Proof of Theorem~\ref{thm:general-discounted-vaw}}%
\label{app:general-discounted-vaw}
\GeneralDiscountedVAW*
\begin{proof}
  Begin by applying the regret template provided by
  \Cref{lemma:discounted-vaw-decomp}:
  \begin{align*}
    R_{T}(\vec{\cmp})
    &\le
      \sumtT D_{\psi_{t}}(\cmp_{t}|\wt)- D_{\psi_{\tpp}}(\cmp_{t}|\wtpp)
      +\sumtT h_{\tpp}(\cmp_{t})-h_{t}(\cmp_{t})+\half\sumtT(\yt-\yhint_{t})^{2}\norm{\xt}^{2}_{\Sigma_{t}^{-1}},\\
    \intertext{bound the first two summations using \Cref{lemma:general-discounted-vaw-divergences}:}
    &\le
      \frac{\gamma\lambda}{2}\norm{\cmp_{1}}^{2}_{2}+h_{T+1}(\cmp_{T})+\gamma\sum_{t=1}^{T-1}\sbrac{F_{t}^{\gamma}(\cmp_{\tpp})-F_{t}^{\gamma}(\cmp_{t})}+\half\sumtT(\yt-\yhint_{t})^{2}\norm{\xt}^{2}_{\Sigma_{t}^{\inv}},\\
      \intertext{and apply a discounted variant of the log-determinant lemma (\Cref{lemma:adaptive-discounted-log-det}) to
      bound the final summation:}
    &\le
      \frac{\gamma\lambda}{2}\norm{\cmp_{1}}^{2}_{2}+h_{T+1}(\cmp_{T})+\frac{d}{2}\max_{t}(\yt-\yhint_{t})^{2}\Log{1+\frac{\sumtT \gamma^{T-t}\norm{\x_{t}}^{2}_{2}}{\lambda d}}\\
    &\qquad
    +\gamma\sum_{t=1}^{T-1}\sbrac{F_{t}^{\gamma}(\cmp_{\tpp})-F_{t}^{\gamma}(\cmp_{t})}
    +\frac{d}{2}\Log{1/\gamma}\sumtT(\yt-\yhint_{t})^{2}
  \end{align*}
  Finally, since the regret does not depend on $h_{T+1}(\cdot)$ we may
  set $h_{T+1}(\cdot)\equiv 0$ in the analysis and hide constants to arrive at the
  stated bound.
\end{proof}

\subsubsection{Proof of Lemma~\ref{lemma:discounted-vaw-decomp}}%
\label{app:discounted-vaw-decomp}
The following lemma provides the base regret decomposition that we use as a
jumping-off point to prove \Cref{thm:general-discounted-vaw}. The result follows using mostly standard mirror descent
analysis, though with a bit of additional care to handle issues related
to the discounted regularizer.
\begin{restatable}{lemma}{DiscountedVAWDecomp}\label{lemma:discounted-vaw-decomp}
  Let $\gamma\in(0,1]$. Then
  for any sequence $\vec{\cmp}=(\cmp_{1},\ldots,\cmp_{T})$ in $\R^{d}$,
  \Cref{alg:discounted-vaw} guarantees
  \begin{align*}
    R_{T}(\vec{\cmp})
    &\le
      \sumtT D_{\psi_{t}}(\cmp_{t}|\wt)- D_{\psi_{\tpp}}(\cmp_{t}|\wtpp)\\
    &\qquad
      +\sumtT h_{\tpp}(\cmp_{t})- h_{t}(\cmp_{t})\\
    &\qquad
      +\sumtT \half(\yt-\yhint_{t})^{2}\norm{\xt}^{2}_{\Sigma_{t}^{\inv}}
  \end{align*}
\end{restatable}
\begin{proof}
  We will proceed following a mirror-descent-based analysis, and thus begin by
  exposing the terms $(\gamma\ell_{t}-\gamma h_{t}+h_{\tpp})(\wtpp)$ observed in
  the mirror-descent interpretation of the update (\Cref{eq:discounted-vaw-update:md}):
  \begin{align}
    R_{T}(\vec{\cmp})
    &=
      \sumtT\ell_{t}(\wt)-\ell_{t}(\cmp_{t})\nonumber\\
    &=
      \sumtT \gamma\sbrac{\ell_{t}(\wt)-\ell_{t}(\cmp_{t})}+(1-\gamma)\sumtT\ell_{t}(\wt)-\ell_{t}(\cmp_{t})\nonumber\\
    &=
      \sumtT \gamma\sbrac{(\ell_{t}-h_{t})(\wt)-(\ell_{t}-h_{t})(\cmp_{t})}+\sumtT\gamma h_{t}(\wt)-\gamma h_{t}(\cmp_{t})\nonumber\\
    &\qquad
      +(1-\gamma)\sumtT\ell_{t}(\wt)-\ell_{t}(\cmp_{t})\nonumber\\
    &=
      \sumtT \gamma\sbrac{(\ell_{t}-h_{t})(\wtpp)-(\ell_{t}-h_{t})(\cmp_{t})}+\sumtT\gamma h_{t}(\wt)-\gamma h_{t}(\cmp_{t})\nonumber\\
    &\qquad
      +\gamma\sumtT (\ell_{t}-h_{t})(\wt)-(\ell_{t}-h_{t})(\wtpp)\nonumber\\
    &\qquad
      +(1-\gamma)\sumtT\ell_{t}(\wt)-\ell_{t}(\cmp_{t})\nonumber\\
    &=
      \sumtT (\gamma\ell_{t}-\gamma h_{t}+h_{\tpp})(\wtpp)-(\gamma \ell_{t}-\gamma h_{t}+h_{\tpp})(\cmp_{t})\nonumber\\
    &\qquad
      +\sumtT\gamma h_{t}(\wt)-h_{\tpp}(\wtpp)+\sumtT h_{\tpp}(\cmp_{t})-\gamma h_{t}(\cmp_{t})\nonumber\\
    &\qquad
      +\gamma\sumtT (\ell_{t}-h_{t})(\wt)-(\ell_{t}-h_{t})(\wtpp)\nonumber\\
    &\qquad
      +(1-\gamma)\sumtT\ell_{t}(\wt)-\ell_{t}(\cmp_{t})\nonumber\\
    \intertext{Re-arranging factors of $\gamma$ from the second-line and
    observing that
    $\sumtT h_{t}(\wt)-h_{\tpp}(\wtpp)=h_{1}(\w_{1})-h_{T+1}(\w_{T+1})$:}
    &=
      \sumtT (\gamma\ell_{t}-\gamma h_{t}+h_{\tpp})(\wtpp)-(\gamma \ell_{t}-\gamma h_{t}+h_{\tpp})(\cmp_{t})\nonumber\\
    &\qquad
      +\sumtT h_{t}(\wt)-h_{\tpp}(\wtpp)+\sumtT -(1-\gamma)h_{t}(\wt)+(1-\gamma)h_{t}(\cmp_{t})+\sumtT h_{\tpp}(\cmp_{t})- h_{t}(\cmp_{t})\nonumber\\
    &\qquad
      +\gamma\sumtT (\ell_{t}-h_{t})(\wt)-(\ell_{t}-h_{t})(\wtpp)\nonumber\\
    &\qquad
      +(1-\gamma)\sumtT\ell_{t}(\wt)-\ell_{t}(\cmp_{t})\nonumber\\
    &=
      \sumtT (\gamma\ell_{t}-\gamma h_{t}+h_{\tpp})(\wtpp)-(\gamma \ell_{t}-\gamma h_{t}+h_{\tpp})(\cmp_{t})\nonumber\\
    &\qquad
      +h_{1}(\w_{1})-h_{T+1}(\w_{T+1})+\sumtT h_{\tpp}(\cmp_{t})- h_{t}(\cmp_{t})\nonumber\\
    &\qquad
      +\gamma\sumtT (\ell_{t}-h_{t})(\wt)-(\ell_{t}-h_{t})(\wtpp)\nonumber\\
    &\qquad
      +(1-\gamma)\sumtT(\ell_{t}-h_{t})(\wt)-(\ell_{t}-h_{t})(\cmp_{t})\label{eq:discounted-vaw-decomp:initial}
  \end{align}
  Moreover, from the first-order optimality condition
  $\wtpp = \argmin_{\w\in \R^{d}}(\gamma \ell_{t}-\gamma h_{t}+h_{\tpp})(\w)+\gamma D_{\psi_{t}}(\w|\wt)$,
  we have
  \begin{align*}
      \inner{\grad(\gamma \ell_{t}-\gamma h_{t}+h_{\tpp})(\wtpp)+\gamma \grad\psi_{t}(\wtpp)-\gamma\grad\psi_{t}(\wt),\wtpp-\cmp_{t}}\le 0
  \end{align*}
  so re-arranging:
  \begin{align*}
    \inner{\grad(\gamma\ell_{t}-\gamma h_{t}+h_{\tpp})(\wtpp),\wtpp-\cmp_{t}}
    &\le
      \gamma\inner{\grad\psi_{t}(\wt)-\grad\psi_{t}(\wtpp),\wtpp-\cmp_{t}}\\
    &=
      \gamma D_{\psi_{t}}(\cmp_{t}|\wt)-\gamma D_{\psi_{t}}(\cmp_{t}|\wtpp)-\gamma D_{\psi_{t}}(\wtpp|\wt),
  \end{align*}
  where the last line uses the three-point relation for bregman divergences,
  $\inner{\grad f(w)-\grad f(\w^{\prime}),\w^{\prime}-\cmp}=D_{f}(\cmp|\w)-D_{f}(\cmp|\w^{\prime})-D_{f}(\w^{\prime}|\w)$.
  Thus,
  \begin{align*}
    &\sumtT(\gamma \ell_{t}-\gamma h_{t}+h_{\tpp})(\wtpp)-(\gamma \ell_{t}-\gamma h_{t}+h_{\tpp})(\cmp_{t})\\
    &\qquad\overset{(a)}{=}
      \sumtT \inner{\grad(\gamma \ell_{t}-\gamma h_{t}+h_{\tpp})(\wtpp),\wtpp-\cmp_{t}}-D_{\gamma\ell_{t}-\gamma h_{t}+h_{\tpp}}(\cmp_{t}|\wtpp)\\
    &\qquad\le
      \sumtT \gamma D_{\psi_{t}}(\cmp_{t}|\wt)-\gamma D_{\psi_{t}}(\cmp_{t}|\wtpp)-\gamma D_{\psi_{t}}(\wtpp|\wt)-D_{\gamma\ell_{t}-\gamma h_{t}+h_{\tpp}}(\cmp_{t}|\wtpp)\\
    &\qquad\overset{(b)}{=}
      \sumtT \gamma D_{\psi_{t}}(\cmp_{t}|\wt)-\gamma D_{\psi_{t}}(\cmp_{t}|\wtpp)-D_{h_{\tpp}}(\cmp_{t}|\wtpp)-\gamma D_{\psi_{t}}(\wtpp|\wt)\\
    &\qquad\overset{(c)}{=}
      \sumtT \gamma D_{\psi_{t}}(\cmp_{t}|\wt)- D_{\psi_{\tpp}}(\cmp_{t}|\wtpp)-\gamma D_{\psi_{t}}(\wtpp|\wt)\\
    &\qquad=
      \sumtT D_{\psi_{t}}(\cmp_{t}|\wt)- D_{\psi_{\tpp}}(\cmp_{t}|\wtpp)-(1-\gamma)D_{\psi_{t}}(\cmp_{t}|\wt)-\gamma D_{\psi_{t}}(\wtpp|\wt),
  \end{align*}
  where $(a)$ uses the definition of Bregman divergence to re-write
  $f(\w)-f(\cmp)=\inner{\grad f(\w),\w-\cmp}-D_{f}(\cmp|\w)$,
  $(b)$ observes that
  $\gamma(\ell_{t}-h_{t})(\w)=\gamma\brac{\half \yt^{2}-\half \yhint_{t}^{2}+(\yhint_{t}-\yt)\inner{\xt,\w}}$,
  so
  $D_{\gamma\ell_{t}-\gamma h_{t}+h_{\tpp}}(\cdot|\cdot)=D_{h_{\tpp}}(\cdot|\cdot)$
  due to the invariance of Bregman divergences to linear terms,
  and $(c)$ recalls that $\Sigma_{\tpp}=\xtpp\xtpp^{\top}+\gamma \Sigma_{t}$ so that
  overall we have:
  \begin{align*}
    \gamma D_{\psi_{t}}(\cmp_{t}|\wtpp)+D_{h_{\tpp}}(\cmp_{t}|\wtpp)
    &=
      \frac{\gamma}{2}\norm{\cmp_{t}-\wtpp}^{2}_{\Sigma_{t}} +\half \inner{\xtpp,\cmp_{t}-\wtpp}^{2}\\
    &=
      \half\norm{\cmp_{t}-\wtpp}^{2}_{\Sigma_{\tpp}}\\
    &=
      D_{\psi_{\tpp}}(\cmp_{t}|\wtpp).
  \end{align*}
  Plugging this back into \Cref{eq:discounted-vaw-decomp:initial}, we have
  \begin{align*}
    R_{T}(\vec{\cmp})
    &\le
      \sumtT D_{\psi_{t}}(\cmp_{t}|\wt)- D_{\psi_{\tpp}}(\cmp_{t}|\wtpp)\\
    &\qquad
      +h_{1}(\w_{1})-h_{T+1}(\w_{T+1})+\sumtT h_{\tpp}(\cmp_{t})- h_{t}(\cmp_{t})\nonumber\\
    &\qquad
      +\gamma\sumtT (\ell_{t}-h_{t})(\wt)-(\ell_{t}-h_{t})(\wtpp)-D_{\psi_{t}}(\wtpp|\wt)\nonumber\\
    &\qquad
      +(1-\gamma)\sumtT(\ell_{t}-h_{t})(\wt)-(\ell_{t}-h_{t})(\cmp_{t})-D_{\psi_{t}}(\cmp_{t}|\wtpp).
  \end{align*}
  Finally, observe that for
  any $u,v\in \R^{d}$,
  $(\ell_{t}-h_{t})(u)-(\ell_{t}-h_{t})(v)=(\yhint_{t}-\yt)\inner{\xt,u-v}$,
  so an application of Fenchel-Young inequality yields
  \begin{align*}
    (\ell_{t}-h_{t})(u)-(\ell_{t}-h_{t})(v)-D_{\psi_{t}}(v|u)
    &=
      (\yhint_{t}-\yt)\inner{\xt,u-v}-\half\norm{u-v}^{2}_{\Sigma_{t}}\\
    &\le
      \half (\yt-\yhint_{t})^{2}\norm{\xt}^{2}_{\Sigma_{t}^{\inv}}.
  \end{align*}
  Applying this in the last two lines of the previous display yields
  \begin{align*}
    R_{T}(\vec{\cmp})
    &\le
      \sumtT D_{\psi_{t}}(\cmp_{t}|\wt)- D_{\psi_{\tpp}}(\cmp_{t}|\wtpp)\\
    &\qquad
      \underbrace{h_{1}(\w_{1})-h_{T+1}(\w_{T+1})}_{\le 0}+\sumtT h_{\tpp}(\cmp_{t})- h_{t}(\cmp_{t})\nonumber\\
    &\qquad
      \gamma\sumtT \half(\yt-\yhint_{t})^{2}\norm{\xt}^{2}_{\Sigma_{t}^{\inv}}+(1-\gamma)\sumtT\half(\yt-\yhint_{t})^{2}\norm{\xt}^{2}_{\Sigma_{t}^{\inv}}\\
    &\le
      \sumtT D_{\psi_{t}}(\cmp_{t}|\wt)- D_{\psi_{\tpp}}(\cmp_{t}|\wtpp)\\
    &\qquad
      +\sumtT h_{\tpp}(\cmp_{t})- h_{t}(\cmp_{t})\nonumber\\
    &\qquad
      +\sumtT \half(\yt-\yhint_{t})^{2}\norm{\xt}^{2}_{\Sigma_{t}^{\inv}}
  \end{align*}
\end{proof}

\subsubsection{Proof of Lemma~\ref{lemma:general-discounted-vaw-divergences}}%
\label{app:general-discounted-vaw-divergences}
The following lemma bounds the sum of divergence terms.
Intuitively, the goal here is to remove all instances of $\wt$ from the
analysis, since in an unbounded domain any terms depending on $\wt$ will
be hard to quantify and could be arbitrarily large in general.
\Cref{lemma:general-discounted-vaw-divergences} shows how get rid of the
$\wt$-dependent terms left in the bound from
\Cref{lemma:discounted-vaw-decomp},
such that only dependencies on the comparators $\cmp_{t}$ remain.
\begin{restatable}{lemma}{GeneralDiscountedVAWDivergences}\label{lemma:general-discounted-vaw-divergences}
  Under the same conditions as \Cref{lemma:discounted-vaw-decomp},
  \begin{align*}
    \sumtT D_{\psi_{t}}(\cmp_{t}|\wt)-D_{\psi_{\tpp}}(\cmp_{t}|\wtpp)+\sumtT h_{\tpp}(\cmp_{t})-h_{t}(\cmp_{t})
    &\le
      \frac{\gamma\lambda}{2}\norm{\cmp_{1}}^{2}_{2}+h_{T+1}(\cmp_{T})+\gamma\sum_{t=1}^{T-1}F_{t}^{\gamma}(\cmp_{\tpp})-F_{t}^{\gamma}(\cmp_{t}).
  \end{align*}
  where
  $F_{t}^{\gamma}(\w)=\sum_{s=0}^{t}\gamma^{t-s}\ell_{s}(\w)$.
\end{restatable}
\begin{proof}
  Observe that by \Cref{lemma:loss-divergence} we have
  $D_{\ell_{t}}(u|v)=\half\inner{\xt,u-v}^{2}=D_{h_{t}}(u|v)$ for any $u,v\in W$.
  Hence, letting
  $F_{t}^{\gamma}(\w)=\sum_{s=0}^{t}\gamma^{t-s}\ell_{s}(\w)$ and
  $\hat F_{t}^{\gamma}(\w)=h_{t}(\w)+\gamma F_{\tmm}^{\gamma}(\w)$,
  and recalling
  $\psi_{t}(\w)=\half\norm{\w}^{2}_{\Sigma_{t}}=\frac{\gamma^{t}\lambda}{2}\norm{\w}^{2}_{2}+\half\sum_{s=1}^{t}\gamma^{t-s}\inner{\x_{s},\w}^{2}$,
  we have $D_{\psi_{t}}(u|v)=D_{\hat F_{t}^{\gamma}}(u|v)$ for any $u,v\in \in\R^{d}$. Thus:
  \begin{align}
    &\sumtT D_{\psi_{t}}(\cmp_{t}|\wt)-D_{\psi_{\tpp}}(\cmp_{t}|\wtpp)\nonumber\\
    &\qquad=
      D_{\psi_{1}}(\cmp_{1}|\w_{1})-D_{\psi_{T+1}}(\cmp_{T}|\w_{T+1})
      +\sum_{t=2}^{T}D_{\psi_{t}}(\cmp_{t}|\wt)-D_{\psi_{t}}(\cmp_{\tmm}|\wt)\nonumber\\
    &\qquad=
      D_{\psi_{1}}(\cmp_{1}|\w_{1})-D_{\psi_{T+1}}(\cmp_{T}|\w_{T+1})
      +\sum_{t=2}^{T}D_{\hat F_{t}^{\gamma}}(\cmp_{t}|\wt)-D_{\hat F_{t}^{\gamma}}(\cmp_{\tmm}|\wt)\nonumber\\
    &\qquad=
      D_{\psi_{1}}(\cmp_{1}|\w_{1})-D_{\psi_{T+1}}(\cmp_{T}|\w_{T+1})
      +\sum_{t=2}^{T}\hat F_{t}^{\gamma}(\cmp_{t})-\hat F_{t}^{\gamma}(\cmp_{\tmm})-\inner{\grad \hat F_{t}^{\gamma}(\wt),\cmp_{t}-\cmp_{\tmm}}\nonumber.
  \end{align}
  Moreover, by \Cref{prop:discounted-vaw-update} we have
  \begin{align*}
    \wt = \argmin_{w\in \R^{d}}h_{t}(\w)+\gamma\sum_{s=0}^{\tmm}\gamma^{t-1-s}\ell_{s}(\w)=\argmin_{w\in \R^{d}} \hat F_{t}^{\gamma}(\w),
  \end{align*}
  hence by convexity of $\hat F_{t}^{\gamma}$ and the first-order optimality
  condition we have $\grad \hat F_{t}^{\gamma}(\wt)=\zeros$,
  so overall we have
  \begin{align*}
    &\sum_{t=1}^{T}D_{\psi_{t}}(\cmp_{t}|\wt)-D_{\psi_{t}}(\cmp_{t}|\wtpp)+\sumtT h_{\tpp}(\cmp_{t})-h_{t}(\cmp_{t})\\
    &\qquad=
      D_{\psi_{1}}(\cmp_{1}|\w_{1})-D_{\psi_{T+1}}(\cmp_{T}|\w_{T+1})
    +\sum_{t=2}^{T} \hat F_{t}^{\gamma}(\cmp_{t})-\hat F_{t}^{\gamma}(\cmp_{\tmm})+\sumtT h_{\tpp}(\cmp_{t})-h_{t}(\cmp_{t})\\
    &\qquad=
      D_{\psi_{1}}(\cmp_{1}|\w_{1})-D_{\psi_{T+1}}(\cmp_{T}|\w_{T+1})
      +\sum_{t=2}^{T} \sbrac{h_{t}(\cmp_{t})-h_{t}(\cmp_{\tmm})+\gamma F_{\tmm}^{\gamma}(\cmp_{t})-\gamma F_{\tmm}^{\gamma}(\cmp_{\tmm})}+\sumtT h_{\tpp}(\cmp_{t})-h_{t}(\cmp_{t})\\
    &\qquad=
      D_{\psi_{1}}(\cmp_{1}|\w_{1})-D_{\psi_{T+1}}(\cmp_{T}|\w_{T+1})
      +\gamma\sum_{t=1}^{T-1}F_{t}^{\gamma}(\cmp_{\tpp})-F_{t}^{\gamma}(\cmp_{t})
      +\sum_{t=2}^{T}h_{\tpp}(\cmp_{t})-h_{t}(\cmp_{\tmm})
      + h_{2}(u_{1})-h_{1}(u_{1})\\
    &\qquad=
      D_{\psi_{1}}(\cmp_{1}|\w_{1})-D_{\psi_{T+1}}(\cmp_{T}|\w_{T+1})
      +h_{T+1}(\cmp_{T})-h_{1}(\cmp_{1})
      +\gamma\sum_{t=1}^{T-1}F_{t}^{\gamma}(\cmp_{\tpp})-F_{t}^{\gamma}(\cmp_{t}).
  \end{align*}
  Finally, observe that with $\w_{1}=\zeros$ and $\yhint_{1}=0$ we have
  \begin{align*}
    D_{\psi_{1}}(\cmp_{1}|\w_{1})=\psi_{1}(\cmp_{1})-\psi_{1}(\zeros)-\inner{\grad\psi_{1}(\zeros),\cmp_{1}}
    =h_{1}(\cmp_{1})+\gamma\ell_{0}(\cmp_{1})=h_{1}(\cmp_{1})+\frac{\gamma\lambda}{2}\norm{\cmp_{1}}^{2}_{2}
  \end{align*}
  so we can express the bound as
  the bound as
  \begin{align*}
    &\sum_{t=1}^{T}D_{\psi_{t}}(\cmp_{t}|\wt)-D_{\psi_{t}}(\cmp_{t}|\wtpp)+\sumtT h_{\tpp}(\cmp_{t})-h_{t}(\cmp_{t})\\
    &\qquad\le
      \frac{\gamma\lambda}{2}\norm{\cmp_{1}}^{2}_{2}+h_{T+1}(\cmp_{T})+\gamma\sum_{t=1}^{T-1}F_{t}^{\gamma}(\cmp_{\tpp})-F_{t}^{\gamma}(\cmp_{t}).
  \end{align*}

\end{proof}

\subsection{Proof of Lemma~\ref{lemma:discounted-terms}}%
\label{app:discounted-terms}
The following lemma bounds the variability and stability terms
from \Cref{thm:general-discounted-vaw} to expose a more explicit
trade-off in terms of the discount factor $\gamma$.
\begin{manuallemma}{\ref{lemma:discounted-terms}}
  Let
  $\ell_{0},\ell_{1},\ldots,\ell_{T}$ be arbitrary non-negative functions,
  $0<\gamma\le \beta<1$,
  and
  $F_{t}^{\gamma}(\w)=\sum_{s=0}^{t}\gamma^{t-s}\ell_{s}(\w)$.
  For all $t$, define
  \begin{align*}
  \bar d_{t}^{\beta}(u,v)=\frac{1}{\sum_{s=0}^{t}\beta^{t-s}}\sum_{s=0}^{t}\beta^{t-s}\sbrac{\ell_{s}(u)-\ell_{s}(v)}_{+}
  \end{align*}
  and let $P_{T}^{\beta}(\vec{\cmp})=\sum_{t=1}^{T-1}\bar{d}_{t}^{\beta}(\cmp_{\tpp},\cmp_{t})$.
  Then for any $V_{T}\ge 0$,
  \begin{align*}
    \gamma \sum_{t=1}^{T-1} \sbrac{F_{t}^{\gamma}(\cmp_{\tpp})-F_{t}^{\gamma}(\cmp_{t})}+\Log{\frac{1}{\gamma}}V_{T}
      \le\frac{\beta}{1-\beta} P_{T}^{\beta}(\vec{\cmp})+\frac{1-\gamma}{\gamma}V_{T}
  \end{align*}
\end{manuallemma}
\begin{proof}
  The first summation can be bounded as
  \begin{align*}
    \gamma \sum_{t=1}^{T-1} \sbrac{F_{t}^{\gamma}(\cmp_{t})-F_{t}^{\gamma}(\cmp_{\tmm})}
    &=
      \gamma\sum_{t=1}^{T-1}\sum_{s=0}^{t}\gamma^{t-s}\sbrac{\ell_{s}(\cmp_{\tpp})-\ell_{s}(\cmp_{t})}\\
    &\le
      \gamma\sum_{t=1}^{T-1}\sum_{s=0}^{t}\gamma^{t-s}\sbrac{\ell_{s}(\cmp_{\tpp})-\ell_{s}(\cmp_{t})}_{+}\\
    &\le
      \beta\sum_{t=1}^{T-1}\sum_{s=0}^{t}\frac{\sum_{s'=0}^{t}\beta^{t-s'}}{\sum_{s'=0}^{t}\beta^{t-s'}}\beta^{t-s}\sbrac{\ell_{s}(\cmp_{\tpp})-\ell_{s}(\cmp_{t})}_{+}\\
    &\le
      \frac{\beta}{1-\beta}\sum_{t=1}^{T-1}\sum_{s=0}^{t}\frac{\beta^{t-s}}{\sum_{s'=0}^{t}\beta^{t-s'}}\sbrac{\ell_{s}(\cmp_{\tpp})-\ell_{s}(\cmp_{t})}_{+}\\
    &=
      \frac{\beta}{1-\beta} P_{T}^{\beta}(\vec{\cmp}),
  \end{align*}
  where the last inequality uses
  $\sum_{s=0}^{t}\beta^{t-s}=\frac{1-\beta^{t+1}}{1-\beta}\le \frac{1}{1-\beta}$.
  Using this along with the elementary inequality
  $\Log{x}\le x-1$, for any $V_{T}\ge 0$ we have
  \begin{align*}
    \gamma \sum_{t=1}^{T-1} \sbrac{F_{t}^{\gamma}(\cmp_{t})-F_{t}^{\gamma}(\cmp_{\tmm})}
    +\Log{\frac{1}{\gamma}}V_{T}
    &\le
      \frac{\beta}{1-\beta} P_{T}^{\beta}(\vec{\cmp})+\brac{\frac{1}{\gamma}-1}V_{T}\\
    &=
      \frac{\beta}{1-\beta} P_{T}^{\beta}(\vec{\cmp})+\frac{1-\gamma}{\gamma}V_{T}
  \end{align*}

\end{proof}

\subsection{Existence of a Good Discount Factor}%
\label{app:good-discount-existence}
The following lemma establishes the existence of a discount factor that
will lead to favorable tuning of the $\gamma$-dependent terms in \Cref{lemma:discounted-terms}.
\begin{restatable}{lemma}{GoodDiscountExistence}\label{lemma:good-discount-existence}
   Let $\ell_{0},\ell_{1},\ldots$ be arbitrary non-negative functions,
   $V_{T}\ge 0$, denote
   $\bar d_{t}^{\gamma}(u,v)=\frac{\sum_{s=0}^{t}\gamma^{t-s}\sbrac{\ell_{s}(u)-\ell_{s}(v)}_{+}}{\sum_{s=0}^{t}\gamma^{t-s}}$
   for $\gamma\in[0,1]$,
   and let
   $ P_{T}^{\gamma}(\vec{\cmp})=\sum_{t=1}^{T-1}\bar d_{t}^{\gamma}(\cmp_{\tpp},\cmp_{t})$.
  Then there is a $\gamma^{*}\in [0,1]$ such that
  \begin{align*}
    \gamma^{*}=\frac{\sqrt{V_{T}}}{\sqrt{V_{T}}+\sqrt{ P_{T}^{\gamma^{*}}(\vec{\cmp})}}.
  \end{align*}
\end{restatable}
\begin{proof}
  First, notice that that any such $\gamma$ with the stated property must be in $[0,1]$ since
  \begin{align*}
    0\le \frac{\sqrt{V_{T}}}{\sqrt{V_{T}}+\sqrt{ P_{T}^{\gamma}(\vec{\cmp})}}\le \frac{\sqrt{V_{T}}}{\sqrt{V_{T}}}= 1.
  \end{align*}
  Next, observe that the condition can be equivalently expressed as follows:
  \begin{align*}
    \gamma
    &=\frac{\sqrt{V_{T}}}{\sqrt{V_{T}}+\sqrt{ P_{T}^{\gamma}(\vec{\cmp})}}\\
    \iff
    \sqrt{V_{T}}(1-\gamma)
    &=
      \gamma\sqrt{ P_{T}^{\gamma}(\vec{\cmp})}\\
    &=
      \gamma\sqrt{\sum_{t=1}^{T-1}\sum_{s=0}^{t}\frac{\gamma^{t-s}}{\sum_{s=0}^{t}\gamma^{t-s}}\sbrac{\ell_{s}(\cmp_{\tpp})-\ell_{s}(\cmp_{t})}_{+}}\\
    &=
    \gamma\sqrt{\sum_{t=1}^{T-1}\sum_{s=0}^{t}\frac{\gamma^{t-s}}{1-\gamma^{t+1}}(1-\gamma)\sbrac{\ell_{s}(\cmp_{\tpp})-\ell_{s}(\cmp_{t})}_{+}}\\
    \iff
    \sqrt{V_{T}(1-\gamma)} &= \gamma \sqrt{\sum_{t=1}^{T-1}\sum_{s=0}^{t}\frac{\gamma^{t-s}}{1-\gamma^{t+1}}\sbrac{\ell_{s}(\cmp_{\tpp})-\ell_{s}(\cmp_{t})}_{+}}.
  \end{align*}
  The quantity on the LHS begins at $\sqrt{V_{T}}$ (for $\gamma=0$) and then
  decreases to $0$ as a function of
  $\gamma$. Likewise, the RHS begins at 0 (for $\gamma=0$) and increases as a function of
  $\gamma$, approaching $\infty$ as
  $\gamma\to 1$. Hence, there must be some $\gamma\in[0,1]$ at which the two
  lines cross, and hence a
  $\gamma\in[0,1]$ which satisfies the above relation,
  so
  there is a
  $\gamma\in[0,1]$
  such that
  \begin{align*}
    \gamma=\frac{\sqrt{V_{T}}}{\sqrt{V_{T}}+\sqrt{ P_{T}^{\gamma}(\vec{\cmp})}}.
  \end{align*}
\end{proof}

\subsection{Proof of Theorem~\ref{thm:oracle-tuning-discounted-vaw}}%
\label{app:oracle-tuning-discounted-vaw}
Now combining everything we've seen in the previous  sections,
we can easily prove the following bound for the discounted VAW forecaster
under \emph{oracle tuning} of the discount factor.
\OracleTuningDiscountedVAW*
\begin{proof}
  \Cref{lemma:good-discount-existence} shows that for any sequence
  $\vec{\cmp}=(\cmp_{1},\ldots,\cmp_{T})$,
  there is a $\gamma^{*}\in[0,1]$ such that
  \begin{align*}
    \gamma^{*}=\frac{\sqrt{d\sumtT\half(\yt-\yhint_{t})^{2}}}{\sqrt{d\sumtT\half(\yt-\yhint_{t})^{2}}+\sqrt{P_{T}^{\gamma^{*}}(\vec{\cmp})}},
  \end{align*}
  so choosing $\gamma=\gamma^{*}$ and
  applying \Cref{thm:general-discounted-vaw}, we have
  \begin{align*}
    R_{T}(\vec{\cmp})
    &\le \frac{\gamma^{*}\lambda}{2}\norm{\cmp_{1}}^{2}_{2}+\frac{d}{2}\max_{t}(\yt-\yhint_{t})^{2}\Log{1+\frac{\sumtT \norm{\x_{t}}^{2}_{2}}{\lambda d}}\\
    &\qquad
    +\gamma^{*}\sum_{t=1}^{T-1}\sbrac{F_{t}^{\gamma^{*}}(\cmp_{\tpp})-F_{t}^{\gamma^{*}}(\cmp_{t})}
    +\frac{d}{2}\Log{1/\gamma^{*}}\sumtT(\yt-\yhint_{t})^{2}\\
    &\overset{(*)}{\le}
      \frac{\lambda}{2}\norm{\cmp_{1}}^{2}_{2}+\frac{d}{2}\max_{t}(\yt-\yhint_{t})^{2}\Log{1+\frac{\sumtT \norm{\x_{t}}^{2}_{2}}{\lambda d}}\\
    &\qquad
      +\frac{\gamma^{*}}{1-\gamma^{*}}P_{T}^{\gamma^{*}}(\vec{\cmp})
      +\frac{1-\gamma^{*}}{\gamma^{*}}\frac{d}{2}\sumtT(\yt-\yhint_{t})^{2}\\
    &=
      \frac{\lambda}{2}\norm{\cmp_{1}}^{2}_{2}+\frac{d}{2}\max_{t}(\yt-\yhint_{t})^{2}\Log{1+\frac{\sumtT \norm{\x_{t}}^{2}_{2}}{\lambda d}}+\sqrt{2d P_{T}^{\gamma^{*}}(\vec{\cmp})\sumtT(\yt-\yhint_{t})^{2}}\\
  \end{align*}
  where $(*)$ uses \Cref{lemma:discounted-terms} (with
  $\beta=\gamma=\gamma^{*}$). The stated result follows by hiding
  lower-order terms.
\end{proof}

\section{Proofs for Section~\ref{sec:small-loss} (\SecSmallLoss)}%
\label{app:small-loss}

\subsection{Proof of Theorem~\ref{thm:oracle-tuning-small-loss}}%
\label{app:oracle-tuning-small-loss}

We split the proof of \Cref{thm:oracle-tuning-small-loss} into
two parts. The following lemma, proven in \Cref{app:small-loss-template}, first
derives an initial regret template
that does most of the heavy lifting. We will later re-use this template
in the proof of \Cref{thm:grid-tuning-small-loss} to avoid repeating the argument.
The high-level intuition is that choosing hints
$\ytilde_{t}\approx\inner{\xt,\wt}$ leads to
$\sumtT(\yt-\yhint_{t})^{2}\approx\sumtT\ell_{t}(\wt)$, which leads to a
self-bounding argument that lets us replace $\sumtT(\yt-\yhint_{t})^{2}$ with
$\sumtT\ell_{t}(\cmp_{t})$ in the regret bound. We defer proof of the
lemma to the next subsection, \Cref{app:small-loss-template}.
\begin{restatable}{lemma}{SmallLossTemplate}\label{lemma:small-loss-template}
  Let $\yref_{t}\in\R$ be an arbitrary reference point, available at the start of
  round $t$, and  let
  $\cB_{t}=\Set{\y\in\R : \yref_{t}-M_{t}\le \y\le \yref_{t}+M_{t}}$ for
  $M_{t}=\max_{s<t}\abs{\y_{s}-\yref_{s}}$.
  Suppose that we apply \Cref{alg:discounted-vaw}
  with hints $\yhint_{t}=\yclip_{t}:=\Clip_{\cB_{t}}(\inner{\xt,\wt})$.
  Then for any sequence $\vec{\cmp}=(\cmp_{1},\ldots,\cmp_{T})$
  in $\R^{d}$ and any $\gamma,\beta\in[0,1]$ such that $\beta\ge \gamma\ge \gamma_{\min}=\frac{2d}{2d+1}$,
  \begin{align*}
    R_{T}(\vec{\cmp})
    &\le
      \gamma\lambda\norm{\cmp_{1}}^{2}_{2}+4d\max_{t}(\yt-\yref_{t})^{2}\Log{1+\frac{\sumtT\gamma^{T-t} \norm{\x_{t}}^{2}_{2}}{\lambda d}}\\
    &\qquad
    +2\frac{\beta}{1-\beta} P_{T}^{\beta}(\vec{\cmp})
      +\frac{1-\gamma}{\gamma}2d\sumtT\ell_{t}(\cmp_{t})
  \end{align*}
\end{restatable}
Now using this template, \Cref{thm:oracle-tuning-small-loss}
is easily proven by plugging in the stated discount factor $\gamma=\gamma^{\Ring}\maxOp\gamma_{\min}$
\OracleTuningSmallLoss*
\begin{proof}
  By \Cref{lemma:small-loss-template} (with $\beta=\gamma$), for any
  $\gamma\ge \gamma_{\min}=\frac{2d}{2d+1}$, we have
  \begin{align*}
    R_{T}(\vec{\cmp})
    &\le
      \gamma\lambda\norm{\cmp_{1}}^{2}_{2}+4d\max_{t}(\yt-\yref_{t})^{2}\Log{1+\frac{\sumtT\gamma^{T-t} \norm{\x_{t}}^{2}_{2}}{\lambda d}}\\
    &\qquad
    +2\frac{\gamma}{1-\gamma} P_{T}^{\gamma}(\vec{\cmp})
      +\frac{1-\gamma}{\gamma}2d\sumtT\ell_{t}(\cmp_{t}).
  \end{align*}
  Now by \Cref{lemma:good-discount-existence}, there is
  a
  $\gamma^{\Ring}\in[0,1]$
  satisfying
  $\gamma^{\Ring}=\frac{\sqrt{d\sumtT\ell_{t}(\cmp_{t})}}{\sqrt{d\sumtT\ell_{t}(\cmp_{t})}+\sqrt{ P_{T}^{\gamma^{\Ring}}(\vec{\cmp})}}$.
  If $\gamma^{\Ring}\ge \gamma_{\min}$, then for
  $\gamma=\gamma^{\Ring}\maxOp \gamma_{\min}$, the terms in the second line
  reduce to
  \begin{align*}
    2\frac{\gamma^{\Ring}}{1-\gamma^{\Ring}} P_{T}^{\gamma^{\Ring}}(\vec{\cmp})
      +\frac{1-\gamma^{\Ring}}{\gamma^{\Ring}}2d\sumtT\ell_{t}(\cmp_{t})
    &=
    4\sqrt{d P_{T}^{\gamma^{\Ring}}(\vec{\cmp})\sumtT\ell_{t}(\cmp_{t})},
  \end{align*}
  and otherwise for $\gamma^{\Ring}\le \gamma_{\min}$ we have
  \begin{align*}
    2\frac{\gamma_{\min}}{1-\gamma_{\min}} P_{T}^{\gamma_{\min}}(\vec{\cmp})
      +\frac{1-\gamma_{\min}}{\gamma_{\min}}2d\sumtT\ell_{t}(\cmp_{t})
    &\le
    2\frac{\gamma_{\min}}{1-\gamma_{\min}} P_{T}^{\gamma_{\min}}(\vec{\cmp})
      +\frac{1-\gamma^{\Ring}}{\gamma^{\Ring}}2d\sumtT\ell_{t}(\cmp_{t})\\
    &\le
    4d P_{T}^{\gamma_{\min}}(\vec{\cmp})
      +2\sqrt{d P_{T}^{\gamma^{\Ring}}(\vec{\cmp})\sumtT\ell_{t}(\cmp_{t})},
  \end{align*}
  so combining these two bounds and plugging back into
  the regret bound above, we have
  \begin{align*}
    R_{T}(\vec{\cmp})
    &\le
      \gamma\lambda\norm{\cmp_{1}}^{2}_{2}+4d\max_{t}(\yt-\yref_{t})^{2}\Log{1+\frac{\sumtT\gamma^{T-t} \norm{\x_{t}}^{2}_{2}}{\lambda d}}\\
    &\qquad
      +4d P_{T}^{\gamma_{\min}}(\vec{\cmp})+ 4\sqrt{d P_{T}^{\gamma^{\Ring}}(\vec{\cmp})\sumtT\ell_{t}(\cmp_{t})}\\
    &\le
      O\brac{
      d P_{T}^{\gamma_{\min}}(\vec{\cmp})+d\max_{t}(\yt-\yref_{t})^{2}\Log{T}+ \sqrt{d P_{T}^{\gamma^{\Ring}}(\vec{\cmp})\sumtT\ell_{t}(\cmp_{t})}.
      }
  \end{align*}

\end{proof}

\subsubsection{Proof of Lemma~\ref{lemma:small-loss-template}}%
\label{app:small-loss-template}
\SmallLossTemplate*
\begin{proof}
  Applying \Cref{thm:general-discounted-vaw} followed by \Cref{lemma:discounted-terms}, for any $\gamma\in(0,1]$ and
  $\beta\ge \gamma$ we have
  \begin{align*}
    R_{T}^{\cA_{\gamma}}(\vec{\cmp})
    &\le
      \frac{\gamma\lambda}{2}\norm{\cmp_{1}}^{2}_{2}+\frac{d}{2}\max_{t}(\yt-\yclip_{t})^{2}\Log{1+\frac{\sumtT\gamma^{T-t} \norm{\x_{t}}^{2}_{2}}{\lambda d}}\\
    &\qquad
    +\gamma\sum_{t=1}^{T-1}\sbrac{F_{t}^{\gamma}(\cmp_{\tpp})-F_{t}^{\gamma}(\cmp_{t})}
      +\frac{d}{2}\Log{1/\gamma}\sumtT(\yt-\yclip_{t})^{2}\\
    &\le
      \frac{\gamma\lambda}{2}\norm{\cmp_{1}}^{2}_{2}+\frac{d}{2}\max_{t}(\yt-\yclip_{t})^{2}\Log{1+\frac{\sumtT\gamma^{T-t} \norm{\x_{t}}^{2}_{2}}{\lambda d}}\\
    &\qquad
    +\frac{\beta}{1-\beta} P_{T}^{\beta}(\vec{\cmp})
      +\frac{1-\gamma}{\gamma}\frac{d}{2}\sumtT(\yt-\yclip_{t})^{2},
  \end{align*}
  Using \Cref{lemma:clipped-predictions} we have
  \begin{align*}
    \sumtT(\yt-\yclip_{t})^{2}
    &\le
      \sumtT\sbrac{ M_{t+1}^{2}-M_{t}^{2} + 2\ell_{t}(\wt)}
      \le
      M_{T+1}^{2}+2\sumtT\ell_{t}(\wt),
  \end{align*}
  so for any $\gamma\ge \frac{2d}{2d+1}$, we have
  \begin{align*}
    \frac{1-\gamma}{\gamma}\frac{d}{2}\sumtT(\yt-\yclip_{t})^{2}
    &\le
    \frac{1-\gamma}{\gamma}d\sbrac{\half M_{T+1}^{2}+\sumtT\ell_{t}(\wt)}\\
    &=
    \frac{1-\gamma}{\gamma}d\sbrac{\half M_{T+1}^{2}+\sumtT\ell_{t}(\wt)-\ell_{t}(\cmp_{t})+\sumtT\ell_{t}(\cmp_{t})}\\
    &\le
      \frac{1}{4}M_{T+1}^{2}+\half\sumtT\ell_{t}(\wt)-\ell_{t}(\cmp_{t})+\frac{1-\gamma}{\gamma}d\sumtT\ell_{t}(\cmp_{t}),
  \end{align*}
  where the final inequality uses
  $\gamma\ge\frac{2d}{2d+1}\implies \frac{1-\gamma}{\gamma}\le\frac{1}{2d}$ and
  bounds
  $\frac{1-\gamma}{\gamma}d\sumtT\ell_{t}(\wt)-\ell_{t}(\cmp_{t})\le \half \sumtT\ell_{t}(\wt)-\ell_{t}(\cmp_{t})$
  (assuming $\sumtT\ell_{t}(\wt)-\ell_{t}(\cmp_{t})\ge 0$, which can be assumed
  without loss of generality since otherwise the stated bound trivially holds).
  Plugging this back into the regret bound and re-arranging terms, we have
  \begin{align*}
    R_{T}(\vec{\cmp})
    &\le
      \frac{\gamma\lambda}{2}\norm{\cmp_{1}}^{2}_{2}+\frac{d}{2}\max_{t}(\yt-\yclip_{t})^{2}\Log{1+\frac{\sumtT\gamma^{T-t} \norm{\x_{t}}^{2}_{2}}{\lambda d}}\\
    &\qquad
    +\frac{\gamma}{1-\gamma} P_{T}^{\gamma}(\vec{\cmp})
      +\half R_{T}(\vec{\cmp})+\frac{1-\gamma}{\gamma}\sumtT\ell_{t}(\cmp_{t})\\
    \implies
      R_{T}(\vec{\cmp})
    &\le
      \gamma\lambda\norm{\cmp_{1}}^{2}_{2}+4d\max_{t}(\yt-\yref_{t})^{2}\Log{1+\frac{\sumtT\gamma^{T-t} \norm{\x_{t}}^{2}_{2}}{\lambda d}}\\
    &\qquad
    +2\frac{\beta}{1-\beta} P_{T}^{\beta}(\vec{\cmp})
      +\frac{1-\gamma}{\gamma}2d\sumtT\ell_{t}(\cmp_{t}),
  \end{align*}
  where we've bounded
  $\max_{t}(\yt-\yclip_{t})^{2}\le 4M_{T+1}=4\max_{t}(\yt-\yref_{t})^{2}$ using \Cref{lemma:clipped-predictions}.
\end{proof}

\section{Proofs for Section~\ref{sec:lb} (\SecLB)}%
\label{app:lb}

\subsection{Proof of Theorem~\ref{thm:tight-lb}}%
\label{app:tight-lb}
\TightLB*
\begin{proof}
  First notice that the
  trivial comparator sequence with $\cmp_{1}=\ldots=\cmp_{T}$
  always satisfies $\sum_{t=2}^{T}\max_{s}\sbrac{\ell_{s}(\cmp_{\tpp})-\ell_{s}(\cmp_{t})}_{+}=0\le P$,
  so we can always lower-bound the dynamic regret
  using the well-known lower bound for
  the static regret in this setting (see, \eg{}, \citet{vovk2001competitive,
    gaillard2019uniform,mayo2022scalefree}).
  In particular, for any $\cmp\in W$
  we have
  \begin{align}
    \sup_{\y_{1},\ldots,\y_{T}}R_{T}(\cmp)
    &\ge
      \Omega\brac{dY^{2}\Log{T}}\label{eq:tight-lb:static}
  \end{align}

  Next,
  let $\sigma\in [0,1]$
  and let $\sigma_{1},\ldots,\sigma_{t}$ be a sequence of
  iid random variables drawn uniformly from $\Set{-\sigma, \sigma}$,
  and let $\yt=Y\sigma_{t}$.
  Choose feature vectors $\xt$ which cycle through the
  standard basis vectors (\eg{} define
  $\iota(t)=t\Mod{d}+1$ and let $\xt = e_{\iota(t)}$).
  Now observe that the comparator sequence
  can always exactly fit a sequence $\y_{1},\ldots,\y_{T}$
  by setting $\cmp_{t}$ to satisfy $\inner{\xt,\cmp_{t}}=\cmp_{t,\iota(t)}=\yt$.
  In particular, by letting $\tilde \cmp_{1}=(\y_{1},\ldots,\y_{d})$,
  $\tilde \cmp_{2}=(\y_{d+1},\ldots,\y_{2d}),\ldots,\tilde\cmp_{\Ceil{T/d}}=(y_{\Ceil{T/d}+1},\ldots,y_{T},0,0,\ldots)$
  we can
  set $\cmp_{t}=\tilde\cmp_{\Ceil{t/d}}$ to guarantee $\inner{\xt,\cmp_{t}}=\yt$ on
  all rounds, while only changing the comparator $\Ceil{T/d}$ times at most.
  From this, we have the following initial bound on the regret:
  \begin{align}
    \sup_{\y_{1},\ldots,\y_{T}}R_{T}(\vec{\cmp})
    &\ge
      \E_{\vec{y}}\sbrac{\sumtT \ell_{t}(\wt)-\ell_{t}(\cmp_{t})}\nonumber\\
    &\ge
      \E_{\vec{y}}\sbrac{\half\yt^{2}+\half\inner{\xt,\wt}^{2}+\yt\inner{\xt,\wt}}\nonumber\\
    &\ge
      \half \sigma^{2}Y^{2}T,\label{eq:tight-lb:initial-bound}
  \end{align}
  where the last line uses $\yt^{2}=Y^{2}\sigma^{2}$ and $\EE{\yt}=0$.
  Moreover,
  since the comparator changes
  only every $d$ rounds, the variability is bounded as
  \begin{align*}
    \sum_{t=1}^{T-1}\max_{s}\sbrac{\ell_{s}(\cmp_{\tpp})-\ell_{s}(\cmp_{t})}_{+}
    &\le
      \sum_{i=1}^{\Ceil{T/d}-1}\max_{s}\sbrac{\ell_{s}(\tilde\cmp_{i+1})-\ell_{s}(\tilde\cmp_{i})}_{+}.
  \end{align*}
  Observe that
  $\ell_{s}(\tilde\cmp_{i+1})-\ell_{s}(\tilde\cmp_{i})$ can only be
  positive when $\inner{\x_{s},\tilde\cmp_{i}}=\y_{s}$ and
  $\inner{\x_{s},\tilde\cmp_{i+1}}=-\y_{s}$, hence
  \begin{align*}
    \sum_{t=1}^{T-1}\max_{s}\sbrac{\ell_{s}(\cmp_{\tpp})-\ell_{s}(\cmp_{t})}_{+}
    &\le
      \sum_{i=1}^{\Ceil{T/d}-1}\max_{s}\sbrac{\ell_{s}(\tilde\cmp_{i+1})-\ell_{s}(\tilde\cmp_{i})}_{+}\\
    &\le
      \sum_{i=1}^{\Ceil{T/d}-1}\half (\y_{s}-(-\y_{s}))^{2}\\
    &\le
    \frac{2TY^{2}}{d}\sigma^{2}.
  \end{align*}
  Hence, setting $\sigma=\sqrt{\frac{dP}{2TY^{2}}}\le 1$ ensures
  $\sum_{t=1}^{T-1}\max_{s}\sbrac{\ell_{s}(\cmp_{\tpp})-\ell_{s}(\cmp_{t})}_{+}\le \frac{2TY^{2}}{d}\sigma^{2}\le P$,
  and the regret is bounded below by
  \begin{align}
    \sup_{\y_{1},\ldots,\y_{T}}R_{T}(\vec{\cmp})
    &\ge
      \half \sigma^{2}Y^{2}T
      =
      \frac{1}{4}dP,\nonumber\\
    \intertext{which we can further lower bound as:}
      &=
        \frac{1}{4}\sqrt{dP \cdot dP}
        \ge
        \frac{1}{4}\sqrt{dP\cdot d\sum_{t=1}^{T-1}\max_{s}\sbrac{\ell_{s}(\cmp_{\tpp})-\ell_{s}(\cmp_{t})}_{+}}\nonumber\\
    &\ge
          \frac{1}{4}\sqrt{dP\sum_{t=1}^{T-1}\sbrac{\ell_{t}(\cmp_{\tpp})-\ell_{t}(\cmp_{t})}_{+}}
      =
      \Omega\brac{\sqrt{dP\sum_{t=2}^{T}\half(\yt-\ytmm)^{2}}}.\label{eq:tight-lb:non-trivial}
  \end{align}
  Taken together with \Cref{eq:tight-lb:static}, we have
  \begin{align*}
    R_{T}(\vec{\cmp})
    &\ge
      \Omega\brac{dY^{2}\Log{T}\maxOp \sqrt{dP\cV_{T}}}
  \end{align*}
  where $\cV_{T}= dP\maxOp \sum_{t=2}^{T}\half(\yt-\ytmm)^{2}$.

\end{proof}

\section{Proofs for Section~\ref{sec:tuning} (\SecTuning)}%
\label{app:tuning}

\subsection{Proof of Lemma~\ref{lemma:clipped-predictions}}%
\label{app:clipped-predictions}
The following lemma shows that by clipping our predictions to some crude
``trust-region'', the loss of the clipped prediction is at worst prortional
to the maximal deviation of the true $\yt$ from the trust region.
Intuitively, we can think of $\yref$ as being some data-dependent
but already-observed quantity, such as $\ytmm$.
\begin{restatable}{lemma}{ClippedPredictions}\label{lemma:clipped-predictions}
  Define $M_{t}=\max_{s<t}\abs{\y_{s}-\yref_{s}}$,
  $\cB_{t}=\Set{x\in\R: \yref_{t}-M_{t}\le x\le \yref_{t}+M_{t}}$, and
  let
  $\yclip_{t}=\Clip_{\cB_{t}}(\inner{\xt,\wt})$ for some $\wt\in\R^{d}$.
  Then for any $t$ we have
  \begin{align*}
    (\yt-\yclip_{t})^{2}\le \Min{4M_{\tpp}^{2},2\ell_{t}(\wt)+M_{\tpp}^{2}-M_{t}^{2}}
  \end{align*}
\end{restatable}
\begin{proof}
  First, observe that we always have
  \begin{align*}
    \brac{\yt-\yclip_{t}}^{2}
    &=
      \brac{\yt-\yref_{t}+\yref_{t}-\yclip_{t}}^{2}
      \le
      2\brac{\yt-\yref_{t}}^{2}+2\brac{\yref_{t}-\yclip_{t}}^{2}
    \le
      2M_{\tpp}^{2}+2M_{t}^{2}\le 4M_{\tpp}^{2}.
  \end{align*}
  Next, observe that if $\inner{\xt,\wt}=\yclip_{t}$, then we trivially have
  $(\yt-\yclip_{t})^{2}=(\yt-\inner{\xt,\wt})^{2}=2\ell_{t}(\wt)$.
  Otherwise, when $\inner{\xt,\wt}\ne \yclip_{t}$, we have clipped
  $\yclip_{t}$ to be a distance of $M_{t}$ away from $\yref_{t}$ and there are two
  cases to consider.
  If $\Sign{\yclip_{t}-\yref_{t}}\ne \Sign{\yt-\yref_{t}}$, then
  the clipping operation $\yclip_{t}=\Clip_{\cB_{t}}(\inner{\xt,\wt})$ moves
  us closer
  to $\yt$, hence
  $\abs{\yt-\yclip_{t}}\le \abs{\yt-\inner{\xt,\wt}}$.
  If $\Sign{\yclip_{t}-\yref_{t}}=\Sign{\yt-\yref_{t}}$, then we precisely have
  $\abs{\yt-\yclip_{t}}=M_{\tpp}-M_{t}$ when $\yt\notin\cB_{t}$ and
  $\abs{\yt-\yclip_{t}}\le \abs{\yt-\inner{\xt,\wt}}$ when $\yt\in \cB_{t}$.
  Hence, combining these cases we have
  \begin{align*}
    (\yt-\yclip_{t})^{2}\le (\yt-\inner{\wt,\xt})^{2} + (M_{\tpp}-M_{t})^{2}
    &\le2\ell_{t}(\wt)+M_{\tpp}^{2}-M_{t}^{2},
  \end{align*}
  where we have used $(u-l)^{2}\le u^{2}-l^{2}$ for $u\ge l\ge 0$.
  Hence, combining with the first display we have
  \begin{align*}
    (\yt-\yclip_{t})^{2}\le \Min{4M_{\tpp}^{2}, M_{\tpp}^{2}-M_{t}^{2}+2\ell_{t}(\wt)}.
  \end{align*}
\end{proof}

\subsection{Proof of Lemma~\ref{lemma:squared-loss-exp-concave}}%
\label{app:squared-loss-exp-concave}

The following lemma shows the following important property
of the meta-learner's losses:
they are $\alpha_{t}$-exp-concave with
$\alpha_{t}=\frac{1}{2\max_{i}\ell_{t}(\y_{t}^{(i)})}$
in the domain $\hat\cY_{t}=\Set{\y=\sum_{i=1}^{N}p_{i}\y_{t}^{(i)}: \sum_{i=1}^{N}p_{i}=1}$.
\begin{restatable}{lemma}{SquaredLossExpConcave}\label{lemma:squared-loss-exp-concave}
  Let $\y^{(1)},\ldots, \y^{(N)}$ be arbitrary real numbers and let
  $\hat\cY_{t}=\Set{\ypred=\sum_{i=1}^{N}p_{i}\y^{(i)}: p\in \R_{\ge0}^{N}, \sum_{i=1}^{N}p_{i}=1}$.
  Then $\ell_{t}(\ypred)=\half(\yt-\ypred)^{2}$ is $\alpha_{t}$-Exp-Concave
  on $\hat\cY_{t}$ for $\alpha_{t}\le \frac{1}{2\max_{i}\ell_{t}\brac{\y^{(i)}}}$.
\end{restatable}

\begin{proof}
Letting $f_{t}(\ypred)=\Exp{-\alpha_{t}\ell_{t}(\ypred)}$ we have
for any $\ypred\in\hat\cY_{t}$:
\begin{align*}
  f_{t}'(\ypred)
  &=
    \sbrac{\Exp{-\frac{\alpha_{t}}{2}(\yt-\ypred)^{2}}}'
    =
    \Exp{-\frac{\alpha_{t}}{2}(\yt-\ypred)^{2}}\alpha_{t}(\yt-\ypred)\\
  f_{t}''(\ypred)
  &=
    \Exp{-\frac{\alpha_{t}}{2}(\yt-\ypred)^{2}}\sbrac{\alpha_{t}^{2}(\yt-\ypred)^{2}-\alpha_{t}}\\
  &=
    \Exp{-\frac{\alpha_{t}}{2}(\yt-\ypred)^{2}}\sbrac{2\alpha_{t}^{2}\ell_{t}(\ypred)-\alpha_{t}}
\end{align*}
Hence for $\alpha_{t}\le\frac{1}{2\max_{i} \ell_{t}(\y^{(i)})}$ we have
\begin{align*}
  f_{t}''(\ypred)
  &\le
    \Exp{-\frac{\alpha_{t}}{2}(\yt-\ypred)^{2}}\alpha_{t}\sbrac{2\alpha_{t}\ell_{t}(\ypred)-1}
  \le 0
\end{align*}
so $f_{t}(\ypred)=\Exp{-\alpha_{t}\ell_{t}(\ypred)}$ is concave and $\ell_{t}$ is
$\alpha_{t}$-Exp-Concave over $\hat\cY$ for $\alpha_{t}\le\frac{1}{2\max_{i}\ell_{t}(\y^{(i)})}$.
\end{proof}

\subsection{Regret of the Range-Clipped Meta-Algorithm}
\label{app:exp-concave-experts-redux}
In this section we prove a simple result showing that
the range-clipping reduction described by \Cref{alg:clipped-meta} incurs only an
constant additional penalty. This lemma will
be used to do most of the heavy-lifting in proving
\Cref{thm:dynamic-regret-tuning}, which simply applies the following lemma
and then chooses a specific meta-algorithm for $\cA_{\Meta}$.
\begin{restatable}{lemma}{ExpConcaveExpertsRedux}\label{lemma:exp-concave-experts-redux}
  For any $[a,b]\subseteq[1,T]$, sequence
  $\vec{\cmp}=(\cmp_{a},\ldots,\cmp_{b})$ in $\R$,
  and $j\in[N]$, \Cref{alg:clipped-meta} guarantees
  \begin{align*}
    R_{[a,b]}(\vec{\cmp})
    &\le
      \half \max_{t}(\yt-\yref_{t})^{2} +R_{[a,b]}^{\cA_{j}}(\vec{\cmp})+R_{[a,b]}^{\Meta}(e_{j}),
  \end{align*}
  where $R_{[a,b]}^{\cA_{j}}(\vec{\cmp})=\sum_{t=a}^{b}\ell_{t}(\wt^{(j)})-\ell_{t}(\cmp_{t})$ is
  the dynamic regret $\cA_{j}$
  and
  $R_{[a,b]}^{\Meta}(e_{j})=\sum_{t=a}^{b}\ell_{t}(\ypred_{t})-\ell_{t}(\yclip_{t}^{(j)})$.
\end{restatable}
\begin{proof}
  For ease of notation we let $\yt^{(i)}=\inner{\xt,\wt^{(i)}}$, where
  $\wt^{(i)}$ is the output of algorithm $\cA_{i}$, and slightly abuse notation
  by writing $\ell_{t}(\y)=\half(\yt-\y)^{2}$ for $\y\in\R$. Hence, we may
  write $\ell_{t}(\wt)\equiv\ell_{t}(\yt^{(i)})$ interchangeably. Note that this
  equivalence is valid in the improper online learning setting
  since the features are observed \emph{before} the learner makes a prediction,
  as discussed in the introduction.

  Now, for for any $j\in[N]$  we have
  \begin{align*}
    R_{[a,b]}(\vec{\cmp})
    &=
      \sum_{t=a}^{b} \ell_{t}(\ypred_{t})-\ell_{t}(\cmp_{t})\\
      &=
      \sum_{t=a}^{b} \ell_{t}(\wt^{(j)})-\ell_{t}(\cmp_{t})
      +\sum_{t=a}^{b} \ell_{t}(\ypred_{t})-\ell_{t}(\wt^{(j)})\\
    &=
      R_{[a,b]}^{\cA_{j}}(\vec{\cmp})+\sum_{t=a}^{b} \ell_{t}(\ypred_{t})-\ell_{t}\brac{\yt^{(j)}},
  \end{align*}
  where we have observed $\yt^{(j)}=\inner{\xt,\wt^{(j)}}$.
  Observe that by \Cref{lemma:clipped-predictions} we have
  \begin{align*}
    \ell_{t}(\yt^{(j)})
    &\ge
      \half M_{t}^{2}-\half M_{\tpp}^{2} + \half(\yt-\yclip_{t}^{(j)})^{2}\\
    &=
      \half M_{t}^{2}-\half M_{\tpp}^{2} + \ell_{t}\brac{\yclip_{t}^{(j)}},
  \end{align*}
  where $M_{t}=\max_{s<t}\abs{\y_{s}-\yref_{s}}$.
  Hence,
  \begin{align*}
    R_{[a,b]}(\vec{\cmp})
    &\le
      R_{[a,b]}^{\cA_{j}}(\vec{\cmp}) + \sum_{t=a}^{b} \ell_{t}(\ypred_{t})-\ell_{t}\brac{\yt^{(j)}}\\
    &\le
      R_{[a,b]}^{\cA_{j}}(\vec{\cmp}) + \sum_{t=a}^{b} \ell_{t}(\ypred_{t})-\ell_{t}\brac{\yclip_{t}^{(j)}}\\
    &\qquad+
      \sum_{t=a}^{b}\half M_{\tpp}^{2}-\half M_{t}^{2}\\
    &\le
      \half M_{b+1}^{2}+R_{[a,b]}^{\cA_{j}}(\vec{\cmp}) + \underbrace{\sum_{t=a}^{b} \ell_{t}(\ypred_{t})-\ell_{t}\brac{\yclip_{t}^{(j)}}}_{=:R_{[a,b]}^{\Meta}(e_{j})}
  \end{align*}
\end{proof}

\subsection{Proof of Theorem~\ref{thm:dynamic-regret-tuning}}%
\label{app:dynamic-regret-tuning}

\begin{minipage}{\columnwidth}
\begin{manualtheorem}{\ref*{thm:dynamic-regret-tuning}}
  Let $\cA_{\Meta}$ be an instance of \Cref{alg:adaptive-fixed-share}
  with $\alpha_{t}=\frac{1}{2\max_{t,i}\ell_{t}(\yclip_{t}^{(i)})}$, $\beta_{t+1}=\frac{1}{(e+t)\log^{2}(e+t)+1}$ and $p_{1}=\mathbf{1}_{N}/N$.
  Then
  for any sequence $\vec{\cmp}=\brac{\cmp_{1},\ldots,\cmp_{T}}$ in $\R$ and any $j\in[N]$,
  \Cref{alg:clipped-meta} guarantees
  \begin{align*}
    R_{[a,b]}(\vec{\cmp})
    &\le
      O\brac{ R_{[a,b]}^{\cA_{j}}(\vec{\cmp})+\max_{t}(\y_{t}-\ytilde_{t})^{2}\Log{Nb\log^{2}(b)}},
  \end{align*}
  where $R_{[a,b]}$ denotes regret over the sub-interval $[a,b]$.
\end{manualtheorem}
\end{minipage}

\begin{proof}
  The proof follows almost immediately using the
  regret guarantee of the range-clipped meta-algorithm
  (\Cref{lemma:exp-concave-experts-redux}),
  from which we have
  \begin{align*}
    R_{[a,b]}(\vec{\cmp})
    &\le
      \half\max_{t} (\yt-\ytilde_{t})^{2}+ R_{[a,b]}^{\cA_{j}}(\vec{\cmp})+R_{[a,b]}^{\Meta}(e_{j}).
  \end{align*}
  Now
  applying
  the guarantee of an appropriate instance of the
  fixed-share algorithm
  (\Cref{thm:adaptive-fixed-share}
  with $\alpha_{t}=\frac{1}{2\max_{t,i}\ell_{t}(\yclip_{t}^{(i)})}$,
  $\beta_{t}=\frac{1}{(e+t)\log^{2}(e+t)+1}$, and $p_{1}=\mathbf{1}_{N}/N$), we have
  \begin{align*}
    R_{[a,b]}^{\Meta}(e_{j})
    &\le
      \frac{1}{\alpha_{b+1}}\sbrac{2\Log{\frac{1}{\beta_{b+1}p_{1j}}}+1}\\
    &\le
      \max_{t,i}\ell_{t}(\yclip_{t}^{(i)})\sbrac{2\Log{((e+b)\log^{2}(e+b)+1)N}+1}\\
    &\le
      O\brac{\max_{t}(\yt-\ytilde_{t})^{2}\Log{b\log^{2}(b)N}},
  \end{align*}
  where the last line applies \Cref{lemma:clipped-predictions} and hides
  constants.
  All together, we have
  \begin{align*}
    R_{[a,b]}(\vec{\cmp})
    &\le
      O\brac{R_{[a,b]}^{\cA_{j}}(\vec{\cmp})+\max_{t}(\yt-\ytilde_{t})^{2}\Log{Nb\log^{2}(b)}}.
  \end{align*}
\end{proof}

\subsection{Proof of Theorem~\ref{thm:grid-tuning-discounted-vaw}}%
\label{app:grid-tuning-discounted-vaw}
The
proof of \Cref{thm:grid-tuning-discounted-vaw} follows by applying \Cref{thm:dynamic-regret-tuning},
and then showing that there exists a $\cA_{\gamma}$ which
attains the desired bound.
We first provide proof of the latter claim in
\Cref{lemma:grid-tuning-discounted-vaw-existence}
for the sake of modularity. In particular, we will also re-use
this result to argue strongly-adaptive guarantees in
\Cref{sec:strongly-adaptive}. Proof of \Cref{thm:grid-tuning-discounted-vaw}
is then easily proven at the end of this section.
\begin{restatable}{lemma}{GridTuningDiscountedVAWExistence}\label{lemma:grid-tuning-discounted-vaw-existence}
  Let $b>1$, $\eta_{\min}=2d$, $\eta_{\max}=dT$, and define
  $\cS_{\eta}=\Set{\eta_{i}=\eta_{\min}b^{i}\minOp \eta_{\max}:i=0,1,\ldots}$
  and
  $\cS_{\gamma}=\Set{\gamma_{i}=\frac{\eta_{i}}{1+\eta_{i}}: i=0,1,\ldots}\cup\Set{0}$.
  For any $\gamma$ in $\cS_{\gamma}$, let $\cA_{\gamma}$ denote an instance of \Cref{alg:discounted-vaw}
  with discount $\gamma$.
  Then for any $\vec{\cmp}=(\cmp_{1},\ldots,\cmp_{T})$ in $\R^{d}$,
  there is a $\gamma^{*}\in[0,1]$ satisfying
  $\gamma^{*}=\frac{\sqrt{d\sumtT\half(\yt-\yhint_{t})^{2}}}{\sqrt{d\sumtT\half(\yt-\yhint_{t})^{2}}+\sqrt{P_{T}^{\gamma^{*}}(\vec{\cmp})}}$
  and a $\gamma\in \cS_{\gamma}$ such that
  \begin{align*}
    R_{T}^{\cA_{\gamma}}(\vec{\cmp})
    &\le
      O\Bigg(d\max_{t}(\yt-\yhint_{t})^{2}\Log{T}
      + b\sqrt{d P_{T}^{\gamma^{*}}(\vec{\cmp})\sumtT (\yt-\yhint_{t})^{2}}\Bigg).
  \end{align*}
\end{restatable}
\begin{proof}
  Denote
  $V_{T}=\frac{d}{2}\sumtT(\yt-\yhint_{t})^{2}$. By
  \Cref{lemma:good-discount-existence},
  there exists a $\gamma^{*}\in[0,1]$ such that
  \begin{align*}
    \gamma^{*}=\frac{\sqrt{V_{T}}}{\sqrt{V_{T}}+\sqrt{ P_{T}^{\gamma^{*}}(\vec{\cmp})}}.
  \end{align*}
  Throughout the proof it will be convenient to work in terms of the related
  quantity $\eta^{*}=\frac{\gamma^{*}}{1-\gamma^{*}}=\sqrt{\frac{V_{T}}{ P_{T}^{\gamma}(\vec{\cmp})}}$. Let us first suppose that
  $0\le \eta^{*}\le \eta_{\min}$. In this case, we have
  \begin{align*}
    \eta^{*}
    &=
      \sqrt{\frac{V_{T}}{ P_{T}^{\gamma^{*}}(\vec{\cmp})}}\le \eta_{\min}
    \implies
      \sqrt{\half\sumtT(\yt-\yhint_{t})^{2}}\le \eta_{\min}\sqrt{\frac{1}{d} P_{T}^{\gamma}(\vec{\cmp})}.
  \end{align*}
  Consider the algorithm $\cA_{0}$ with $\gamma=0$: in this case we have
  $\wt=\argmin_{\w\in \R^{d}}h_{t}(\w)$, so $\inner{\xt,\wt}=\yhint_{t}$
  and the regret is trivially
  \begin{align}
    \sumtT\ell_{t}(\wt^{\cA_{0}})-\ell_{t}(\cmp_{t})
    &\le
      \sumtT\half(\yt-\yhint_{t})^{2}\nonumber\\
    &=
      \sqrt{\sumtT\half(\yt-\yhint_{t})^{2}\sumtT \half(\yt-\yhint_{t})^{2}}\nonumber\\
    &\le
      \frac{\eta_{\min}}{\sqrt{d}}\sqrt{ P_{T}^{\gamma^{*}}(\vec{\cmp})\sumtT\half (\yt-\yhint_{t})^{2}}\nonumber\\
    &=
      2\sqrt{V_{T} P_{T}^{\gamma^{*}}(\vec{\cmp})}
  \end{align}
  for $\eta_{\min}=2d$.

  Otherwise, for $\eta^{*}\ge \eta_{\min}$,
  using
  \Cref{thm:general-discounted-vaw}
  we have that for any $\gamma\in\cS_{\gamma}$,
  \begin{align*}
    R_{T}^{\cA_{\gamma}}(\vec{\cmp})
    &\le
      \frac{\gamma\lambda}{2}\norm{\cmp_{1}}^{2}_{2}+\frac{d}{2}\max_{t}(\yt-\yhint_{t})^{2}\Log{1+\frac{\sumtT \gamma^{T-t}\norm{\x_{t}}^{2}_{2}}{\lambda d}}\\
    &\qquad
    +\gamma\sum_{t=1}^{T-1}\sbrac{F_{t}^{\gamma}(\cmp_{\tpp})-F_{t}^{\gamma}(\cmp_{t})}
    +\Log{1/\gamma}V_{T}\\
    &\overset{(*)}{\le}
      \frac{\gamma\lambda}{2}\norm{\cmp_{1}}^{2}_{2}+\frac{d}{2}\max_{t}(\yt-\yhint_{t})^{2}\Log{1+\frac{\sumtT \gamma^{T-t}\norm{\x_{t}}^{2}_{2}}{\lambda d}}\\
    &\qquad
    +\eta^{*}P_{T}^{\gamma^{*}}(\vec{\cmp})
    +\frac{V_{T}}{\eta}\\
  \end{align*}
  where $(*)$ observes that
  $\eta_{\min}=\frac{\gamma_{\min}}{1-\gamma_{\min}}\le \eta^{*}=\frac{\gamma^{*}}{1-\gamma^{*}}\implies \gamma_{\min}\le \gamma^{*}$
  and applies \Cref{lemma:discounted-terms} (with $\beta=\gamma^{*}$)
  and substitutes $\eta=\frac{\gamma}{1-\gamma}$.
  If
  $\eta^{*}\ge \eta_{\max}$ then
  choosing $\eta=\eta_{\max}=dT$ yields
  \begin{align*}
    \frac{V_{T}}{\eta}=\frac{d}{2dT}\sumtT (\yt-\yhint_{t})^{2}\le \half\max_{t}(\yt-\yhint_{t})^{2},
  \end{align*}
  and otherwise, there is an $\eta_{k}$ in $\cS_{\eta}$ such that
  $\eta_{k}\le \eta^{*}\le b\eta_{k}$, so choosing $\eta=\eta_{k}$ yields
  \begin{align*}
    \frac{V_{T}}{\eta_{k}}\le b\frac{V_{T}}{\eta^{*}}=b\sqrt{P_{T}^{\gamma^{*}}(\vec{\cmp})V_{T}}
  \end{align*}
  Hence, overall we have that there is a $\gamma\in\cS_{\gamma}$ such that
  \begin{align*}
    R_{T}^{\cA_{\gamma}}(\vec{\cmp})
    &\le
      \frac{\gamma\lambda}{2}\norm{\cmp_{1}}^{2}_{2}+\half\max_{t}(\yt-\yhint_{t})^{2}\sbrac{d\Log{1+\frac{\sumtT \gamma^{T-t}\norm{\x_{t}}^{2}_{2}}{\lambda d}}\maxOp 1}
    +\eta^{*}P_{T}^{\gamma^{*}}(\vec{\cmp})
    +b\frac{V_{T}}{\eta^{*}}\\
    &=
      \frac{\gamma\lambda}{2}\norm{\cmp_{1}}^{2}_{2}+\half\max_{t}(\yt-\yhint_{t})^{2}\sbrac{d\Log{1+\frac{\sumtT \gamma^{T-t}\norm{\x_{t}}^{2}_{2}}{\lambda d}}\maxOp 1}
      +(b+1)\sqrt{V_{T} P_{T}^{\gamma^{*}}(\vec{\cmp})}\\
    &\le
      O\brac{d\max_{t}(\yt-\yhint_{t})^{2}\Log{T}\maxOp b\sqrt{d P_{T}^{\gamma^{*}}(\vec{\cmp})\sumtT (\yt-\yhint_{t})^{2}}}.
  \end{align*}
\end{proof}

With the previous lemma in hand, the proof
of \Cref{thm:grid-tuning-discounted-vaw} follows easily. The theorem is
re-stated for convenience.
\GridTuningDiscountedVAW*
\begin{proof}
Applying \Cref{thm:dynamic-regret-tuning}, for any sequence
$\vec{\cmp}=(\cmp_{1},\ldots,\cmp_{T})$ in $\R^{d}$ and any $\gamma\in\cS_{\gamma}$ we have
\begin{align}
  R_{T}(\vec{\cmp})
  &\le
    \hat O\brac{R_{T}^{\cA_{\gamma}}(\vec{\cmp}) + \max_{t}(\yt-\yref_{t})^{2}\Log{NT}}\nonumber\\
  &\le
    \hat O\brac{R_{T}^{\cA_{\gamma}}(\vec{\cmp}) + \max_{t}(\yt-\yref_{t})^{2}\Log{T}}\label{eq:grid-tuning-discounted-vaw:full-bound},
\end{align}
where the last line uses
$N=\abs{\cS_{\gamma}}=\log_{b}(\eta_{\max}/\eta_{\min})\le O(\log_{b}(T))$,
then hides $\log(\log)$ factors. Finally,
by \Cref{lemma:grid-tuning-discounted-vaw-existence},
there is indeed a $\gamma^{*}\in[0,1]$ satisfying
$\gamma^{*}=\frac{\sqrt{d\sumtT\half(\yt-\yhint_{t})^{2}}}{\sqrt{d\sumtT\half(\yt-\yhint_{t})^{2}}+\sqrt{P_{T}^{\gamma^{*}}(\vec{\cmp})}}$
and a $\gamma\in\cS_{\gamma}$ such that
\begin{align*}
  R_{T}^{\cA_{\gamma}}(\vec{\cmp})\le O\brac{d\max_{t}(\yt-\yhint_{t})^{2}\Log{T}+b\sqrt{dP_{T}^{\gamma^{*}}(\vec{\cmp})\sumtT(\yt-\yhint_{t})^{2}}}.
\end{align*}
Plugging this back into \Cref{eq:grid-tuning-discounted-vaw:full-bound}
and choosing $\yref_{t}=\yhint_{t}$ proves the result.
\end{proof}

\subsection{Proof of Theorem~\ref{thm:grid-tuning-small-loss}}%
\label{app:grid-tuning-small-loss}
As in \Cref{app:grid-tuning-discounted-vaw},
the proof of \Cref{thm:grid-tuning-small-loss}
follows by applying \Cref{thm:dynamic-regret-tuning}
and then showing that there is a $\cA_{\gamma}$ attaining
the desired regret bound.
We first provide proof of the latter claim in
\Cref{lemma:grid-tuning-small-loss-existence}
for the sake of modularity, so that we can
use it when arguing strongly-adaptive guarantees in \Cref{sec:strongly-adaptive}.
Proof of \Cref{thm:grid-tuning-small-loss}
is proven at the end of this section.
\begin{restatable}{lemma}{GridTuningSmallLossExistence}\label{lemma:grid-tuning-small-loss-existence}
  Under the same conditions as \Cref{lemma:grid-tuning-discounted-vaw-existence},
  suppose each $\cA_{\gamma}$ sets hints
  $\yhint_{t}=\yclip_{t}^{\gamma}=\Clip_{\cB_{t}}(\inner{\xt,\wt^{\gamma}})$,
  where $\cB_{t}=[\yref_{t}-M_{t}, \yref_{t}+M_{t}]$
  and $M_{t}=\max_{s<t}\abs{\y_{s}-\yref_{s}}$.
  Then for any $\vec{\cmp}=(\cmp_{1},\ldots,\cmp_{T})$ in $W$,
  there is a $\gamma^{\Ring}\in[0,1]$ satisfying
  $\gamma^{\Ring}=\frac{\sqrt{d\sumtT\ell_{t}(\cmp_{t})}}{\sqrt{d\sumtT\ell_{t}(\cmp_{t})}+\sqrt{dP_{T}^{\gamma^{\Ring}}(\vec{\cmp})}}$
  and a $\gamma\in\cS_{\gamma}$ such that
  \begin{align*}
    R_{T}^{\cA_{\gamma}}(\vec{\cmp})
    &\le
      O\bigg(d P_{T}^{\gamma_{\min}}(\vec{\cmp})+ d\max_{t}\brac{\yt-\yref_{t}}^{2}\Log{T}\\
    &\qquad
      + b\sqrt{d P_{T}^{\gamma^{\Ring}}(\vec{\cmp})\sumtT\ell_{t}(\cmp_{t})}\bigg),
  \end{align*}
  where $\gamma^{\min}=\min\Set{\gamma\in \cS_{\gamma}}=\frac{2d}{2d+1}$.
\end{restatable}
\begin{proof}
  Using \Cref{lemma:small-loss-template}, for any
  $\vec{\cmp}=(\cmp_{1},\ldots,\cmp_{T})$, $\gamma\in(0,1)$, and
  $\beta\ge\gamma\ge\gamma_{\min}=\frac{2d}{2d+1}$, we have
  \begin{align*}
      R_{T}(\vec{\cmp})
    &\le
      \gamma\lambda\norm{\cmp_{1}}^{2}_{2}+4d\max_{t}(\yt-\yref_{t})^{2}\Log{1+\frac{\sumtT\gamma^{T-t} \norm{\x_{t}}^{2}_{2}}{\lambda d}}\\
    &\qquad
    +2\frac{\beta}{1-\beta} P_{T}^{\beta}(\vec{\cmp})
      +\frac{1-\gamma}{\gamma}2d\sumtT\ell_{t}(\cmp_{t}),
  \end{align*}

  We will
  proceed by showing that there is a $\beta$ and $\gamma$ that suitably balances the summations in
  the last line.
  To this end, recall that by
  \Cref{lemma:good-discount-existence}, there is a $\gamma^{\Ring}$
  satisfying
  \begin{align*}
    \gamma^{\Ring}
    =
    \frac{\sqrt{d\sumtT\ell_{t}(\cmp_{t})}}{\sqrt{d\sumtT\ell_{t}(\cmp_{t})}+\sqrt{ P_{T}^{\gamma^{\Ring}}(\vec{\cmp})}}
  \end{align*}
  Denote $\eta=\frac{\gamma}{1-\gamma}$ and
  $ \eta^{\Ring}=\frac{ \gamma^{\Ring}}{1-\gamma^{\Ring}}=\sqrt{\frac{d\sumtT\ell_{t}(\cmp_{t})}{ P_{T}^{\gamma^{\Ring}}(\vec{\cmp})}}$.
  If $\eta^{\Ring}\ge \eta_{\max}=\frac{\gamma_{\max}}{1-\gamma_{\max}}$, then we can
  take $\beta=\gamma^{\Ring}$ and $\gamma=\gamma_{\max}$ to get
  \begin{align*}
    \frac{\beta}{1-\beta} P_{T}^{\beta}(\vec{\cmp})+\frac{\gamma}{1-\gamma}d\sumtT\ell_{t}(\cmp_{t})
    &=
    \eta^{\Ring} P_{T}^{\gamma^{\Ring}}(\vec{\cmp})+\frac{d\sumtT\ell_{t}(\cmp_{t})}{\eta_{\max}}\\
    &=
      \sqrt{d P_{T}^{\gamma^{\Ring}}(\vec{\cmp})\sumtT\ell_{t}(\cmp_{t})} + \frac{d\sumtT\ell_{t}(\cmp_{t})}{\eta_{\max}}\\
    &\le
      \sqrt{d P_{T}^{\gamma^{\Ring}}(\vec{\cmp})\sumtT\ell_{t}(\cmp_{t})} +\max_{t}\ell_{t}(\cmp_{t}),
  \end{align*}
  where the last line recalls $\eta_{\max}=dT$.
  Otherwise, if
  $\eta^{\Ring}\le\eta_{\min}=\frac{\gamma_{\min}}{1-\gamma_{\min}}=2d$, then
  taking $\beta=\gamma=\gamma_{\min}$ yields
  \begin{align*}
    \eta_{\min} P_{T}^{\gamma_{\min}}(\vec{\cmp})+\frac{d\sumtT\ell_{t}(\cmp_{t})}{\eta_{\min}}
    &\le
    \eta_{\min} P_{T}^{\gamma_{\min}}(\vec{\cmp})+\frac{d\sumtT\ell_{t}(\cmp_{t})}{\eta^{\Ring}}\\
    &=
    2d P_{T}^{\gamma_{\min}}(\vec{\cmp})
      +
      \sqrt{d P_{T}^{\gamma^{\Ring}}(\vec{\cmp})\sumtT\ell_{t}(\cmp_{t})}.
  \end{align*}
  Lastly, if $\eta_{\min}\le \eta^{\Ring}\le\eta_{\max}$, there is a $\eta_{k}=\frac{\gamma_{k}}{1-\gamma_{k}}\in\cS_{\eta}$ such that
  $\eta_{k}\le \eta^{\Ring}\le b\eta_{k}$, so choosing $\beta=\gamma^{\Ring}$ and $\gamma=\gamma_{k}$ yields
  \begin{align*}
    \eta^{\Ring} P_{T}^{\gamma^{\Ring}}(\vec{\cmp})+\frac{d\sumtT\ell_{t}(\cmp_{t})}{\eta_{k}}
    &\le
    \eta^{\Ring} P_{T}^{\gamma^{\Ring}}(\vec{\cmp})+b\frac{d\sumtT\ell_{t}(\cmp_{t})}{ \eta^{\Ring}}\\
    &=
      (b+1)\sqrt{d P_{T}^{\gamma^{\Ring}}(\vec{\cmp})\sumtT\ell_{t}(\cmp_{t})}
  \end{align*}
  Combining the three cases, we have
  \begin{align*}
    2\frac{\beta}{1-\beta} P_{T}^{\beta}(\vec{\cmp})+\frac{1-\gamma}{\gamma}2d\sumtT\ell_{t}(\cmp_{t})
    &\le
    4d  P_{T}^{\gamma_{\min}}(\vec{\cmp})+2\max_{t}\ell_{t}(\cmp_{t})
      +2(b+1)\sqrt{d P_{T}^{\gamma^{\Ring}}(\vec{\cmp})\sumtT\ell_{t}(\cmp_{t})}\\
  \end{align*}
  Hence, overall the regret can be bound as
  \begin{align*}
    R_{T}^{\cA_{\gamma}}(\vec{\cmp})
    &\le
      \gamma\lambda\norm{\cmp_{1}}^{2}_{2}
      +d\max_{t}(\yt-\yclip_{t}^{\gamma})^{2}\Log{1+\frac{\sumtT\gamma^{T-t}\norm{\xt}^{2}_{2}}{\lambda d}}\\
    &\qquad
    +4d  P_{T}^{\gamma_{\min}}(\vec{\cmp})+2\max_{t}\ell_{t}(\cmp_{t})
      +2(b+1)\sqrt{d P_{T}^{\gamma^{\Ring}}(\vec{\cmp})\sumtT\ell_{t}(\cmp_{t})}\\
    &\le
    O\brac{d P_{T}^{\gamma_{\min}}(\vec{\cmp})+ d\max_{t}(\yt-\yref_{t})^{2}\Log{T}+ b\sqrt{d P_{T}^{\gamma^{\Ring}}(\vec{\cmp})\sumtT\ell_{t}(\cmp_{t})}},
  \end{align*}
  where we've applied \Cref{lemma:clipped-predictions} to bound
  $\max_{t}(\yt-\yclip_{t}^{\gamma})^{2}\le 4M_{T+1}^{2} = 4\max_{t}(\yt-\yref_{t})^{2}$.
  Plugging this back into \Cref{eq:grid-tuning-small-loss:full-bound}
  proves the stated bound.
\end{proof}

Now the proof of \Cref{thm:grid-tuning-small-loss} follows
by composing \Cref{thm:dynamic-regret-tuning} and
\Cref{lemma:grid-tuning-small-loss-existence}. The theorem is
restated below for convenience.
\GridTuningSmallLoss*
\begin{proof}
  As in the proof of \Cref{thm:grid-tuning-discounted-vaw}, we
  apply
  \Cref{thm:dynamic-regret-tuning}, from which it follows
  that for any $\vec{\cmp}=(\cmp_{1},\ldots,\cmp_{T})$ in $\R^{d}$ and any $\gamma\in\cS_{\gamma}$,
  the dynamic regret is bounded as
  \begin{align}
    R_{T}(\vec{\cmp})
    &\le
      \hat O\brac{R_{T}^{\cA_{\gamma}}(\vec{\cmp}) + \max_{t}(\yt-\yref_{t})^{2}\Log{NT}}\nonumber\\
    &\le
      \hat O\brac{R_{T}^{\cA_{\gamma}}(\vec{\cmp}) + \max_{t}(\yt-\yref_{t})^{2}\Log{T}}\label{eq:grid-tuning-small-loss:full-bound},
  \end{align}
  where the last line uses
  $N=\abs{\cS_{\gamma}}=\log_{b}(\eta_{\max}/\eta_{\min})\le O(\log_{b}(T))$,
  then hides $\log(\log)$ factors. And using
  \Cref{lemma:grid-tuning-small-loss-existence},
  for any $\vec{\cmp}=(\cmp_{1},\ldots,\cmp_{T})$
  there is a $\gamma^{\Ring}\in[0,1]$ satisfying
  $\gamma^{\Ring}=\frac{\sqrt{d\sumtT\ell_{t}(\cmp_{t})}}{\sqrt{d\sumtT\ell_{t}(\cmp_{t})}+\sqrt{P_{T}^{\Ring}(\vec{\cmp})}}$
  and a $\gamma\in\cS_{\gamma}$ such that
  \begin{align*}
    R_{T}^{\cA_{\gamma}}(\vec{\cmp})
    &\le
    O\brac{d P_{T}^{\gamma_{\min}}(\vec{\cmp})+ d\max_{t}(\yt-\yref_{t})^{2}\Log{T}+ b\sqrt{d P_{T}^{\gamma^{\Ring}}(\vec{\cmp})\sumtT\ell_{t}(\cmp_{t})}},
  \end{align*}
  Plugging this back into \Cref{eq:grid-tuning-small-loss:full-bound}
  completes the proof.
\end{proof}

\section{Adaptive Fixed-share}%
\label{app:adaptive-fixed-share}

\begin{algorithm}
\KwInput{Experts $\cA_1,\ldots,\cA_N$, $p_1\in \Delta_N$}\\
\For{$t=1:T$}{
    Get $y_{t}^{(i)}$ from $\cA_i$ for all $i$\\
    Play $\ybar_t = \sum_{i=1}^Np_{ti}y_{t}^{(i)}$\\
    Observe loss $\ell_{t}(y)=\half(\yt-\y)^{2}$ and let $\ell_{ti}=\ell_{t}(y_{t}^{(i)})$ for all $i$
    \hfill\\
    Let $q_{t+1,i}=\frac{p_{ti}\Exp{-\alpha_t\ell_{ti}}}{\sum_{j=1}^Np_{tj}\Exp{-\alpha_t \ell_{tj}}}$ for all $i$\\
    Choose $\beta_\tpp$ and set $p_{\tpp}=(1-\beta_\tpp)q_\tpp + \beta_\tpp p_1$
}
\caption{Adaptive Fixed-Share}
\label{alg:adaptive-fixed-share}
\end{algorithm}

In this section, we provide for completeness analysis related to the
fixed-share algorithm \cite{cesa2012mirror} with time-varying modulus.
The following is a modest generalization of the analysis of \citet[Theorem
10.3]{hazan2019introduction}. Throughout this section we assume
that the losses $\ell_{t}:\hat\cY\to\R$ are exp-concave in their domain.

\begin{restatable}{theorem}{AdaptiveFixedShare}\label{thm:adaptive-fixed-share}
For all \(t\) let \(\ell_{t}\) be an \(\alpha_{t}\text{-Exp-Concave}\) function and
assume that \(\alpha_t\ge \alpha_\tpp\) for all \(t\). For all $t$, set $\beta_{t}\le \frac{1}{(e+t)\log^{2}(e+t)+1}$.
Then for any $j\in[N]$ and any $[a,b]\subseteq[1,T]$, \Cref{alg:adaptive-fixed-share} guarantees
  \begin{align*}
    \sum_{t=a}^{b}\ell_t(\ypred_{t})-\ell_t(\y_{t}^{(j)})
    &\le
      \frac{1}{\alpha_{b+1}}\sbrac{2\Log{\frac{1}{\beta_{b+1}p_{1j}}}+1}
\end{align*}
\end{restatable}
\begin{proof}
  The heavy lifting is done mostly using \Cref{lemma:fs-one-step-error},
  after which the proof follows by choosing the sequence of mixing parameters $\beta_{t}$.
  Applying \Cref{lemma:fs-one-step-error} and observing the telescoping sum, we have
  \begin{align*}
    \sum_{t=a}^b \ell_t(\ybar_t)-\ell_t\brac{y_t^{(j)}}
    &\le
      \sum_{t=a}^b\frac{1}{\alpha_{t}}\Log{\frac{1}{p_{tj}}}-\frac{1}{\alpha_{\tpp}}\Log{\frac{1}{p_{\tpp,j}}}\\
    &\qquad
      +\sum_{t=a}^b\frac{1}{\alpha_{t}}\Log{\frac{1}{1-\beta_\tpp}}\\
    &\qquad
      +\sum_{t=a}^b\abs{\frac{1}{\alpha_{\tpp}}-\frac{1}{\alpha_{t}}}\Log{\frac{1}{\beta_\tpp p_{1j}}}\\
    &=
      \frac{1}{\alpha_{a}}\Log{\frac{1}{p_{aj}}} -\frac{1}{\alpha_{b+1}}\Log{\frac{1}{p_{b+1,j}}}\\
    &\qquad
      +\sum_{t=a}^b\frac{1}{\alpha_{t}}\Log{\frac{1}{1-\beta_\tpp}}\\
    &\qquad
      +\sum_{t=a}^b\abs{\frac{1}{\alpha_{\tpp}}-\frac{1}{\alpha_{t}}}\Log{\frac{1}{\beta_\tpp p_{1j}}}.
  \end{align*}
  Now observe that with \(\beta_\tpp \le \frac{1}{(e+t)\log^2(e+t)+1}\), using the
  elementary inequality $\Log{1+y}\le y$ we have
  \begin{align*}
    \Log{\frac{1}{1-\beta_\tpp}} = \Log{1+\frac{\beta_\tpp}{1-\beta_\tpp}} \le \frac{\beta_\tpp}{1-\beta_\tpp} = \frac{1}{(e+t)\log^2(e+t)}
  \end{align*}
  so for non-increasing \(\alpha_t\) we have
  \begin{align*}
    \sum_{t=a}^b \frac{1}{\alpha_t}\Log{\frac{1}{1-\beta_\tpp}}
    &\le
      \sum_{t=a}^b\frac{1}{\alpha_t}\frac{1}{(e+t)\log^2(e+t)}\\
    &\le
      \frac{1}{\alpha_{b}}\sum_{t=a}^b\frac{1}{(e+t)\log^2(e+t)}\\
    &\le
      \frac{1}{\alpha_{b}}\int_{e}^{e+b}\frac{1}{y\log^2y}dy \\
    &=
      \frac{1}{\alpha_{b}}\frac{-1}{\Log{y}}\Bigg|_{y=e}^{e+b}\le \frac{1}{\alpha_{b}}
  \end{align*}
  and similarly,
  \begin{align*}
    \sum_{t=a}^b \abs{\frac{1}{\alpha_\tpp}-\frac{1}{\alpha_t}}\Log{\frac{1}{\beta_\tpp p_{1j}}}
    &\le
      \Log{\frac{1}{\beta_{b+1}p_{1j}}}\sum_{t=a}^b \frac{1}{\alpha_\tpp}-\frac{1}{\alpha_t}\\
    &\le
      \frac{1}{\alpha_{b+1}}\Log{\frac{1}{\beta_{b+1}p_{1j}}},
  \end{align*}
  so overall we have
  \begin{align*}
    \sum_{t=a}^b\ell_t(\ybar_t)-\ell_t\brac{y_{t}^{(j)}}
    &\le
    \frac{1}{\alpha_{a}}\Log{\frac{1}{p_{aj}}} -\frac{1}{\alpha_{b+1}}\Log{\frac{1}{p_{b+1,j}}}
      +\frac{\Log{\frac{1}{\beta_{b+1}p_{1j}}}+1}{\alpha_{b+1}}\\
    &=
      \frac{1}{\alpha_{a}}\Log{\frac{1}{p_{aj}}}
      +\frac{\Log{\frac{p_{b+1,j}}{\beta_{b+1}p_{1j}}}+1}{\alpha_{b+1}}\\
    &\le
      \frac{1}{\alpha_{b+1}}\Log{\frac{1}{(1-\beta_{a})q_{aj}+\beta_{a}p_{1j}}}
      +\frac{\Log{\frac{p_{b+1,j}}{\beta_{b+1}p_{1j}}}+1}{\alpha_{b+1}}\\
    &\le
      \frac{1}{\alpha_{b+1}}\sbrac{2\Log{\frac{1}{\beta_{b+1}p_{1j}}}+1}
    &\le
  \end{align*}
\end{proof}

\subsection{Proof of Lemma~\ref{lemma:fs-one-step-error}}%
\label{app:fs-one-step-error}
The following provides an initial one-step bound to work from, which we use
in the proof of \Cref{thm:adaptive-fixed-share}.
\begin{restatable}{lemma}{FSOneStepError}\label{lemma:fs-one-step-error}
  For all \(t\) let \(\ell_{t}\) be an \(\alpha_{t}\text{-Exp-Concave}\) function. Then
  for any \(j\in[N]\), \Cref{alg:adaptive-fixed-share} guarantees
  \begin{align*}
    \ell_{t}(\ybar_{t})-\ell_{t}(y_{t}^{(j)})
    &\le
      \frac{1}{\alpha_{t}}\Log{\frac{1}{p_{tj}}}-\frac{1}{\alpha_{\tpp}}\Log{\frac{1}{p_{\tpp,j}}}\\
    &\qquad
      +\frac{1}{\alpha_{t}}\Log{\frac{1}{1-\beta_\tpp}}\\
    &\qquad
      +\abs{\frac{1}{\alpha_{\tpp}}-\frac{1}{\alpha_{t}}}\Log{\frac{1}{\beta_\tpp p_{1j}}}
  \end{align*}
\end{restatable}
\begin{proof}
By $\alpha_{t}$-Exp-Concavity of \(\ell_{t}\), we have that \(y\mapsto \Exp{-\alpha_{t}\ell_{t}(y)}\) is
concave. Hence, applying Jensen's inequality:
\begin{align*}
  \Exp{-\alpha_{t}\ell_{t}(\ybar_{t})}
  &\ge
    \sum_{i=1}^{N}p_{ti}\Exp{-\alpha_{t}\ell_{t}\brac{y_{t}^{(i)}}}
    =
    \sum_{i=1}^{N}p_{ti}\Exp{-\alpha_{t}\ell_{ti}}
\end{align*}
and taking the natural logarithm of both sides we have
\begin{align*}
  -\alpha_{t}\ell_{t}(\ybar_{t})&\ge \Log{\sum_{i=1}^{N}p_{ti}\Exp{-\alpha_{t}\ell_{ti}}}\\
  \ell_{t}(\ybar_{t})&\le-\frac{1}{\alpha_{t}}\Log{\sum_{i=1}^{N}p_{ti}\Exp{-\alpha_{t}\ell_{ti}}}.
\end{align*}
Hence, for any \(j\in[N]\) we have
\begin{align*}
  \ell_{t}(\ybar_{t})-\ell_{t}\brac{y_{t}^{(j)}}
  &\le
    -\frac{1}{\alpha_{t}}\Log{\sum_{i=1}^{N}p_{ti}\Exp{-\alpha_{t}\ell_{ti}}}-\ell_{tj}\\
  &=
    -\frac{1}{\alpha_{t}}\Log{\sum_{i=1}^{N}p_{ti}\Exp{-\alpha_{t}\ell_{ti}}}+\frac{1}{\alpha_{t}}\Log{\Exp{-\alpha_{t}\ell_{tj}}}\\
  &=
    \frac{1}{\alpha_{t}}\Log{\frac{\Exp{-\alpha_{t}\ell_{tj}}}{\sum_{i=1}^{N}p_{ti}\Exp{-\alpha_{t}\ell_{ti}}}}\\
  &=
    \frac{1}{\alpha_{t}}\Log{\frac{p_{tj}\Exp{-\alpha_{t}\ell_{tj}}}{p_{tj}\sum_{i=1}^{N}p_{ti}\Exp{-\alpha_{t}\ell_{ti}}}}\\
  &=
    \frac{1}{\alpha_{t}}\sbrac{\Log{\frac{q_{\tpp,j}}{p_{tj}}}}\\
  &=
    \frac{1}{\alpha_{t}}\sbrac{\Log{\frac{1}{p_{tj}}}-\Log{\frac{1}{q_{\tpp,j}}}}.
\end{align*}
Adding and subtracting \(\frac{1}{\alpha_{\tpp}}\Log{\frac{1}{p_{\tpp,j}}}\),
\begin{align*}
  \ell_{t}(\ybar_{t})-\ell_{t}\brac{y_{t}^{(j)}}
  &\le
    \frac{1}{\alpha_{t}}\Log{\frac{1}{p_{tj}}}-\frac{1}{\alpha_{\tpp}}\Log{\frac{1}{p_{\tpp,j}}}\\
  &\qquad
    +\frac{1}{\alpha_{\tpp}}\Log{\frac{1}{p_{\tpp,j}}} -\frac{1}{\alpha_{t}}\Log{\frac{1}{q_{\tpp,j}}}\\
  &=
    \frac{1}{\alpha_{t}}\Log{\frac{1}{p_{tj}}}-\frac{1}{\alpha_{\tpp}}\Log{\frac{1}{p_{\tpp,j}}}\\
  &\qquad
    +\underbrace{\frac{1}{\alpha_{t}}\Log{\frac{1}{p_{\tpp,j}}} -\frac{1}{\alpha_{t}}\Log{\frac{1}{q_{\tpp,j}}}}_{\Log{q_{\tpp,j}/p_{\tpp,j}}/\alpha_t}\\
  &\qquad
    +\sbrac{\frac{1}{\alpha_{\tpp}}-\frac{1}{\alpha_{t}}}\Log{\frac{1}{p_{\tpp,j}}}\\
  \intertext{recalling $p_{\tpp,j}=(1-\beta_{\tpp})q_{\tpp,j}+\beta_{\tpp}p_{1j}$,}
  &=
    \frac{1}{\alpha_{t}}\Log{\frac{1}{p_{tj}}}-\frac{1}{\alpha_{\tpp}}\Log{\frac{1}{p_{\tpp,j}}}\\
  &\qquad
    +\frac{1}{\alpha_{t}}\Log{\frac{q_{\tpp,j}}{(1-\beta_\tpp)q_{\tpp,j} + \beta_{\tpp}p_{1j}}}\\
  &\qquad
    +\sbrac{\frac{1}{\alpha_{\tpp}}-\frac{1}{\alpha_{t}}}\Log{\frac{1}{(1-\beta_\tpp)q_{\tpp,j} + \beta_\tpp p_{1j}}}\\
  &\le
    \frac{1}{\alpha_{t}}\Log{\frac{1}{p_{tj}}}-\frac{1}{\alpha_{\tpp}}\Log{\frac{1}{p_{\tpp,j}}}\\
  &\qquad
    +\frac{1}{\alpha_{t}}\Log{\frac{1}{1-\beta_\tpp}}\\
  &\qquad
    +\abs{\frac{1}{\alpha_{\tpp}}-\frac{1}{\alpha_{t}}}\Log{\frac{1}{\beta_\tpp p_{1j}}}
\end{align*}
\end{proof}

\newpage
\section{Strongly-Adaptive Guarantees}%
\label{app:strongly-adaptive}
In this section we provide a formal statement of
the result sketched in \Cref{sec:strongly-adaptive}.
The result follows easily from the results in \Cref{sec:tuning},
after borrowing the geometric covering intervals from \citet{daniely2015strongly}.

\begin{restatable}{theorem}{StronglyAdaptive}\label{thm:strongly-adaptive}
  Let $\cS_{\gamma}$ be the set of discount factors defined in
  \Cref{thm:grid-tuning-discounted-vaw}, let $S$ denote
  a set of geometric covering intervals over $[1,T]$,
  and for each $\gamma\in\cS_{\gamma}$ and $I\in S$, let $\cA_{\gamma,I}$ be
  an instance of \Cref{alg:discounted-vaw} using discount $\gamma$ and applied
  during interval $I$ (and predicts $\yref_{t}$ for $t\notin I$).
  Let $\cA_{\Meta}$
  be an instance of the meta-algorithm characterized in
  \Cref{thm:dynamic-regret-tuning}. Then for any $[s,\tau]\subseteq[1,T]$,
  there is a
  set of disjoint intervals $I_{1},\ldots,I_{K}$ in $S$
  such that $\cup_{i=1}^{K}I_{i}=[s,\tau]$, and moreover,
  for any $\vec{\cmp}=(\cmp_{s},\ldots,\cmp_{\tau})$
  \Cref{alg:clipped-meta} with $\yref_{t}=\yhint_{t}$ guarantees
  \begin{align*}
    R_{[s,\tau]}(\vec{\cmp})\le\hat O\brac{
    d\max_{t}(\yt-\yref_{t})^{2}\log^{2}(T)+
    b\sqrt{d P_{[s,\tau]}^{\gamma^{*}}(\vec{\cmp})\sum_{t\in[s,\tau]}(\yt-\yhint_{t})^{2}}
    }
  \end{align*}
  where
  $P_{[s,\tau]}^{\gamma^{*}}(\vec{\cmp})=\sum_{i=1}^{K}P_{I_{i}}^{\gamma_{i}^{*}}(\vec{\cmp})$
  and each
  $\gamma^{*}_{i}\in[0,1]$ satisfies
  $\gamma_{i}^{*}=\frac{\sqrt{\frac{d}{2}\sum_{t\in I_{i}}(\yt-\yhint_{t})^{2}}}{\sqrt{\frac{d}{2}\sum_{t\in I_{i}}(\yt-\yhint_{t})^{2}}+\sqrt{P_{I_{i}}^{\gamma^{*}_{i}}(\vec{\cmp})}}$.

  If we instead suppose each $\cA_{\gamma,I}$ sets hints
  as in \Cref{thm:grid-tuning-small-loss}, then
  for any $\vec{\cmp}=(\cmp_{s},\ldots,\cmp_{\tau})$
  \Cref{alg:clipped-meta} guarantees
  \begin{align*}
    R_{[s,\tau]}(\vec{\cmp})
    &\le
      \hat O\brac{dP_{[s,\tau]}^{\gamma_{\min}}(\vec{\cmp}) + d \max_{t}(\yt-\yref_{t})^{2}\log^{2}(T)
      +b\sqrt{dP_{[s,\tau]}^{\gamma^{\Ring}}(\vec{\cmp})\sum_{t\in[s,\tau]}\ell_{t}(\cmp_{t})}
      }
  \end{align*}
  where
  $P_{[s,\tau]}^{\gamma^{\Ring}}(\vec{\cmp})=\sum_{i=1}^{K}P_{I_{i}}^{\gamma^{\Ring}_{i}}(\vec{\cmp})$
  and each $\gamma^{\Ring}_{i}\in[0,1]$ satisfies
  $\gamma^{\Ring}_{i}=\frac{\sqrt{d\sum_{t\in I_{i}}\ell_{t}(\cmp_{t})}}{\sqrt{d\sum_{t\in I_{i}}\ell_{t}(\cmp_{t})}+\sqrt{P_{I_{i}}^{\gamma^{\Ring}_{i}}(\vec{\cmp})}}$.
\end{restatable}
\begin{proof}
  For any $[s,\tau]\subseteq[1,T]$, \citet[Lemma 1.2]{daniely2015strongly}
  shows that there exists a disjoint set of intervals $I_{1},\ldots, I_{K}$
  in $S$ such that $\cup_{i=1}^{K} I_{i}=[s,\tau]$ and $K\le O(\log(\tau-s))$.
  Hence, we can decompose $\sum_{i=1}^{K}R_{I_{i}}(\vec{\cmp})$,
  so
  applying \Cref{thm:dynamic-regret-tuning} to each of these
  sub-intervals, for any $\gamma\in\cS_{\gamma}$ we have:
  \begin{align}
    R_{[s,\tau]}(\vec{\cmp})= \sum_{i=1}^{K}R_{I_{i}}(\vec{\cmp})
    &\le
      \sum_{i=1}^{K}\hat O\brac{R_{I_{i}}^{\cA_{\gamma, I_{i}}}(\vec{\cmp})+\max_{t}(\yt-\yhint_{t})^{2}\Log{N\abs{I_{i}}}}\nonumber\\
    &\le
      \hat O\brac{\sum_{i=1}^{K} R_{I_{i}}^{\cA_{\gamma, I_{i}}}(\vec{\cmp})+K\max_{t}(\yt-\yhint_{t})^{2}\Log{N(\tau-s)}}\nonumber\\
    &\le
      \hat O\brac{\sum_{i=1}^{K} R_{I_{i}}^{\cA_{\gamma, I_{i}}}(\vec{\cmp})+\max_{t}(\yt-\yhint_{t})^{2}\log^{2}(T)}\label{eq:strongly-adaptive:full-bound},
  \end{align}
  where $\hat O(\cdot)$ hides $\log(\log)$ factors and the last line bounds $K\le O(\log(\tau-s))\le O(\log(T))$
  and $N\le O(T\Log{T})$.
  The bound on $N$ can be seen from the fact that
  $\abs{\cS_{\gamma}}\le O(\log(T))$, and from the fact that $S$ is
  constructed as $S=\cup_{i=1}^{\Floor{\log(T)}}S_{i}$ where
  $S_{i}=\Set{[k2^{i},(k+1)2^{i}-1]:k=0,1,\ldots}$, from which it
  is easily seen that $\abs{S}\le O(T)$ by observing that each $S_{i}$ has at
  most $T/2^{i}$ intervals, hence summing them all up yields $\abs{S}=\sum_{i=1}^{\Floor{\Log{T}}}\abs{S_{i}}\le O(T)$.

  Now for any interval $I_{i}$, \Cref{lemma:grid-tuning-discounted-vaw-existence}
  shows that there is a $\gamma^{*}_{i}\in[0,1]$ satisfying
  $\gamma^{*}_{i}=\frac{\sqrt{d\sum_{t\in I_{i}}\half(\yt-\yhint_{t})^{2}}}{\sqrt{d\sum_{t\in I_{i}}\half(\yt-\yhint_{t})^{2}}+\sqrt{P_{I_{i}}^{\gamma^{*}_{i}}(\vec{\cmp})}}$
  and a $\gamma\in \cS_{\gamma}$ such that
  \begin{align*}
    R_{I_{i}}^{\cA_{\gamma,I_{i}}}(\vec{\cmp})
    &\le
      O\brac{d\max_{t}(\yt-\yhint_{t})^{2}\Log{\abs{I_{i}}}+ b\sqrt{dP_{I_{i}}^{\gamma^{*}_{i}}(\vec{\cmp})\sum_{t\in I_{i}}(\yt-\yhint_{t})^{2}}}
  \end{align*}
  so summing these up and applying Cauchy-Schwarz inequlity leads to
  \begin{align*}
    \sum_{i=1}^{K}R_{I_{i}}^{\cA_{\gamma,I_{i}}}(\vec{\cmp})
    &\le
      O\brac{Kd\max_{t}(\yt-\yhint_{t})^{2}\Log{\abs{I_{i}}}+\sum_{i=1}^{K} b\sqrt{dP_{I_{i}}^{\gamma^{*}_{i}}(\vec{\cmp})\sum_{t\in I_{i}}(\yt-\yhint_{t})^{2}}}\\
    &\le
      O\brac{d\max_{t}(\yt-\yhint_{t})^{2}\log^{2}(\tau-s)+b\sqrt{dP_{[s,\tau]}^{\gamma^{*}}(\vec{\cmp})\sum_{t\in [s,\tau]}(\yt-\yhint_{t})^{2}}}
  \end{align*}
  where we've defined $P_{[s,\tau]}^{\gamma^{*}}(\vec{\cmp})=\sum_{i=1}^{K}P_{I_{i}}^{\gamma^{*}_{i}}(\vec{\cmp})$.
  Plugging this back into \Cref{eq:strongly-adaptive:full-bound},
  overall we may bound:
  \begin{align*}
    R_{[s,\tau]}(\vec{\cmp})
    &\le
      \hat O\brac{d\max_{t}(\yt-\yref_{t})^{2}\log^{2}(T)+b\sqrt{dP_{[s,\tau]}^{\gamma^{*}}(\vec{\cmp})\sum_{t\in [s,\tau]}(\yt-\yhint_{t})^{2}}}
  \end{align*}
  where we've chosen $\yhint_{t}=\yref_{t}$ for simplicity.

  An identical argument holds for the second statement:
  for any interval $I_{i}$, \Cref{lemma:grid-tuning-small-loss-existence}
  shows that there is a $\gamma^{\Ring}_{i}\in[0,1]$ satisfying
  $\gamma^{\Ring}_{i}=\frac{\sqrt{d\sum_{t\in I_{i}}\ell_{t}(\cmp_{t})}}{\sqrt{d\sum_{t\in I_{i}}\ell_{t}(\cmp_{t})}+\sqrt{P_{I_{i}}^{\gamma^{\Ring}_{i}}(\vec{\cmp})}}$
  and a $\gamma\in \cS_{\gamma}$ such that
  \begin{align*}
    R_{I_{i}}^{\cA_{\gamma,I_{i}}}(\vec{\cmp})
    &\le
      O\brac{dP_{I_{i}}^{\gamma_{\min}}(\vec{\cmp})+d\max_{t}(\yt-\yref_{t})^{2}\Log{\abs{I_{i}}}+ b\sqrt{dP_{I_{i}}^{\gamma^{\Ring}_{i}}(\vec{\cmp})\sum_{t\in I_{i}}\ell_{t}(\cmp_{t})}}
  \end{align*}
  so summing these up and applying Cauchy-Schwarz inequality
  again leads to
  \begin{align*}
    \sum_{i=1}^{K}R_{I_{i}}^{\cA_{\gamma,I_{i}}}(\vec{\cmp})
    &\le
      O\brac{dP_{[s,\tau]}^{\gamma_{\min}}(\vec{\cmp})+Kd\max_{t}(\yt-\yhint_{t})^{2}\Log{\abs{I_{i}}}+\sum_{i=1}^{K} b\sqrt{dP_{I_{i}}^{\gamma^{\Ring}}(\vec{\cmp})\sum_{t\in I_{i}}\ell_{t}(\cmp_{t})}}\\
    &\le
      O\brac{dP_{[s,\tau]}^{\gamma_{\min}}(\vec{\cmp})+d\max_{t}(\yt-\yhint_{t})^{2}\log^{2}(\tau-s)+b\sqrt{dP_{[s,\tau]}^{\gamma^{\Ring}}(\vec{\cmp})\sum_{t\in [s,\tau]}\ell_{t}(\cmp_{t})}}
  \end{align*}
  where we've defined
  $P_{[s,\tau]}^{\gamma^{*}}(\vec{\cmp})=\sum_{i=1}^{K}P_{I_{i}}^{\gamma^{*}_{i}}(\vec{\cmp})$,
  so
  plugging this back into \Cref{eq:strongly-adaptive:full-bound},
  overall we may bound:
  \begin{align*}
    R_{[s,\tau]}(\vec{\cmp})
    &\le
      \hat O\brac{dP_{[s,\tau]}^{\gamma_{\min}}(\vec{\cmp})+d\max_{t}(\yt-\yref_{t})^{2}\log^{2}(T)+b\sqrt{dP_{[s,\tau]}^{\gamma^{\Ring}}(\vec{\cmp})\sum_{t\in [s,\tau]}\ell_{t}(\cmp_{t})}},
  \end{align*}
  where we've defined $P_{[s,\tau]}^{\gamma^{\Ring}}=\sum_{i=1}^{K}P_{I_{i}}^{\gamma^{\Ring}_{i}}(\vec{\cmp})$.

\end{proof}

\subsection{Matching the Exp-concave Guarantee in Unbounded Domains}%
\label{app:baby-comparison}
Recall from \Cref{sec:lb} that in the Exp-concave
setting, the algorithm of \citet{baby2021optimal}
achieves a dynamic regret bound of the form
$R_{T}(\vec{\cmp})\le \tilde O\brac{T^{1/3}C_{T}^{2/3}}$ for $C_{T}=\sum_{t=1}^{T-1}\norm{\cmp_{t}-\cmp_{\tmm}}_{1}$.
Our strongly-adaptive guarantees in \Cref{thm:strongly-adaptive}
show that a bound of this form can be achieved even in the
unbounded domain setting.
To see why, note that the essential intuition of \citet{baby2021optimal} is that
if we have access to a \emph{strongly-adaptive} algorithm
guaranteeing $R_{[a,b]}(u)\le O(\log(b-a))$ \emph{static} regret on all intervals
$[a,b]\subseteq[1,T]$, then to attain the desired bound up to log terms
it suffices to
show
that there \emph{exists} a set of intervals $\left\{I_{1},\ldots,I_{N}\right\}$ partitioning
$[1,T]$ such that $N\le T^{1/3}C_{T}^{2/3}$ and
that the dynamic regret is bounded by the static regrets
over the partition, leading to regret matching  $O(T^{1/3}C_{T}^{2/3})$ up to
logarithmic terms.

Our strongly-adaptive
guarantee in \Cref{thm:strongly-adaptive} actually achieves a stronger guarantee than is
necessary to invoke the above argument, by
guaranteeing
$O\big(\log(b-a)\vee \sqrt{dP_{[a,b]}^{\gamma}(\vec{\cmp})|b-a|}\big)$ \emph{dynamic} regret on every
interval $[a,b]$, and hence as a special case we have $O(\log(b-a))$ static regret on
each interval as well. A similar partitioning argument
then provides an analogous $T^{1/3}C_{T}^{2/3}$ bound, even in unbounded domains.
If this is surprising, note that the exp-concave (and hence bounded
domain) restriction is only really used
to provide an algorithm which achieves logarithmic static regret, not to
construct the essential partition. In the online linear regression setting, we do not
need exp-concavity to guarantee logarithmic static regret --- the VAW forecaster
can provide the necessary guarantee even in an unbounded domain.

\newpage
\section{Supporting Lemmas}%
\label{app:lemmas}

The following provides a useful relation between the squared loss and its
Bregman divergence.
\begin{restatable}{lemma}{LossDivergence}\label{lemma:loss-divergence}
  Let $\ell_{t}(\w)=\half(\yt-\inner{\xt,\wt})^{2}$. Then for any $\cmp,\w\in W$,
  \begin{align*}
    D_{\ell_{t}}(\cmp|\w)=\half\inner{\xt, \cmp-\w}^{2}
  \end{align*}
\end{restatable}
\begin{proof}
  By definition of Bregman divergence, we have:
  \begin{align*}
    D_{\ell_{t}}(\cmp|\w)=\ell_{t}(\cmp)-\ell_{t}(\w)-\inner{\grad\ell_{t}(\w),\cmp-\w}.
  \end{align*}
  Expanding the definition of $\ell_{t}$, we have
  \begin{align*}
    \ell_{t}(\cmp)-\ell_{t}(\w)
    &=
      \half(\yt-\inner{\xt,\cmp})^{2}-\half(\yt-\inner{\xt,\w})^{2}\\
    &=
      \half\yt^{2}+\half\inner{\xt,\cmp}^{2}-\yt\inner{\xt,\cmp}-\half\yt^{2}-\half\inner{\xt,\w}^{2}+\yt\inner{\xt,\w}\\
    &=
      \half\inner{\xt,\cmp}^{2}-\half\inner{\xt,\w}^{2}+\yt\inner{\xt,\w-\cmp}.
  \end{align*}
  Moreover, we have
  \begin{align*}
    -\inner{\grad\ell_{t}(\w),\cmp-\w}
    &=
      \inner{(y_{t}-\inner{\xt,\w})\xt, \cmp-\w}\\
    &=
      -\yt\inner{\xt,\w-\cmp}+\inner{\xt,\w}^{2}-\inner{\xt,\w}\inner{\xt,\cmp},
  \end{align*}
  so combining with the previous display we have
  \begin{align*}
    \ell_{t}(\cmp)-\ell_{t}(\w)-\inner{\grad\ell_{t}(\w),\cmp-\w}
    &=
      \half\inner{\xt,\cmp}^{2}-\half\inner{\xt,\w}^{2}+\yt\inner{\xt,\w-\cmp}\\
    &\qquad
      -\yt\inner{\xt,\w-\cmp}+\inner{\xt,\w}^{2}-\inner{\xt,\w}\inner{\xt,\cmp}\\
    &=
      \half\inner{\xt,\cmp}^{2}+\half\inner{\xt,\w}^{2}-\inner{\xt,\w}\inner{\xt,\cmp}\\
    &=
      \half\brac{\inner{\xt,\cmp}-\inner{\xt,\w}}^{2}\\
    &=
      \half\inner{\xt,\cmp-\w}^{2}.
  \end{align*}
\end{proof}

The following provides a discounted version of the log-determinant lemma.
\begin{restatable}{lemma}{AdaptiveDiscountedLogDet}\label{lemma:adaptive-discounted-log-det}
  Let $\gamma\in(0,1]$, $\lambda > 0$, $\xt\in\R^{d}$, and define
  $M_{0}=\lambda I$ and $M_{t}=x_{t}x_{t}^{\top}+\gamma M_{\tmm}$ for each $t>0$.
  Then for any sequence $\Delta_{1},\Delta_{2},\ldots$ in $\R$,
  \begin{align*}
    \sumtT\Delta_{t}^{2}\norm{\xt}^{2}_{M_{t}^{\inv}}\le
      d\Log{1/\gamma}\Delta_{1:T}^{2}+\max_{t}\Delta_{t}^{2}d\Log{1+\frac{\sumtT\gamma^{T-t}\norm{\xt}^{2}_{2}}{\lambda d}}
  \end{align*}
\end{restatable}
\begin{proof}
  By definition we have $M_{t}=\xt\xt^{\top}+\gamma M_{\tmm}$, so re-arranging and
  taking the determinant of both sides we have
  \begin{align*}
    \Det{\gamma M_{\tmm}}
    &=
      \Det{M_{t}-\xt\xt^{\top}}
      =
      \Det{M_{t}}\Det{I-M_{t}^{-\half}\xt\xt^{\top}M_{t}^{-\half}}\\
    &=
      \Det{M_{t}}(1-\norm{\xt}^{2}_{M_{t}^{\inv}})
  \end{align*}
  where the last line uses the fact that $\Det{I-yy^{\top}}=1-\norm{y}^{2}_{2}$.
  Re-arranging, using $\Det{\gamma M_{\tmm}}=\gamma^{d}\Det{M_{\tmm}}$, and using the fact that
  $1-x\le -\Log{x}$ we have
  \begin{align*}
    \sumtT\Delta_{t}^{2}\norm{\xt}^{2}_{M_{t}^{\inv}}
    &=
      \sumtT \Delta_{t}^{2}\sbrac{1-\frac{\gamma^{d}\Det{M_{\tmm}}}{\Det{M_{t}}}}\\
    &\le
      \sumtT \Delta_{t}^{2}\Log{\frac{\Det{M_{t}}}{\gamma^{d}\Det{M_{\tmm}}}}\\
    &=
      \sumtT \Delta_{t}^{2}d\Log{1/\gamma}+\sumtT\Delta_{t}^{2}\Log{\frac{\Det{M_{t}}}{\Det{M_{\tmm}}}}\\
    &\le
      d\Log{1/\gamma}\Delta_{1:T}^{2}+\max_{t}\Delta_{t}^{2}\Log{\prod_{t=1}^{T}\frac{\Det{M_{t}}}{\Det{M_{\tmm}}}}\\
    &=
      d\Log{1/\gamma}\Delta_{1:T}^{2}+\max_{t}\Delta_{t}^{2}\Log{\frac{\Det{M_{T}}}{\Det{M_{0}}}}.
  \end{align*}
  Observe that $\Det{M_{0}}=\Det{\lambda I}=\lambda^{d}$, and using AM-GM inequality
  we have
  \begin{align*}
    \Det{M_{T}}
    &\le
      \brac{\frac{\Tr{M_{t}}}{d}}^{d}
      =\brac{\frac{\Tr{\lambda \gamma^{T} I + \sumtT\gamma^{T-t}\xt\xt^{\top}}}{d}}^{d}\\
    &=
      \brac{\frac{d\lambda\gamma^{T}+\sumtT\gamma^{T-t}\norm{\xt}^{2}_{2}}{d}}^{d},
  \end{align*}
  Hence
  $\frac{\Det{M_{T}}}{\Det{M_{0}}}\le \brac{\frac{d\lambda\gamma^{T}+\sumtT\gamma^{T-t}\norm{\xt}^{2}_{2}}{d \lambda}}^{d}$,
  so overall we have
  \begin{align*}
    \sumtT\Delta_{t}^{2}\norm{\xt}^{2}_{M_{t}^{\inv}}
    &\le
      d\Log{1/\gamma}\Delta_{1:T}^{2}+\max_{t}\Delta_{t}^{2}\Log{\brac{\frac{d\lambda\gamma^{T}+\sumtT\gamma^{T-t}\norm{\xt}^{2}_{2}}{\lambda d}}^{d}}\\
    &=
      d\Log{1/\gamma}\Delta_{1:T}^{2}+\max_{t}\Delta_{t}^{2}d\Log{\frac{d\lambda\gamma^{T}+\sumtT\gamma^{T-t}\norm{\xt}^{2}_{2}}{\lambda d}}\\
    &\le
      d\Log{1/\gamma}\Delta_{1:T}^{2}+\max_{t}\Delta_{t}^{2}d\Log{1+\frac{\sumtT\gamma^{T-t}\norm{\xt}^{2}_{2}}{\lambda d}}\\
  \end{align*}
\end{proof}

Note that the \Cref{lemma:adaptive-discounted-log-det} also
immediately gives us the usual log determinant lemma as a special case where $\gamma=1$:
\begin{restatable}{lemma}{LogDet}\label{lemma:log-det}
  Let $\lambda > 0$, $\xt\in\R^{d}$, and define
  Let $M_{0}=\lambda I$ and $M_{t}=x_{t}x_{t}^{\top}+M_{\tmm}$ for each $t>0$.
  Then for any sequence $\Delta_{1},\Delta_{2},\ldots$ in $\R$,
  \begin{align*}
    \sumtT\Delta_{t}^{2}\norm{\xt}^{2}_{M_{t}^{\inv}}\le
      d\max_{t}\Delta_{t}^{2}\Log{1+\frac{\sumtT\norm{\xt}^{2}_{2}}{\lambda d}}
  \end{align*}
\end{restatable}

The following lemma is common in adaptive online learning and provided for completeness.
\begin{restatable}{lemma}{SqrtBounds}\label{lemma:sqrt-bounds}
  Let $a_{1},\ldots, a_{T}$ be arbitrary non-negative numbers in $\R$.
  Then
  \begin{align*}
    \sqrt{\sumtT a_{t}}\le \sumtT\frac{a_{t}}{\sqrt{\sum_{s=1}^{t}a_{s}}}\le 2\sqrt{\sumtT a_{t}}
  \end{align*}
\end{restatable}
\begin{proof}
  By concavity of $x\mapsto\sqrt{x}$, we have
  \begin{align*}
    \sqrt{a_{1:t}}-\sqrt{a_{1:\tmm}}\ge \frac{a_{t}}{2\sqrt{a_{1:t}}},
  \end{align*}
  so summing over $t$ and observing the resulting telescoping sum yields
  \begin{align*}
    \sumtT\frac{a_{t}}{\sqrt{a_{1:t}}} \le 2 \sumtT\sqrt{a_{1:t}}-\sqrt{a_{1:\tmm}} = 2\sqrt{a_{1:T}}.
  \end{align*}
  For the lower bound, observe that
  \begin{align*}
    \sumtT\frac{a_{t}}{\sqrt{a_{1:t}}}\ge \sumtT\frac{a_{t}}{\sqrt{a_{1:T}}} = \frac{a_{1:T}}{\sqrt{a_{1:T}}}=\sqrt{a_{1:T}}
  \end{align*}
\end{proof}

\end{document}